\documentclass[10pt,twocolumn,letterpaper]{article}

\usepackage[pagenumbers]{wacv}              %
\usepackage[accsupp]{axessibility}  %

\definecolor{wacvblue}{rgb}{0.21,0.49,0.74}
\usepackage[pagebackref,breaklinks,colorlinks,allcolors=wacvblue]{hyperref}

\usepackage{colortbl}
\usepackage{lipsum}
\usepackage{mdframed}
\usepackage{multirow}
\usepackage{overpic}
\usepackage{soul}
\usepackage{tikz}
\usepackage[normalem]{ulem}
\usepackage{xcolor}

\newcommand{\best}{\cellcolor{red!20}}
\newcommand{\sbest}{\cellcolor{orange!20}}
\newcommand{\tbest}{\cellcolor{yellow!20}}

\newcommand{\hlc}[2][yellow]{{%
    \colorlet{foo}{#1}%
    \sethlcolor{foo}\hl{#2}}%
}

\def\wacvPaperID{1453} %
\def\confName{WACV}
\def\confYear{2026}

\title{Pointmap-Conditioned Diffusion for Consistent Novel View Synthesis}

\author{Thang-Anh-Quan Nguyen$^{1,2}$\quad 
Nathan Piasco$^{1}$\quad
Luis Rold{\~a}o$^{1}$\quad
Moussab Bennehar$^{1}$\\
Dzmitry Tsishkou$^{1}$\quad
Laurent Caraffa$^{3}$\quad
Jean-Philippe Tarel$^{2}$\quad
Roland Br{\'e}mond$^{2}$\\
$^{1}$Noah's Ark, Huawei Paris Research Center, France\\
$^{2}$COSYS, Gustave Eiffel University, France\\
$^{3}$LASTIG, IGN-ENSG, Gustave Eiffel University, France\\
{}
}

\begin{document}

\twocolumn[{
\renewcommand\twocolumn[1][]{#1}
\maketitle
\vspace{-1cm}
\begin{center}
    \centering
            \begin{overpic}[width=1.0\linewidth]{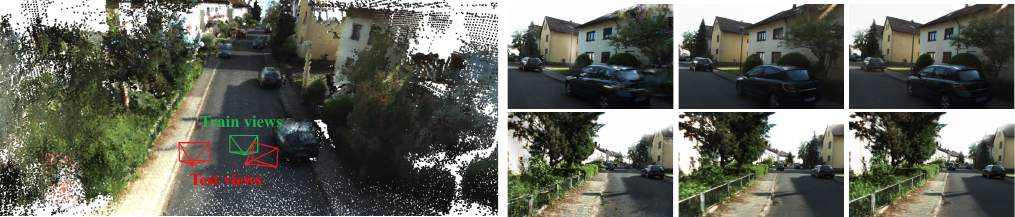}
            \put(55.9, 21.7){\small 3DGS}
            \put(70.2, 21.7){\small PointmapDiff}
            \put(83.9, 21.7){\small 3DGS+PointmapDiff}
            \end{overpic} 
    \captionof{figure}{PointmapDiff is a method that can perform extrapolated view synthesis in urban scenes. We present viewpoints generated at $45^\circ$ angle to the right (first row) and at $1.5m$ position to the left (second row). Our approach significantly outperforms the baselines when rendering viewpoints beyond the original recorded trajectory, whereas 3DGS~\cite{kerbl20233d} struggles with severe artifacts.}
    \label{fig:teaser}
\end{center}
}]

\maketitle
\begin{abstract}
Synthesizing extrapolated views remains a difficult task, especially in urban driving scenes, where the only reliable sources of data are limited RGB captures and sparse LiDAR points. To address this problem, we present PointmapDiff, a framework for novel view synthesis that utilizes pre-trained 2D diffusion models. Our method leverages point maps (\ie, rasterized 3D scene coordinates) as a conditioning signal, capturing geometric and photometric priors from the reference images to guide the image generation process. With the proposed reference attention layers and ControlNet for point map features, PointmapDiff can generate accurate and consistent results across varying viewpoints while respecting geometric fidelity. Experiments on real-life driving data demonstrate that our method achieves high-quality generation with flexibility over point map conditioning signals (\eg, dense depth map or even sparse LiDAR points) and can be used to distill to 3D representations such as 3D Gaussian Splatting for improving view extrapolation.
\end{abstract}
    
\section{Introduction}
\label{sec:intro}

Reconstruction of urban driving scenes plays a crucial role in understanding and advancing autonomous driving systems. Recently, neural rendering techniques such as Neural Radiance Fields (NeRFs)~\cite{mildenhall2021nerf, ost2021neural, wu2023mars, yang2023emernerf, nguyen2024rodus} and 3D Gaussian Splatting (3DGS)~\cite{kerbl20233d, zhou2024drivinggaussian, yan2024street} have demonstrated remarkable potential in synthesizing photorealistic street views, allowing autonomous vehicles to be trained and tested in more diverse and complex scenarios.

Despite these advancements, a significant challenge persists in the form of extrapolation, where the model struggles to render images from viewpoints that differ significantly from the recorded data. Since most training images are captured from vehicle-mounted cameras with low overlap on a single trajectory, the neural reconstruction models primarily learn to interpolate on this trajectory rather than extrapolate to far-away views. This results in a degraded rendering quality, with noticeable blur and artifacts when synthesizing views from extreme angles and positions. Addressing this limitation is crucial for maximizing the utility of reconstructed street scenes, ensuring autonomous driving simulations remain accurate even when operating in regions that are not directly captured during training.

To this extent, we introduce PointmapDiff, a framework designed to leverage pre-trained 2D diffusion models for novel view synthesis (NVS) by incorporating 3D structure into 2D diffusion features. 
In urban driving environments, LiDAR provides reliable geometric information with broader coverage, whereas relying solely on RGB images is insufficient for capturing the scene. To address this, PointmapDiff leverages ControlNet conditioning on point maps, 2D projections of 3D coordinates from the scene’s point cloud, and combines them with features extracted from reference images.
This conditioning allows the model to capture relevant geometric relationships between viewpoints.
Additionally, PointmapDiff utilizes a reference cross-view attention module, ensuring the implicit transfer of information from reference views to the generated target views.

Our approach offers two key benefits: first, point maps establish better correspondences between viewpoints compared to RGB, which require texture details and constant lighting; and second, the model can adapt point maps derived from sparse LiDAR data as geometric guidance for generation, in scenarios where establishing correspondences is particularly challenging.
Finally, we show that PointmapDiff can be used to generate views that are aligned with LiDAR scans. This also serves as supplementary supervision for refining scene representation, such as 3DGS, beyond the initial training trajectories. By integrating the benefits of distilling 3D scene modeling with diffusion-based image generation, PointmapDiff achieves state-of-the-art performance in extrapolated-view synthesis, delivering high-quality, consistent renderings of unobserved views.
\noindent To summarize, our main contributions are as follows:
\begin{itemize}
    \item we propose a point map-conditioned generative framework that can synthesize viewpoints from a single or multiple reference views,
    \item by effectively capturing features and correspondences from point maps, our results consistently respect both geometric and appearance information from reference views and LiDAR scans,
    \item we showcase PointmapDiff's performance in urban reconstruction, given restrictions in sparse points input, as well as its effectiveness in single-image NVS and object manipulation.
\end{itemize}

\section{Related Work}
\label{sec:related_work}

\begin{figure}[t]
    \centering
    \begin{subfigure}{0.40\linewidth}
        \begin{tikzpicture}
        \node[anchor=south west,inner sep=0] (image) at (0,0) {                \includegraphics[width=\linewidth]{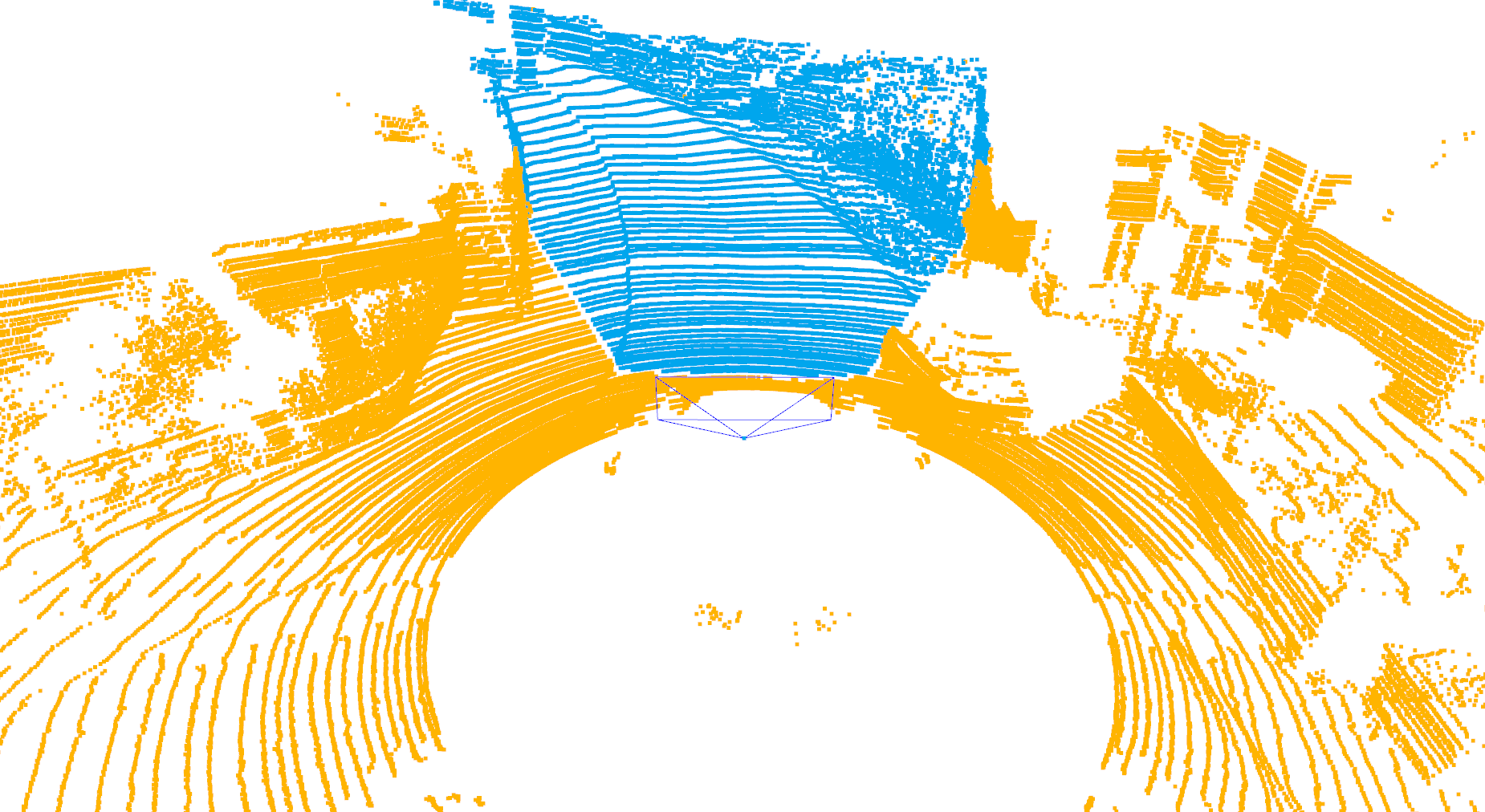}};
        \begin{scope}[x={(image.south east)},y={(image.north west)}]
            \draw[red,thick] (0.58,0.42) rectangle (0.7,0.62);
        \end{scope}
        \end{tikzpicture}
        \caption*{LiDAR Scan}
    \end{subfigure}
    \begin{subfigure}{0.29\linewidth}
        \includegraphics[width=\linewidth]{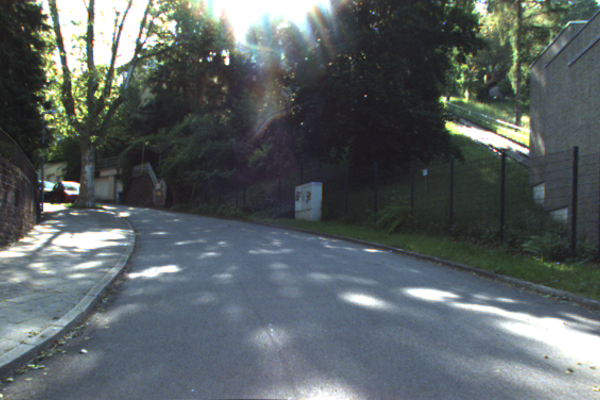}
        \caption*{Source view}
    \end{subfigure}
    \begin{subfigure}{0.32\linewidth}
        \includegraphics[width=\linewidth]{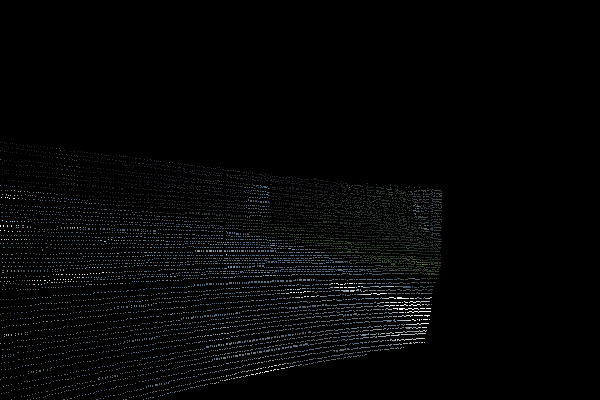}
        \caption*{Warped view}
    \end{subfigure}
    \begin{subfigure}{0.32\linewidth}
        \includegraphics[width=\linewidth]{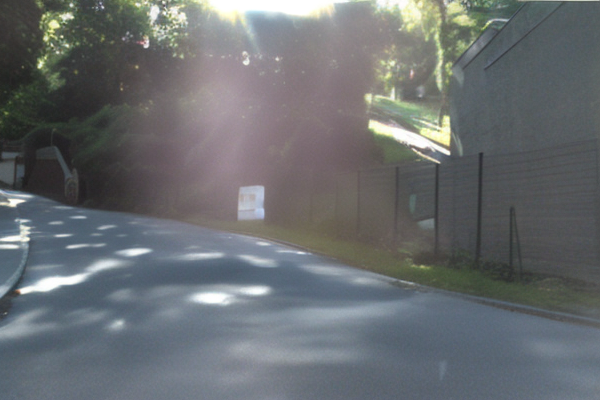}
        \caption*{Baseline}
    \end{subfigure}
    \begin{subfigure}{0.32\linewidth}
        \begin{tikzpicture}
        \node[anchor=south west,inner sep=0] (image) at (0,0) {                \includegraphics[width=\linewidth]{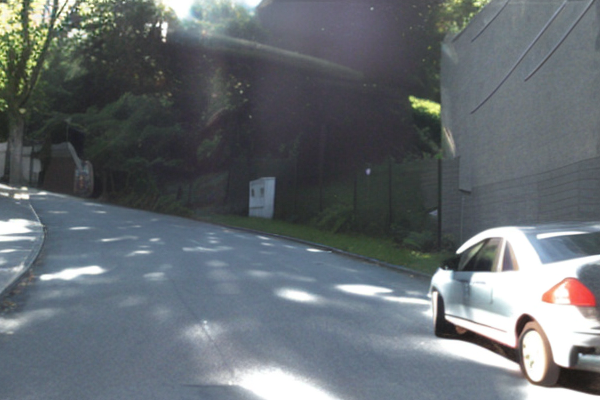}};
        \begin{scope}[x={(image.south east)},y={(image.north west)}]
            \draw[red,thick] (0.7,0.45) rectangle (0.99,0.01);
        \end{scope}
        \end{tikzpicture}  
        \caption*{Ours}
    \end{subfigure}
    \caption{From a reference image and synchronized LiDAR scan, while the image can observe only a small part (\textcolor{blue}{blue}) of the scene, the geometric information from the rest of the LiDAR scan (\textcolor{orange}{orange}) can still be used to generate meaningful content. We label the cars that appear in both the LiDAR scan and the generated image in \textcolor{red}{red}, denoting the advantage of our method compared to other baselines.}
    \label{fig:rebuttal}
    \vspace{-0.5cm}
\end{figure}

\noindent\textbf{Novel View Synthesis.}
The goal of NVS is to generate realistic and visually coherent images of a specific instance or scene from camera viewpoints that have not been observed before. This involves taking one or more existing views of the scene and synthesizing new views while ensuring consistency. NVS can be categorized into two types based on how viewpoints are generated: \textit{View Interpolation}, where the synthesized viewpoints lie within the given input views distribution, and \textit{View Extrapolation}, which involves generating viewpoints outside the input range, often requiring the model to infer a larger amount of content.
 
Many modern view interpolation methods are \textit{reconstruction-based} and built upon NeRF~\cite{mildenhall2021nerf}, 3GDS~\cite{kerbl20233d}, and their derivatives~\cite{tewari2022advances}, which describe a scene as radiance fields to fit the observed images. They enable 3D representation by capturing photos of a real scene and optimizing the underlying geometry and appearance. However, these methods typically require extensive per-scene fitting and only allow for rendering the scene from viewpoints in the training pose distribution. As a result, they usually struggle to generate realistic details in faraway viewpoints. Moreover, capturing detailed scenes requires hundreds to thousands of images, while insufficient scene coverage can lead to optimization issues, resulting in inaccurate geometry, artifacts, and blurry renderings.

On the other hand, most existing extrapolation methods are \textit{generative-based} and rely on training generative models to take available reference images and camera viewpoints as conditions, and directly generate new views. These methods are designed to work with minimal input (\eg, a single image) and rely on general knowledge from large models and datasets.
ReconFusion~\cite{wu2024reconfusion} use priors from CLIP image embedding~\cite{radford2021learning} and pixelNeRF's~\cite{yu2021pixelnerf} features for enabling 3D-awareness. Other works~\cite{ren2022look, yu2023long, tseng2023consistent, tang2023mvdiffusion, gao2024cat3d, yu2024polyoculus} designed special attention mechanisms based on epipolar geometry, local neighborhoods, or camera's ray embeddings~\cite{sitzmann2021light}. GenWarp~\cite{seo2024genwarp},  MultiDiff~\cite{muller2024multidiff}, and ViewCrafter~\cite{yu2024viewcrafter} focus on implicit geometric warping signals using Monocular Depth Estimation (MDE)~\cite{ranftl2020towards, bhat2023zoedepth}. 
Still, closing the domain gap between indoor and outdoor scenes remains challenging as MDE becomes less accurate, while sparse LiDAR data makes it difficult to obtain sufficient warping information. 

\noindent\textbf{Extrapolation in Street View Reconstruction.} 
In driving scenes, the training camera distribution is often biased towards forward-facing movements, which severely limits the vehicle's field of view. Additionally, these straight trajectories could lead to overfitting in methods that rely only on camera parameters. To improve the rendering quality of neural rendering models for viewpoints distant from the training views, several methods augment training with synthesized viewpoints from external generative models. For instance, 
VEGS~\cite{hwang2024vegs} employs a diffusion model with per-scene LoRA~\cite{hu2022lora} and Perturb-and-Average Scoring~\cite{wang2023score} to enhance details in extrapolated views. However, this approach lacks control over geometry and thus depends heavily on strong geometric priors, such as normal supervision. Without such priors, it would require longer training times for the distillation loss to converge. Similarly, SGD~\cite{yu2025sgd} trains a diffusion model with a ControlNet conditioned on two adjacent frames and the dense depth prediction of the current frame, focusing on few-shot setups. Yet, it uses patchified CLIP image embeddings~\cite{radford2021learning} as guidance, which provide high-level semantic information but lack precise spatial detail. This results in inconsistencies in the generated images that can negatively impact the overall 3DGS training. In contrast, our approach focuses on improving the quality of the generative model, enabling more accurate and stable results. We believe that a reliable model with better adaptation to the data modalities commonly found in driving scenarios can significantly boost the performance. 
FreeVS~\cite{wang2024freevs} utilizes colorized LiDAR point clouds to generate pseudo-images for conditioning, making it the most comparable approach to ours. However, like many other baselines, it frequently struggles in regions with absent RGB coverage (\cref{fig:rebuttal}). This limitation presents a significant challenge to the safety and reliability of autonomous perception.
Moreover, instead of training the entire architecture across large datasets with numerous parameters, we design our method to optimize only a minimal subset of parameters by leveraging ControlNet~\cite{zhang2023adding}. We demonstrate that it eliminates the need for extensive fine-tuning and still achieves satisfactory results.

\section{Method}
\label{sec:method}

\begin{figure*}
    \centering    
    \begin{subfigure}{0.85\linewidth}
        \includegraphics[width=\linewidth]{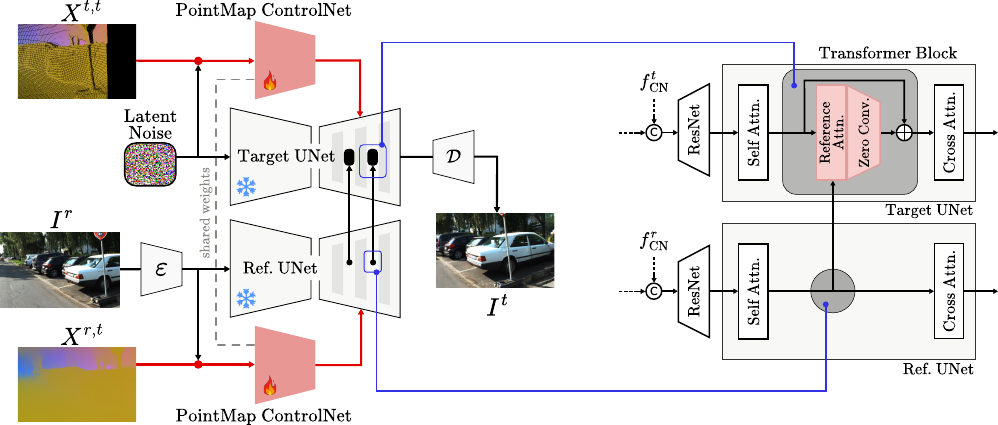}
    \end{subfigure}
    \caption{\textbf{Method overview.} (left) PointmapDiff is trained in the latent space of a fixed VAE with encoder $\mathcal{E}$ and decoder $\mathcal{D}$. Given a reference RGB image $I^{r}$ and the corresponding geometry $D^{r}$, we obtain a pair of point maps $\{X^{r,t}, X^{t,t}\}$ as inputs. We predict the target image $I^{t}$ given the geometry signal from the target point map, and information comes from the reference U-Net. Particularly, two Pointmap ControlNets are employed to extract geometric feature correspondences and concatenate \textcircled{c} them with the intermediate SD feature maps. We freeze the original SD model and only train the Pointmap ControlNet and the reference attention module.
    (right) We extract reference features using our reference U-Net. These augmented features are integrated into the target U-Net through a reference-guided cross-view attention mechanism, which is added \footnotesize{$\bigoplus$} \small throughout the target U-Net.} 
    \label{fig:architecture}
    \vspace{-0.5cm}
\end{figure*}

\subsection{Preliminaries}
\textbf{Diffusion Models}~\cite{ho2020denoising}
are probabilistic models designed to learn the underlying data distribution $p(x)$ by starting from a Gaussian distributed variable $x_{T}$ and gradually denoising it to recover the original data sample $x_{0}$, which simulates the reverse process of a fixed forward (noise-adding) Markov Chain.

In particular, we leverage Latent Diffusion Models (LDM)~\cite{rombach2022high}, which utilize a pre-trained Variational Auto-Encoder (VAE)~\cite{kingma2013auto} to map image data from pixel space into a compressed latent space with lower dimensionality and perform diffusion process in that latent space. This reduces computational complexity, memory footprint, and enables conditioning on other modalities, such as text during generation, while still preserving details.
Typically, to learn the denoising process, the network, U-Net~\cite{ronneberger2015u} in this case, is trained to predict the noise by minimizing:
\begin{equation}
    \mathcal{L}(\theta)=\mathbb{E}_{\epsilon, \tau}\left[\|\epsilon_{\theta}(z_{\tau}, \tau,\mathbf{c})-\epsilon\|^{2}_{2}\right],
\end{equation}
where $\epsilon_{\theta}$ is the noise prediction network with parameters $\theta$, $\tau\sim\mathcal{U}(0, T)$ is the time step, $z_{\tau}$ is the noisy latent at $\tau$, $\epsilon\sim \mathcal{N}(0, I)$ is the additive Gaussian noise, and $\mathbf{c}$ denotes the user-specified condition signal, which is used for the conditional generation.

\noindent\textbf{ControlNet}~\cite{zhang2023adding}
is a versatile network that allows the addition of conditioning into a pre-trained Stable Diffusion (SD) model. 
It has been demonstrated to support various types of input conditioning, such as depth, sketches, and semantic maps, by injecting conditional image features into trainable copies of the original SD encoder blocks, allowing SD to generate images that are coherent with the input condition.
A key advantage of ControlNet is its ability to resist overfitting during fine-tuning, allowing it to retain the original model’s performance. This makes it particularly useful for incorporating 3D awareness~\cite{wang2023freereg, xu20243difftection} into diffusion models without compromising their 2D semantic quality. Nonetheless, this ability has not yet been fully exploited in view synthesis applications.

\noindent\textbf{Problem Statement.} 
Given a reference RGB image $I^{r}$ with its geometric cue $D^{r}$ (derived from a depth sensor, estimated depth map or LiDAR data), we aim to generate a novel view $I^{t}$ from a relative viewpoint $P_{r\rightarrow t}\in SE(3)$ and RGB camera intrinsics $K\in\mathbb{R}^{3\times 3}$. In the latent space, our objective becomes:
\begin{equation}
    z^{t}\sim p(z^{t}|z^{r}, D^{r}, P_{r\rightarrow t}, K),
\end{equation}
where $z^{t}, z^{r}$ are the latent representations for $I^{t}, I^{r}$ and can be decoded through the VAE's decoder.

\subsection{Architecture}
Our approach comprises a two-stream architecture, the reference U-Net takes an input view image $I^{r}$ and produces a feature $f^{r}$. Concurrently, the target U-Net takes a noisy latent and generates a novel view image $I^{t}$, by integrating the input feature $f^{r}$ into its internal novel view feature $f^{t}$. To provide the diffusion model with the depth-based correspondence, we generate a pair of reference and target point maps $\{X^{r,t}, X^{t,t}\}$ and inject them into the model using ControlNets. An overview of the model's architecture is shown in \cref{fig:architecture}.

\subsection{Pointmap ControlNet.}
The advantages of point map have been explored in DUSt3R~\cite{wang2024dust3r}. Point maps encapsulate the geometry of the scene, the relation between pixels and 3D points, and the relationship between two viewpoints.
Such powerful representation can be easily applied to a variety of Multi-View downstream tasks, such as point matching and relative pose estimation.
Since the potential of point maps remains underexplored, this work delves into their advantages within the framework of diffusion models.

We first visit the term \textit{point map}, a point map $X\in\mathbb{R}^{W\times H\times 3}$ describes a mapping between image pixels and 3D scene points. The point map $X$ of the observed scene can be straightforwardly obtained from the camera intrinsic $K$ and the ground-truth depth $D$ as $X_{i,j}=K^{-1}D_{i,j}[i,j,1]^{T}$, where each pixel represents the projected point coordinate. Here, $X$ is expressed in the camera coordinate frame, but in practice, it can further be denoted as $X^{n,m}$, which is the point map $X^{n}$ from camera $n$ expressed in camera $m$’s coordinate frame:
\begin{equation}
    X^{n,m}=h^{-1}(P_{n\rightarrow m}h(X^n)),
\end{equation}
where $P_{n\rightarrow m}$ is the relative pose for images $m$ and $n$, and $h : (x, y, z) \rightarrow (x, y, z, 1)$ is the homogeneous mapping.
We show that the use of point maps provides robust alignment across views by encoding consistent 3D spatial information; when two pixels across different views share the same point map value, they correspond to the same location in the global coordinate frame. Unlike RGB images, which are sensitive to variations in textures, lighting, and shading, point maps offer explicit and more stable representations, making them particularly beneficial for enforcing consistency in 3D-aware tasks.

Inspired by depth-to-image generation, we utilize ControlNet and inject point maps to enhance the 3D awareness of diffusion features. Specifically, we select pairs of images with known relative camera poses and train the ControlNet to condition on the two point maps $\{X^{r,t}, X^{t,t}\}$.
Suppose $\mathcal{F}(\cdot; \Theta)$ is an SD block, with parameters $\Theta$, in particular, the original ControlNet block copies from pre-trained SD's as $\mathcal{F}(\cdot; \Theta')$ and accompanied by two zero convolutions $\mathcal{Z}(\cdot; \Theta_{z1})$, $\mathcal{Z}(\cdot; \Theta_{z2})$. 
Since the geometric features induced by the point map condition in ControlNet are designed to be aligned with the latent inputs, they are processed through the zero-initialized convolutions and subsequently added to the spatial layers of the U-Net:
\begin{equation}
\begin{split}
    f_{CN}^{m} = &\underbrace{\mathcal{F}(z^{m}; \Theta)}_{\text{semantic features}}
    \\
    + &\underbrace{\mathcal{Z}\!\left(\mathcal{F}(z^{m} + \mathcal{Z}(\gamma(X^{m,t}); \Theta_{z1}); \Theta'); \Theta_{z2}\right)}_{\text{geometric features}},
\end{split}
\end{equation}
where $f_{CN}^{m} \text{~with~} m\in\{r, t\}$ is the set of residual signals, which are augmented with the extracted geometric features to join in the features of the U-Net. The point map is processed using positional encoding~\cite{tancik2020fourier} function $\gamma(\cdot)$.

These two shared-weight ControlNets help extract the intermediate feature correspondences between the reference and target point maps. Since both point maps are aligned in the same target view coordinate system, the model can naturally follow not only semantic but also geometric correlations between the reference and target views.

\subsection{Reference-Guided Cross-View Attention.}
Our next step is to learn an attention mechanism between the reference and target features, ensuring consistency between views. We introduce reference attention and inject it after the self-attention layer in the main target U-Net. This enables the model to capture the corresponding relationships from the reference views.
In this module, we change the keys and values corresponding to the output image $I^{t}$ with those of the reference image $I^{r}$. Formally, the output of our reference attention layer is given by:
\begin{multline}
    \text{RefAttn}(Q^{t}, K^{r}, V^{r}) = \text{softmax}\left(\frac{Q^{t}{K^{r}}^{T}}{\sqrt{d}}\right)\cdot V^{r}\\
    \text{with}~Q^{t}=W^{Q}f^{t}; K^{r}=W^{K}f^{r}; V^{r}=W^{V}f^{r},
\end{multline}
where $W^{Q}, W^{K}, W^{V}$ are learnable projection matrices~\cite{vaswani2017attention} for the feature inputs $f^{r}, f^{t}$. We further initialize this attention module with the weights from the self-attention module. The output is then passed through a zero convolution layer and added back to the information flow:
\begin{equation}
    f = f + \mathcal{Z}\left(\text{RefAttn}(Q^{t}, K^{r}, V^{r}), \Theta_{z}\right).
\end{equation}
We further demonstrate the effectiveness of this design in \cref{sec:implementation}.

\subsection{Training}
\noindent\textbf{Implementation Details.}
To construct the point maps used in training, we first generate depth maps of reference views using an MDE or depth completion model~\cite{zhang2023completionformer, wang2024dust3r}. Additionally, we incorporate raw LiDAR data to introduce sparse modality by randomly providing point maps derived directly from LiDAR scans for the reference and target views.
Using raw LiDAR input, our approach achieves two key benefits: (1) it improves point correspondences for more effective pixel transfer, even when dealing with sparse point distribution, and (2) it encourages the novel view to adhere to the geometric structure imposed by the LiDAR data, leading to more coherent and structure-aware output.

The point maps are then normalized to a fixed range of $[-1, 1]$, followed by a positional encoding. This reduces the model's sensitivity to 3D scale ambiguities.
To ensure sufficient correspondence between training pairs, we measure the overlap ratio between two point maps and select only training pairs where the overlap is more than 20\%.

\noindent\textbf{Data Augmentation.}
We apply random cropping to the image to simulate lateral translation. Our method utilizes 3D geometry directly and does not require conditioning on the camera's intrinsic or extrinsic parameters.
This strategy helps reduce camera-related ambiguities and allows us to generate training pairs from a single image, making it particularly beneficial for urban driving datasets, where movement is restricted to forward and backward motion.

\noindent\textbf{Training Objective.}
We leverage the pre-trained SD v1.5 model for both U-Net branches to inherit its generalization ability. We freeze the VAE, the U-Nets, and train only the reference attention modules with the Pointmap ControlNet by minimizing the following cost function:
\begin{equation}
    \mathcal{L}(\theta)=\mathbb{E}_{\epsilon, \tau}\left[\|\epsilon_{\theta}(z^{t}_{\tau}, z^{r}, X^{r,t}, X^{t,t}, P_{r\rightarrow t}, K, \tau)-\epsilon\|^{2}_{2}\right]
\end{equation}
on a dataset containing pairs of source view image $I^{r}$, target view image $I^{t}$ that are encoded into $z^{r}, z^{t}$ respectively, their camera information $\{P_{r\rightarrow t}, K\}$, and point maps $\{X^{r,t}, X^{t,t}\}$. We adopt DDIM sampler~\cite{song2020denoising} and add noise at $\tau$ to the target latent $z^{t}_{\tau}$ while keeping the reference latent $z^{r}$ clean.
We use AdamW optimizer~\cite{loshchilov2017decoupled} with a learning rate of 1e-4, applying cosine scheduler and train the model for 500k iterations with a batch size of 4 at the resolution of $512 \times 320$. Other training parameters remain set to their default values. During inference, we fix the sampling steps at 50.

\section{Experiments}
\label{sec:eval}

We prepare two experiments to evaluate our model. First, we demonstrate how PointmapDiff can be used to enhance the rendering quality of 3DGS~\cite{kerbl20233d} at extrapolated viewpoints. Second, we assess the model’s performance without 3DGS, showcasing its ability to generate views from a single input. We train and evaluate our model on the KITTI-360~\cite{liao2022kitti} and Waymo~\cite{sun2020scalability} datasets.

\subsection{Extrapolation in Street View Reconstruction}
We fine-tune 3DGS on synthesized views to enhance the rendering quality on extrapolated views. We construct evaluation sets that (1) look left and right by rotating the camera by $\pm 45^\circ$ around the camera's y-axis, (2) shift the camera lateral to the heading direction by $\pm\{2,4\}m$, and (3) look downward by rotating the
camera by $10^\circ$ around the camera's x-axis while flying upward by $1m$.

\def\fgsize{0.16}

\begin{figure*}
\centering
\setlength{\tabcolsep}{0.002\linewidth}
\renewcommand{\arraystretch}{0.8}
\begin{tabular}{lcccccc}
    {} & Reference & NeRF~\cite{tancik2023nerfstudio} & 3DGS~\cite{kerbl20233d} & SGD~\cite{yu2025sgd} & VEGS~\cite{hwang2024vegs} & 3DGS+PointmapDiff\\
    
    \multirow{1}{*}[13mm]{\rotatebox[origin=c]{90}{Interpolate}} & 
    \includegraphics[width=\fgsize\textwidth]{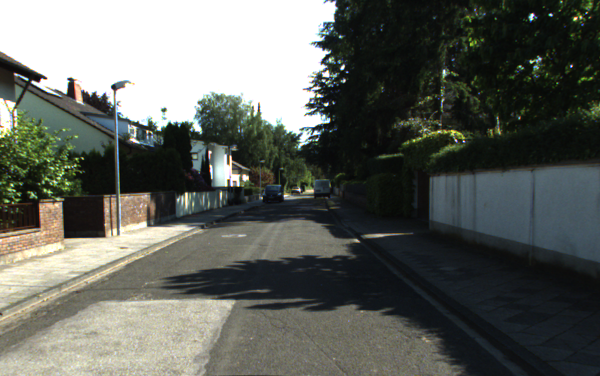} &
    \includegraphics[width=\fgsize\textwidth]{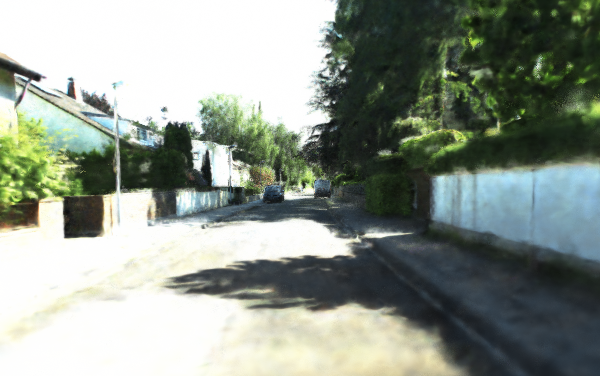} &
    \includegraphics[width=\fgsize\textwidth]{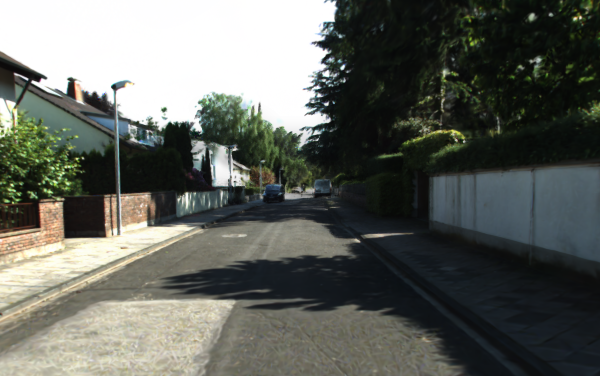} & 
    \includegraphics[width=\fgsize\textwidth]{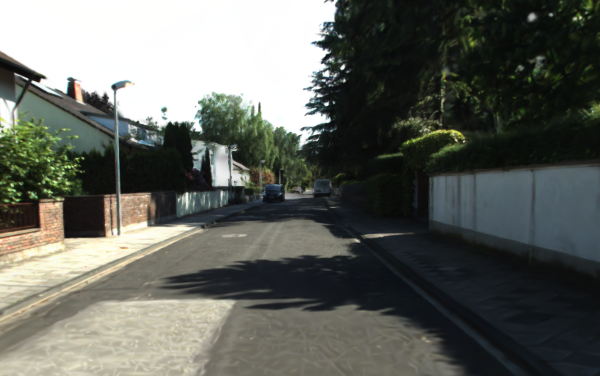} & 
    \includegraphics[width=\fgsize\textwidth]{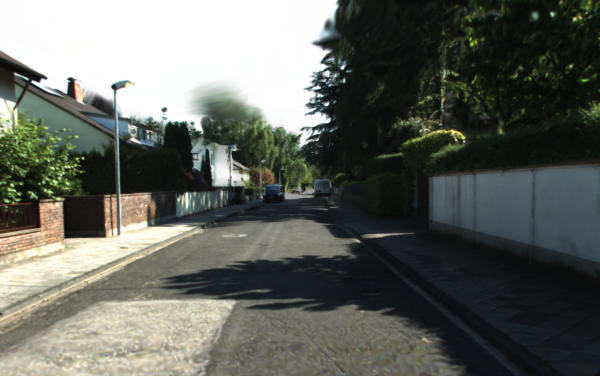} & 
    \includegraphics[width=\fgsize\textwidth]{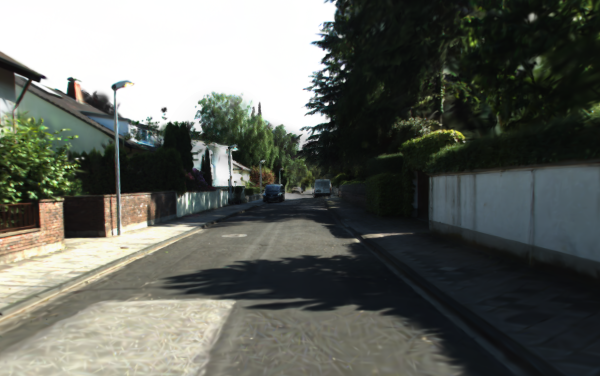}\\
    
    \multirow{2}{*}[14mm]{\rotatebox[origin=c]{90}{Rotate $45^\circ$}} & 
    \includegraphics[width=\fgsize\textwidth]{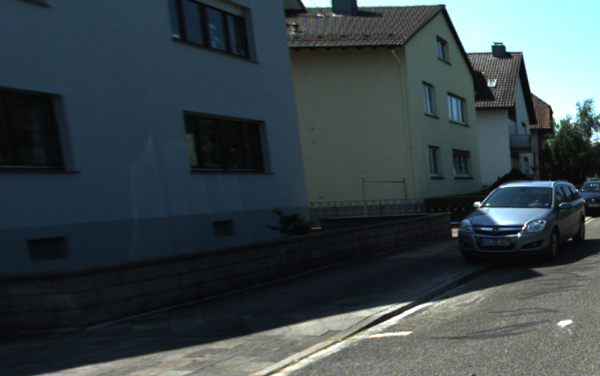} & 
    \includegraphics[width=\fgsize\textwidth]{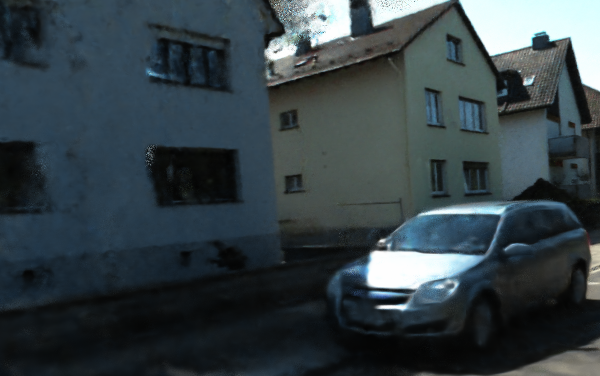} &
    \includegraphics[width=\fgsize\textwidth]{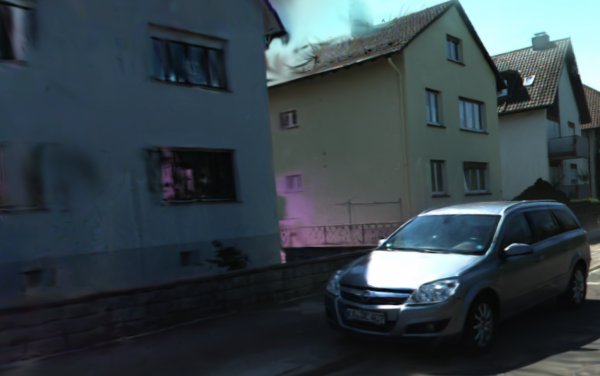} & 
    \includegraphics[width=\fgsize\textwidth]{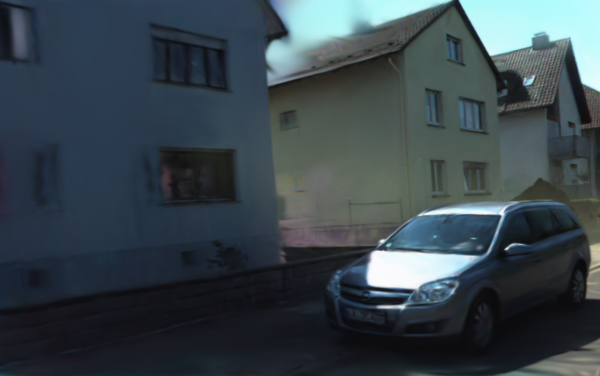} & 
    \includegraphics[width=\fgsize\textwidth]{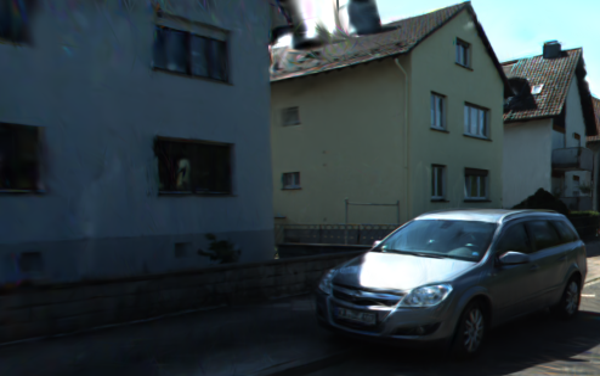} & 
    \includegraphics[width=\fgsize\textwidth]{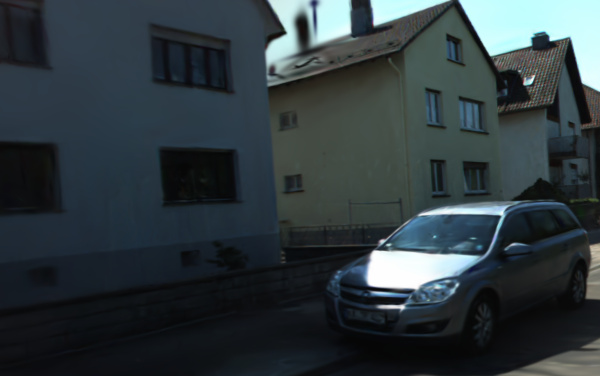}\\
    
    \multirow{2}{*}[13mm]{\rotatebox[origin=c]{90}{Shift 2m}} &
    \includegraphics[width=\fgsize\textwidth]{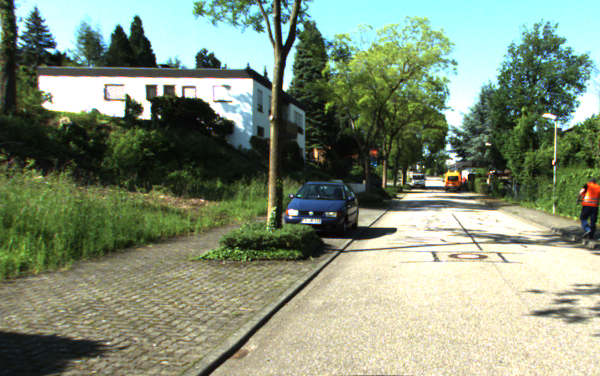} & 
    \includegraphics[width=\fgsize\textwidth]{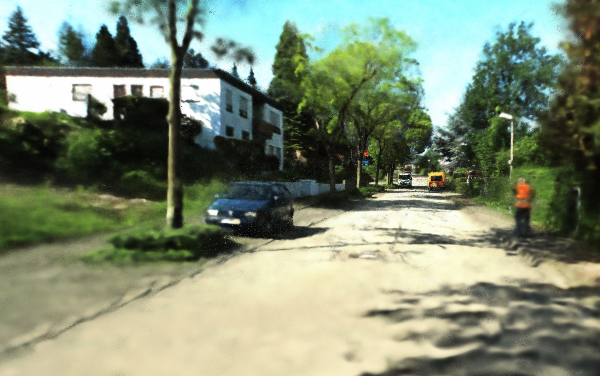} &
    \includegraphics[width=\fgsize\textwidth]{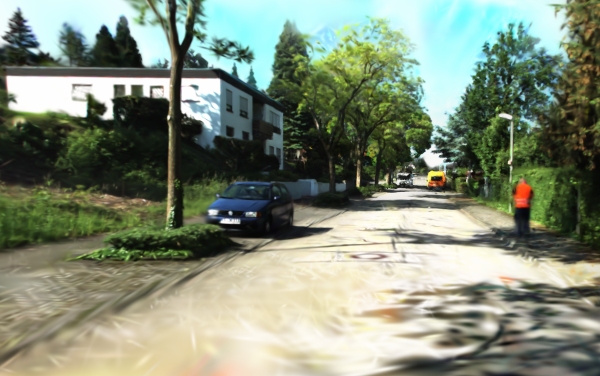} & 
    \includegraphics[width=\fgsize\textwidth]{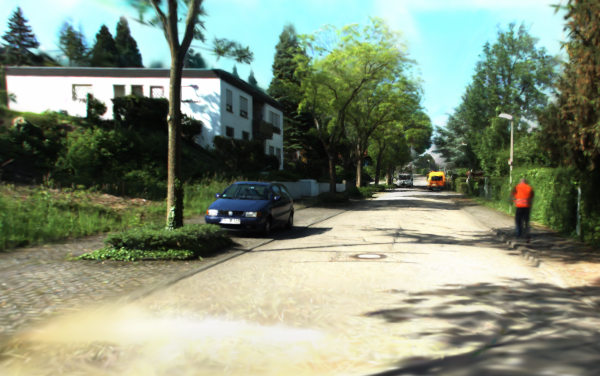} & 
    \includegraphics[width=\fgsize\textwidth]{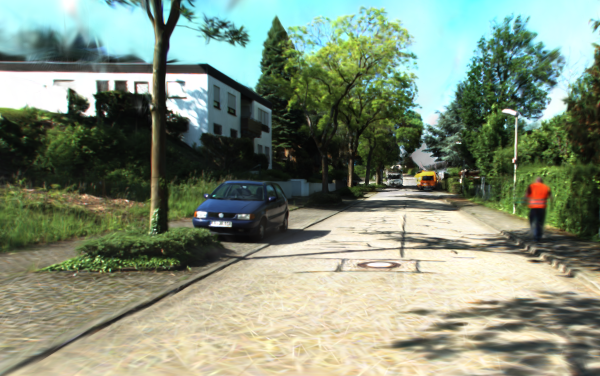} & 
    \includegraphics[width=\fgsize\textwidth]{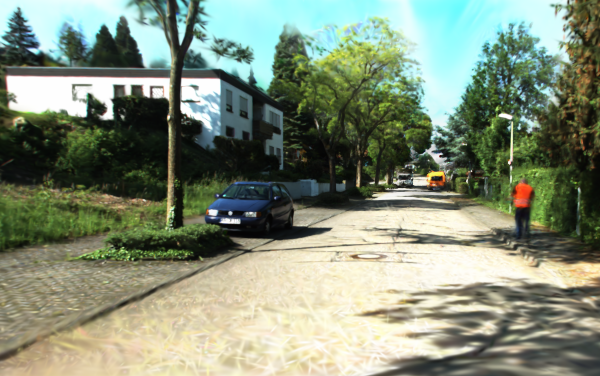}\\
    
    \multirow{2}{*}[15mm]{\rotatebox[origin=c]{90}{Upward 1m}} &
    \includegraphics[width=\fgsize\textwidth]{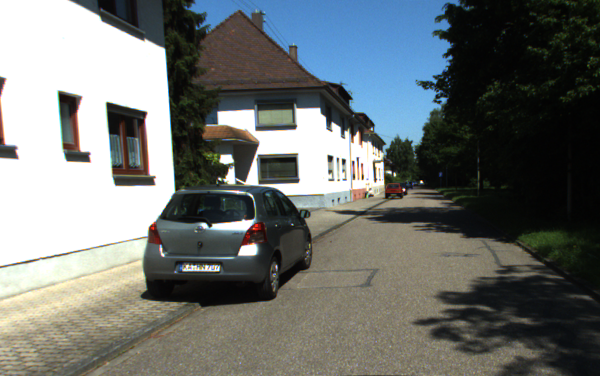} &
    \includegraphics[width=\fgsize\textwidth]{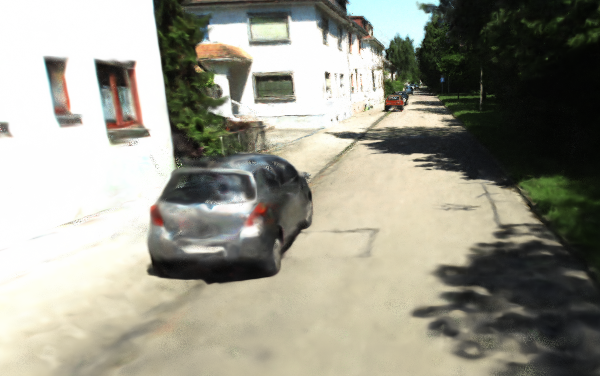} &
    \includegraphics[width=\fgsize\textwidth]{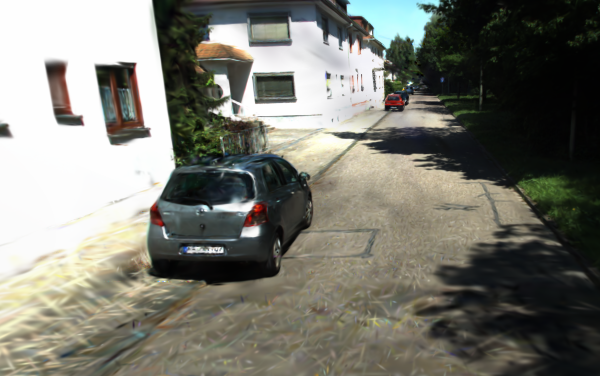} & 
    \includegraphics[width=\fgsize\textwidth]{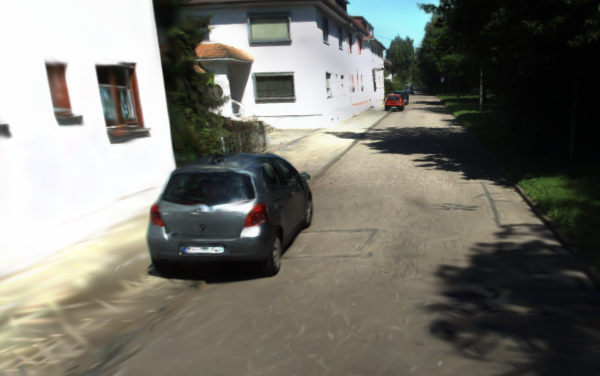} & 
    \includegraphics[width=\fgsize\textwidth]{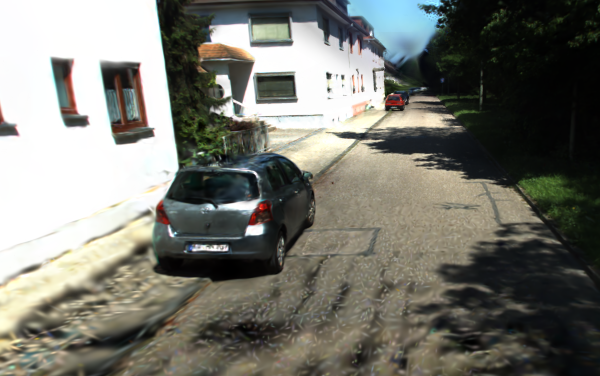} & 
    \includegraphics[width=\fgsize\textwidth]{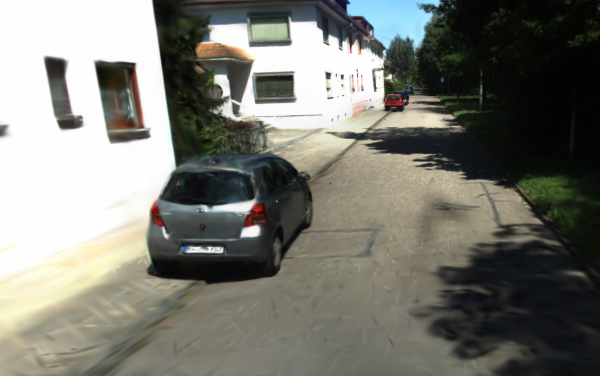}\\
\end{tabular}
\caption{Qualitative comparison on KITTI-360~\cite{liao2022kitti}.
We demonstrate three scenarios: rotating, shifting, and flying upward. The test view represents the conventional camera sampled from forward-facing trajectories. We also include training images that provide the best available coverage as a reference.}
\label{fig:extrapolate_kitti}
\end{figure*} 

\def\fgsize{0.16}

\begin{figure*}
\centering
\setlength{\tabcolsep}{0.002\linewidth}
\renewcommand{\arraystretch}{0.8}
\begin{tabular}{lcccccc}
    {} & Reference & NeRF~\cite{tancik2023nerfstudio} & 3DGS~\cite{kerbl20233d} & 3DGS+ViewCrafter & 
    FreeVS~\cite{wang2024freevs} &
    3DGS+PointmapDiff\\
    \multirow{1}{*}[15mm]{\rotatebox[origin=c]{90}{Interpolate}} & 
    \includegraphics[width=\fgsize\textwidth]{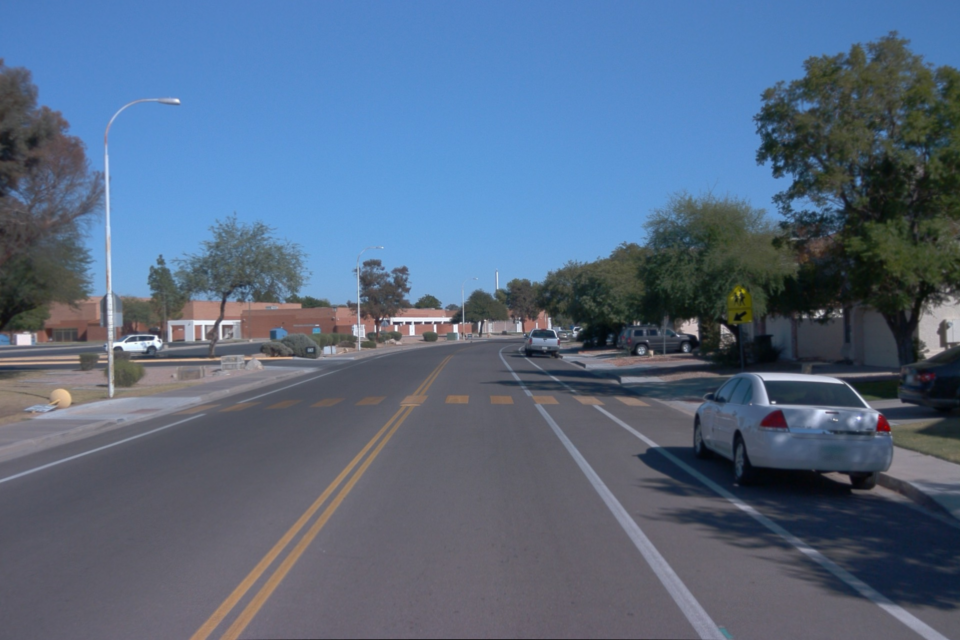} &
    \includegraphics[width=\fgsize\textwidth]{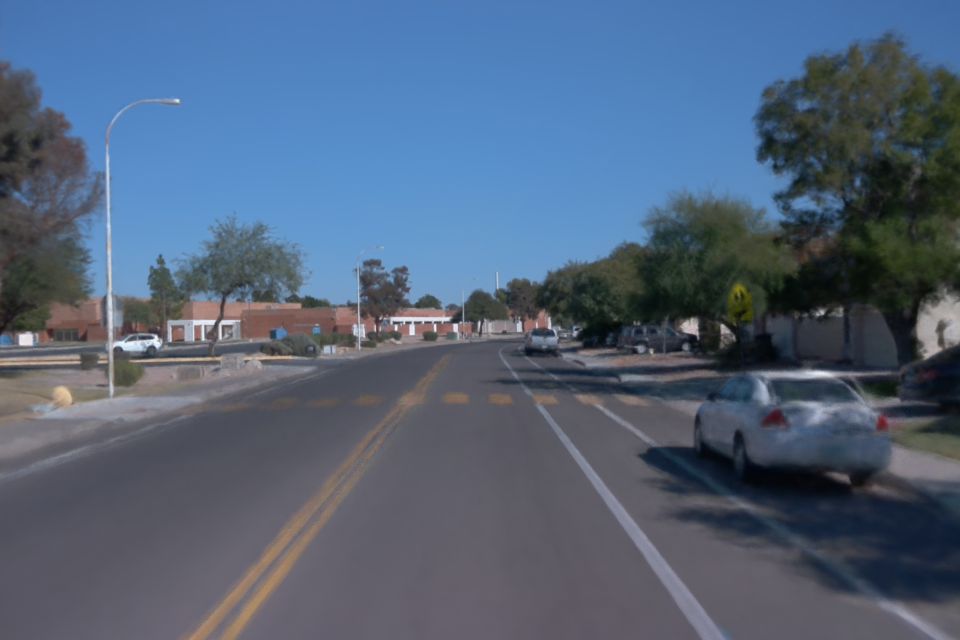} &
    \includegraphics[width=\fgsize\textwidth]{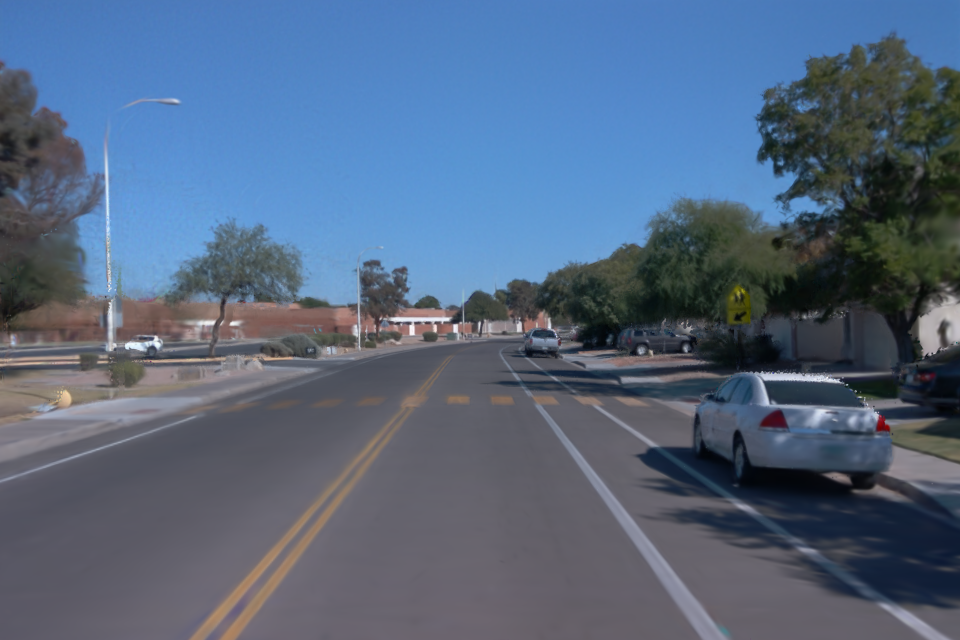} & 
    \includegraphics[width=\fgsize\textwidth]{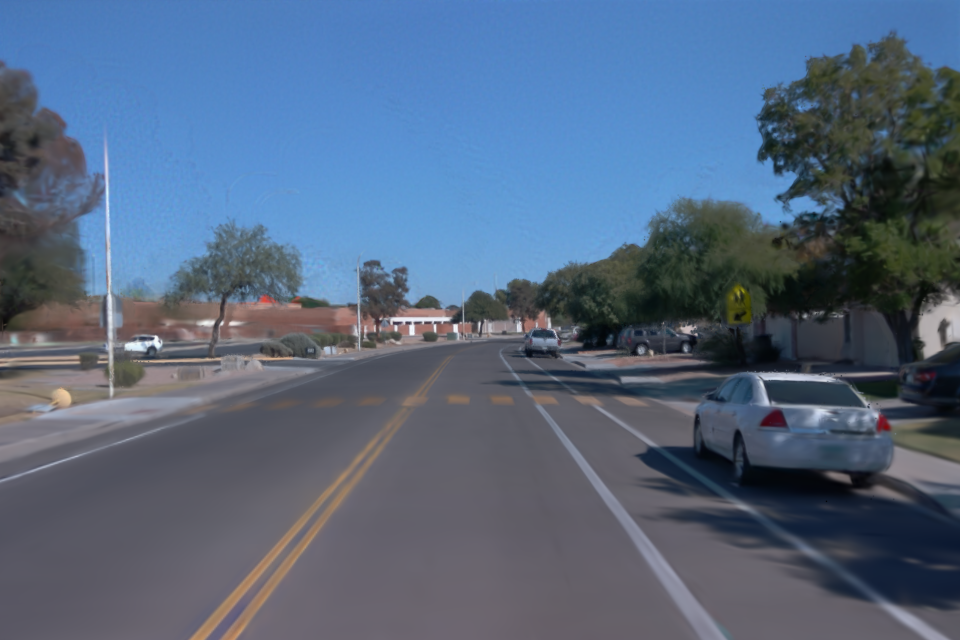} & 
    \includegraphics[width=\fgsize\textwidth]{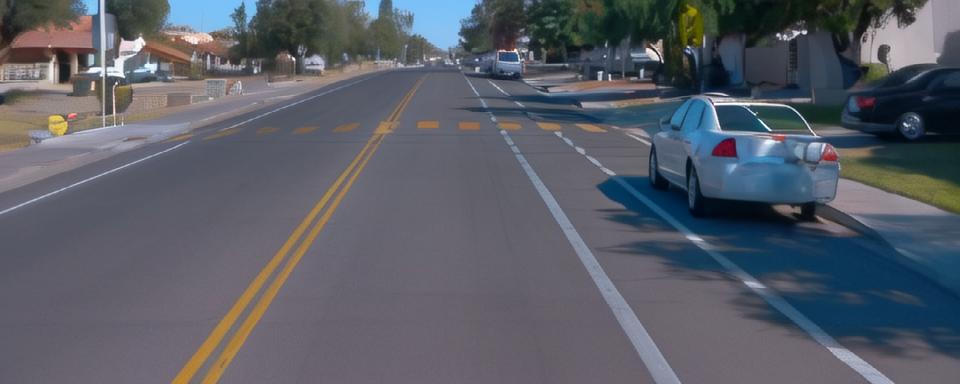} & 
    \includegraphics[width=\fgsize\textwidth]{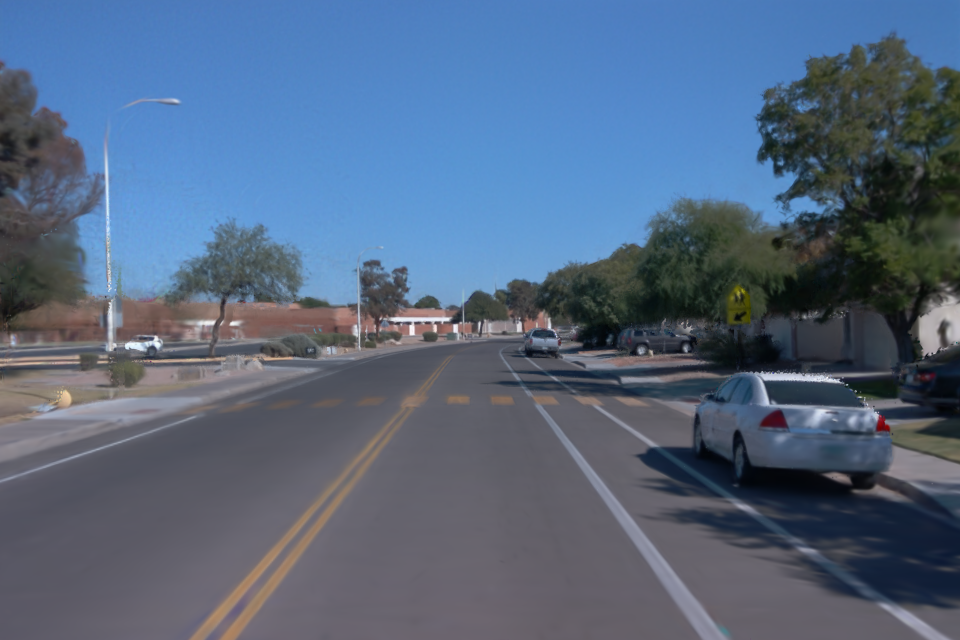}\\
    
    \multirow{2}{*}[13mm]{\rotatebox[origin=c]{90}{Shift 2m}} &
    {} & 
    \includegraphics[width=\fgsize\textwidth]{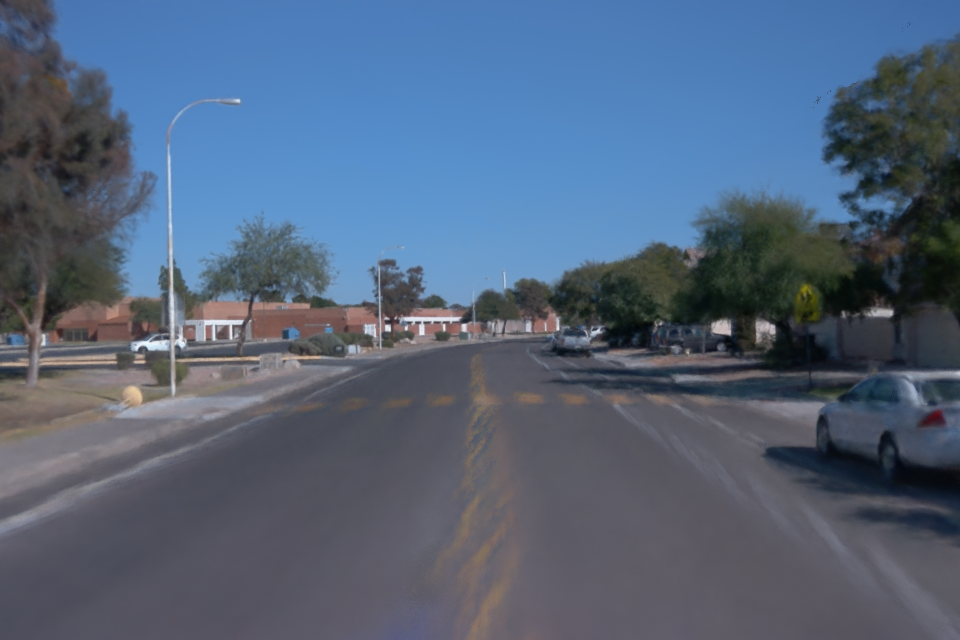} &
    \includegraphics[width=\fgsize\textwidth]{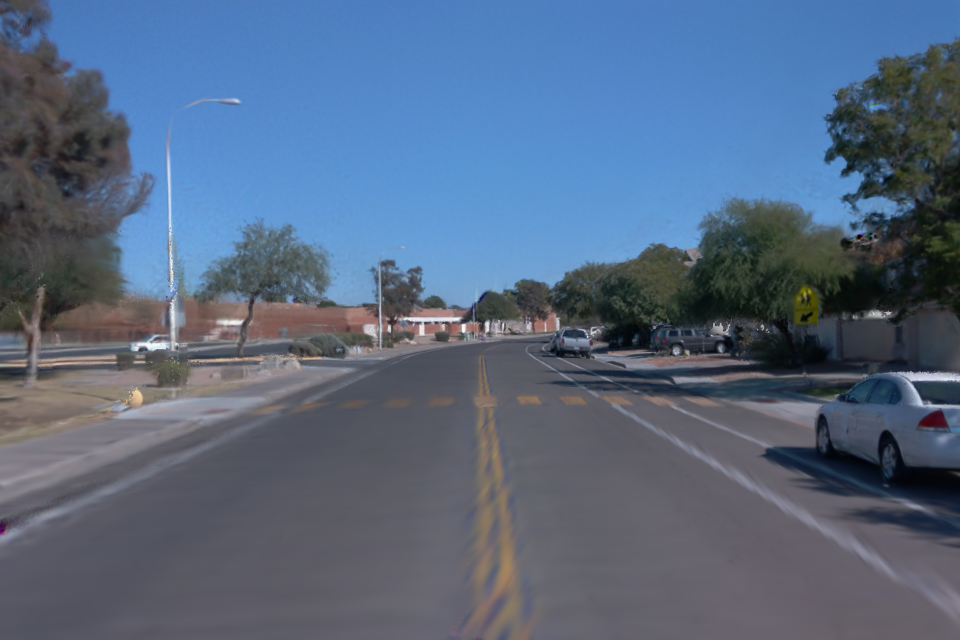} & 
    \includegraphics[width=\fgsize\textwidth]{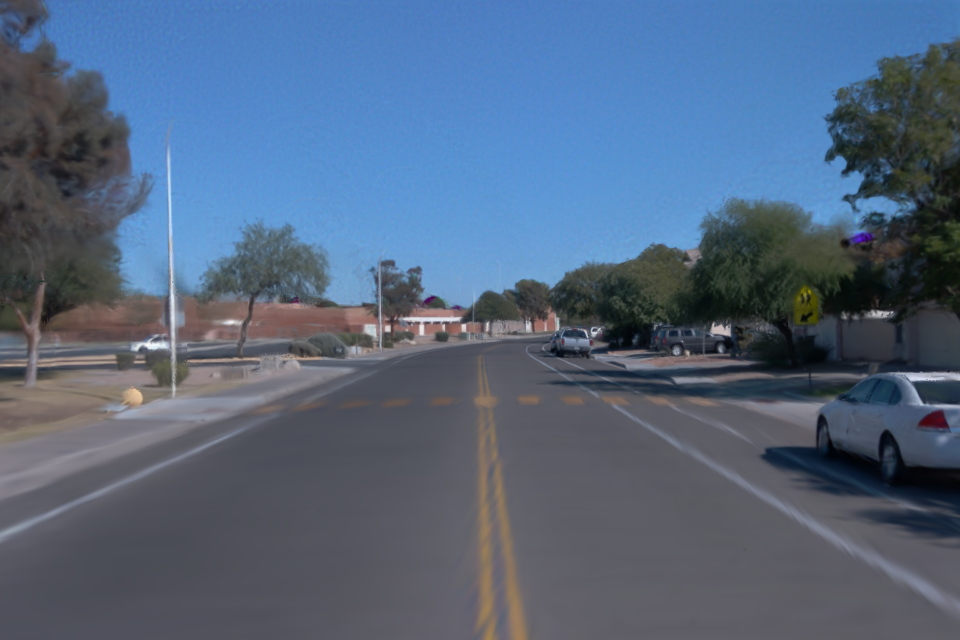} & 
    \includegraphics[width=\fgsize\textwidth]{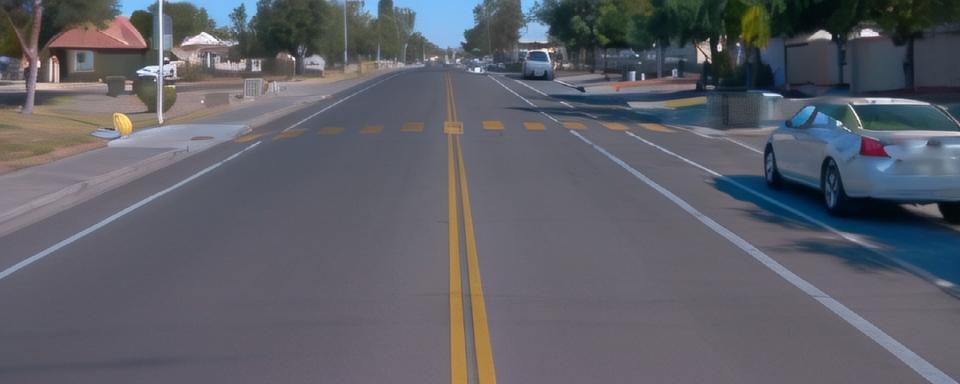} & 
    \includegraphics[width=\fgsize\textwidth]{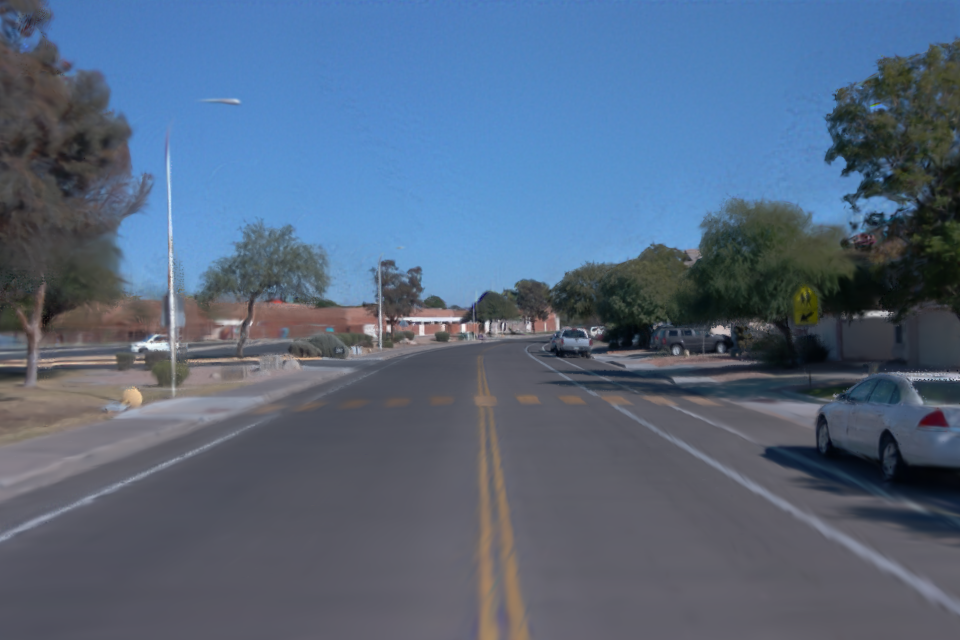}\\
    
    \multirow{2}{*}[13mm]{\rotatebox[origin=c]{90}{Shift 4m}} &
    {} &
    \includegraphics[width=\fgsize\textwidth]{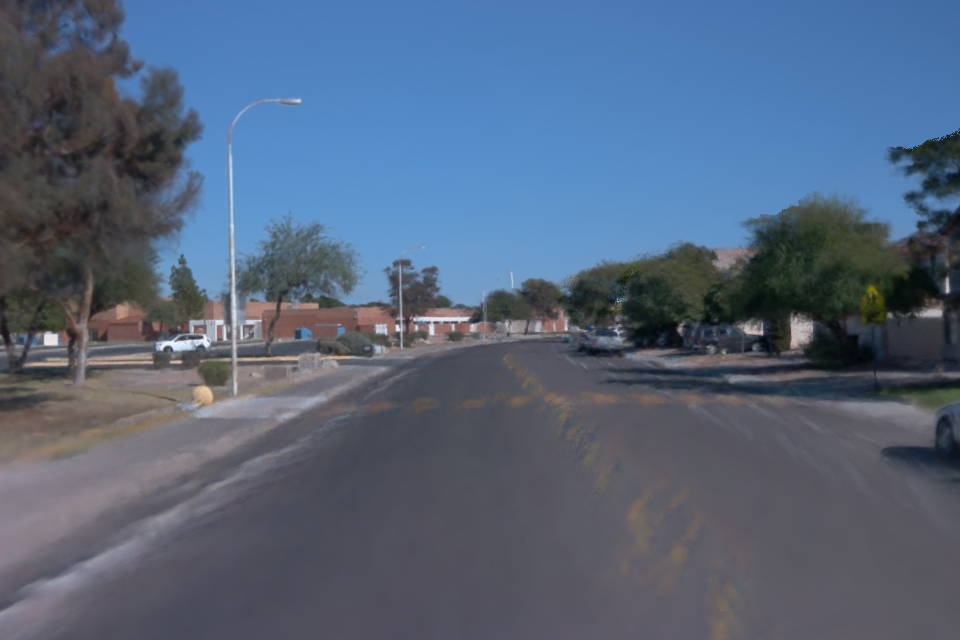} &
    \includegraphics[width=\fgsize\textwidth]{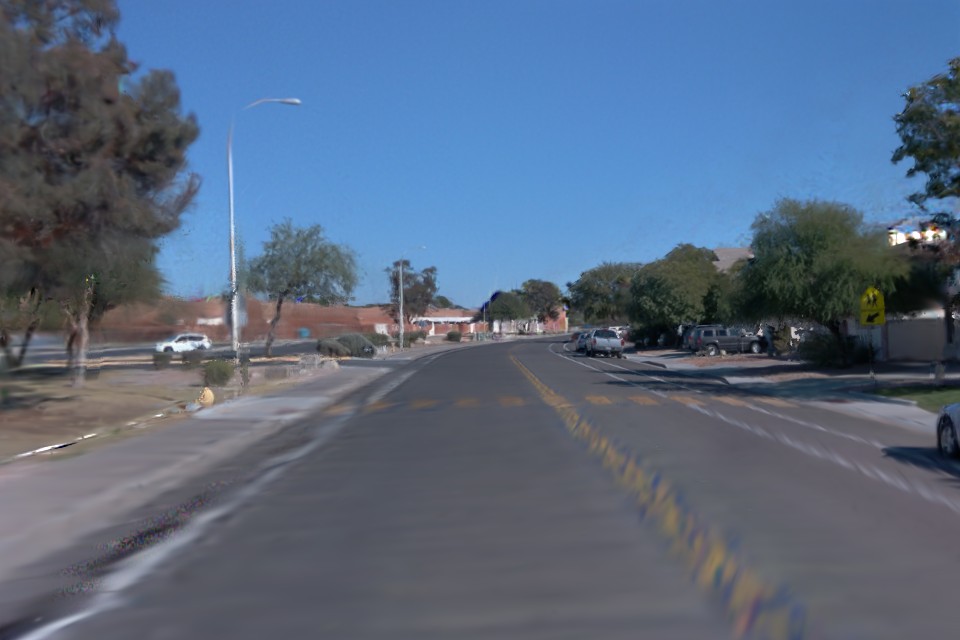} & 
    \includegraphics[width=\fgsize\textwidth]{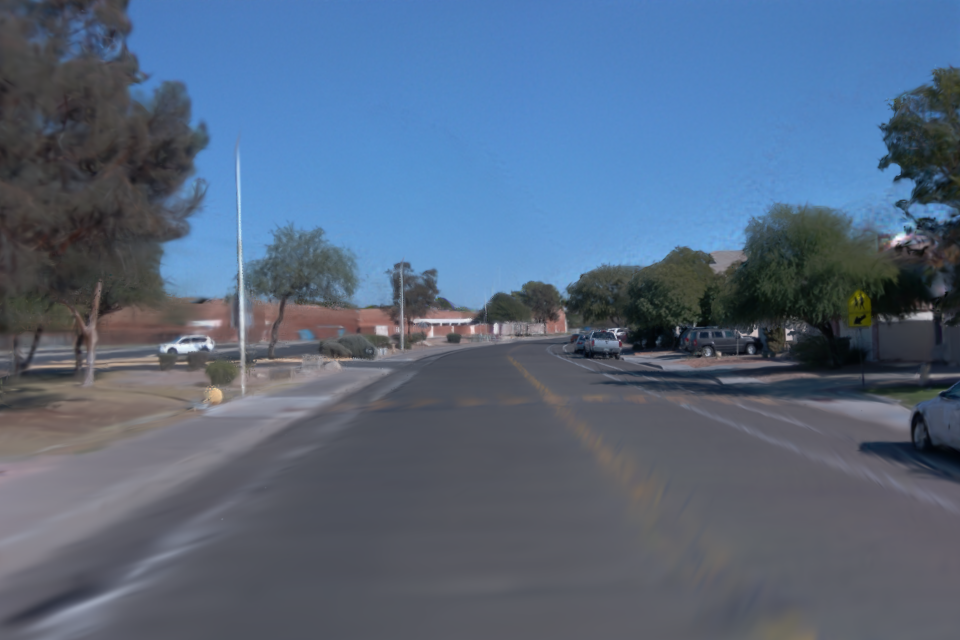} & 
    \includegraphics[width=\fgsize\textwidth]{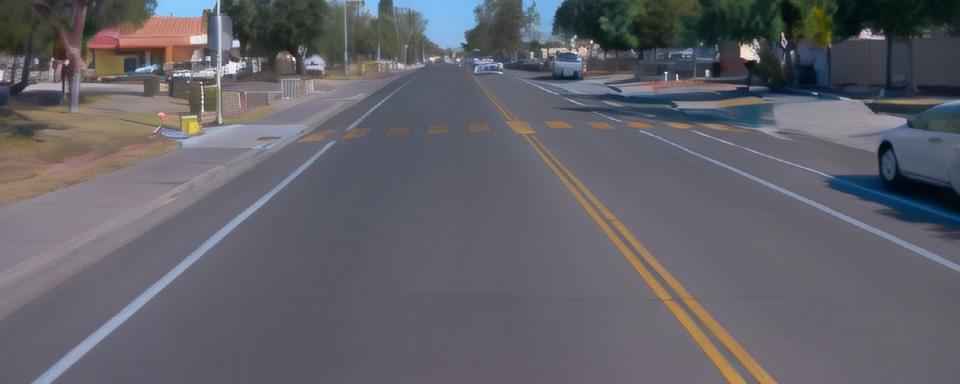} & 
    \includegraphics[width=\fgsize\textwidth]{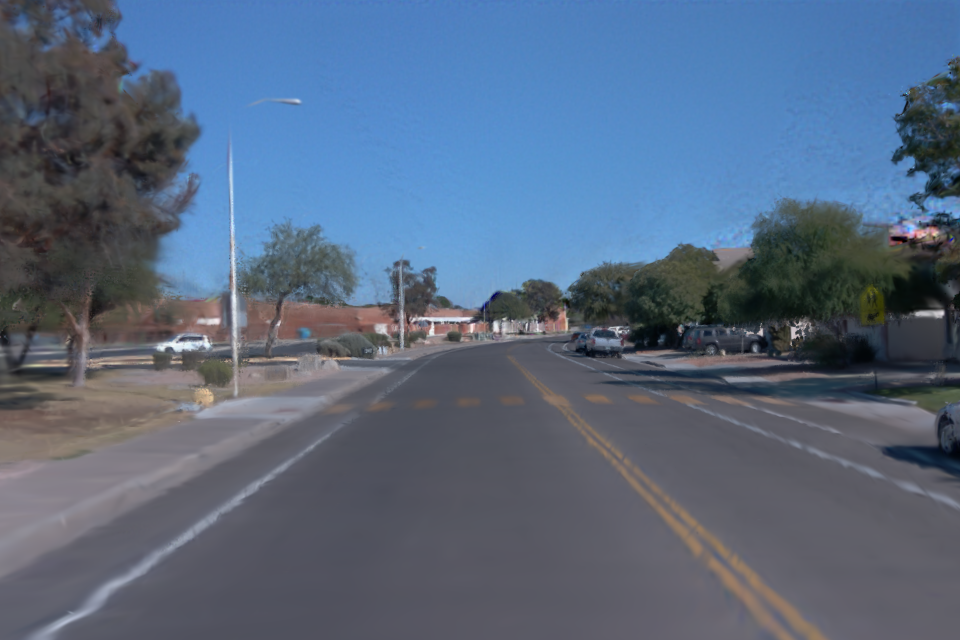}\\

\end{tabular}
\caption{Qualitative comparison on Waymo~\cite{sun2020scalability} with different shifting distances.}
\label{fig:extrapolate_waymo}
\vspace{-0.5cm}
\end{figure*}

\noindent\textbf{3D Gaussian Splatting.}
We first prepare a pre-trained 3DGS on available ground truth images for 20000 iterations. We refer to the original 3DGS paper~\cite{kerbl20233d} for the fundamental training pipeline.
Next, rendered images from augmented viewpoints are encoded and perturbed into noisy latents based on a noise scale $s=\frac{\tau}{T}$. PointmapDiff is used to refine these noisy novel-view images, which are then added to the training data. We choose the reference image with the highest overlap and the closest training LiDAR scan as the condition. We continue fine-tuning the same 3DGS model for another 20000 iterations, in which for every 200 steps, we repeatedly update the augmented set with a linearly reduced $s$.
The scene representation is optimized using two different loss functions:
\begin{align}
    \mathcal{L}_{train}&=\lambda_{rgb}\mathcal{L}_{rgb}+\lambda_{ssim}\mathcal{L}_{ssim}+\lambda_{d}\mathcal{L}_{d},\\
    \mathcal{L}_{aug}&=\lambda_{aug}\mathcal{L}_{rgb}+\lambda_{lpips}\mathcal{L}_{lpips}+\lambda_{d}\mathcal{L}_{d},
\end{align}

\noindent for ground truth training data $\mathcal{L}_{train}$ and augmented data $\mathcal{L}_{aug}$, where $\mathcal{L}_{rgb}$, $\mathcal{L}_{ssim}$ denote the L1 and SSIM losses, respectively. We treat augmented views differently by lowering the weight of L1 loss and incorporating LPIPS loss $\mathcal{L}_{lpips}$ between 3DGS-rendered and diffusion-generated images to prioritize high-level semantic similarity over strict photometric consistency~\cite{gao2024cat3d}. Furthermore, we utilize depth loss $\mathcal{L}_{d}=\|\hat{D}-D_{lidar}\|_{1}$ across all views, as the extrapolated views are designed to be geometrically aligned with the LiDAR data. More details can be found in \cref{sec:supmat_extrapolation}.

\noindent\textbf{Baselines.}
We compare our method with existing state-of-the-art urban driving scene reconstruction and extrapolation methods, including SGD~\cite{yu2025sgd}, VEGS~\cite{hwang2024vegs}, ViewCrafter~\cite{yu2024viewcrafter} and FreeVS~\cite{wang2024freevs} as well as original 3DGS~\cite{kerbl20233d} and Nerfacto~\cite{tancik2023nerfstudio}. For a fair comparison, we allow 3DGS and incorporate accumulated LiDAR data for Gaussian initialization and apply depth supervision to both 3DGS and Nerfacto. Furthermore, we redefine the set of augmented samples for SGD and VEGS to align with our mentioned evaluation criteria.

\noindent\textbf{Metrics.}
We select every $8^{\text{th}}$ frame as conventional test frame for interpolation. We adopt PSNR, SSIM~\cite{wang2004image}, and LPIPS~\cite{zhang2018unreasonable} and evaluate at resolution $1408\times 376$ for KITTI-360 and $960\times 640$ for Waymo.
On the other hand, we report FID~\cite{heusel2017gans} and KID ($\times 100$)~\cite{binkowski2018demystifying} under extrapolation settings since no ground truth image is available. In these cases, KITTI's images are cropped to $600\times 376$, ensuring there is not a lot of unobserved space that could disturb the results, following Hwang~\etal~\cite{hwang2024vegs}.

\begin{table}[htbp]
\centering
\resizebox{\columnwidth}{!}{%
\begin{tabular}{lccccc}
    \toprule
     & \multicolumn{3}{c}{Interpolation} & \multicolumn{2}{c}{Extrapolation}
    \\
    \cmidrule(rl){2-4} 
    \cmidrule(rl){5-6}
     
    & PSNR$\uparrow$ & SSIM$\uparrow$ & LPIPS$\downarrow$ & FID$\downarrow$ & KID$\downarrow$\\
    \midrule
    
    Nerfacto~\cite{tancik2023nerfstudio} & 22.02 & 0.731 & 0.270 & 101.60 & 5.858\\
    3DGS~\cite{kerbl20233d} & \best{23.44} & \best{0.792} & \best{0.122} & 85.08 & 4.926\\
    SGD~\cite{yu2025sgd} & \tbest{23.23} & \tbest{0.779} & 0.162 & \tbest{82.77} & \sbest{4.207}\\
    VEGS~\cite{hwang2024vegs} & 22.77 & \sbest{0.790} & \sbest{0.143} & \sbest{80.79} & \tbest{4.273}\\
    3DGS+PointmapDiff & \sbest{23.39} & \sbest{0.790} & \tbest{0.144} & \best{78.07} & \best{3.799}\\
    \bottomrule
\end{tabular}}
\caption{Quantitative results for interpolated and extrapolated view synthesis on KITTI-360~\cite{liao2022kitti}. Highlighted with \hlc[red!20]{best}, \hlc[orange!20]{second}, and \hlc[yellow!20]{third}.}
\label{tab:extrapolate_kitti}  
\end{table}

\begin{table}[htbp]
\centering
\resizebox{\columnwidth}{!}{%
\begin{tabular}{lccccc}
    \toprule
     & \multicolumn{3}{c}{Interpolation} & \multicolumn{2}{c}{Extrapolation}
    \\
    \cmidrule(rl){2-4} 
    \cmidrule(rl){5-6}
     
    & PSNR$\uparrow$ & SSIM$\uparrow$ & LPIPS$\downarrow$ & FID$\downarrow$ & KID$\downarrow$\\
    \midrule
    Nerfacto~\cite{tancik2023nerfstudio} & 29.75 & 0.884 & 0.189 & 102.79 & 5.650\\
    3DGS~\cite{kerbl20233d} & \sbest{30.18} & \best{0.914} & \best{0.076} & 62.50 & 2.401\\
    3DGS+ViewCrafter~\cite{yu2024viewcrafter} & \tbest{29.83} & \tbest{0.910} & \sbest{0.078} & \tbest{54.16} & \tbest{2.035}\\
    FreeVS~\cite{wang2024freevs} & 23.37 & 0.743 & 0.180 & \sbest{51.95} & \sbest{1.838}\\
    3DGS+PointmapDiff & \best{30.45} & \sbest{0.913} & \tbest{0.079} & \best{48.22} & \best{1.035}\\
    \bottomrule
\end{tabular}}
\caption{Quantitative results for interpolated and extrapolated view synthesis on Waymo~\cite{sun2020scalability}.}
\label{tab:extrapolate_waymo}  
\end{table}

As illustrated in \cref{fig:extrapolate_kitti} and \cref{fig:extrapolate_waymo}, all methods demonstrate improvements over the original 3DGS and NeRF on extrapolated viewpoints. However, while SGD reduces the artifacts, it does not completely resolve the issue, as some blur and traces of artifacts still appear in the generated views. VEGS performs well on planar regions such as roads and buildings, thanks to normal supervision, but struggles with far-away background. ViewCrafter works effectively with small viewpoint shifts but underperforms on larger displacements. FreeVS slightly suffers from color change and is limited to areas where LiDAR data is available, which affects its performance in distant areas and sky regions.
On the other hand, our method portrays clear improvement across all scenarios.
We report quantitative comparison in \cref{tab:extrapolate_kitti} and \cref{tab:extrapolate_waymo}. Combining 3DGS with PointmapDiff outperforms the baselines in inception distances while retaining the most rendering quality of conventional test cameras.

\subsection{Single-image NVS on Street View}

\def\fgsize{0.16}

\begin{figure*}
\centering
\setlength{\tabcolsep}{0.002\linewidth}
\renewcommand{\arraystretch}{0.8}
\begin{tabular}{cccccc}
    Source view & Inpainting~\cite{rombach2022high} & GenWarp~\cite{seo2024genwarp} & BTS~\cite{wimbauer2023behind} & PointmapDiff & Target view (GT)\\

    \includegraphics[width=\fgsize\textwidth]{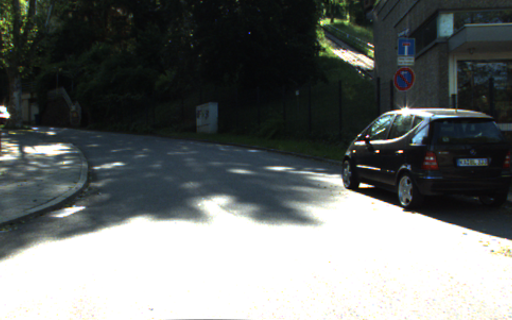} & 
    \includegraphics[width=\fgsize\textwidth]{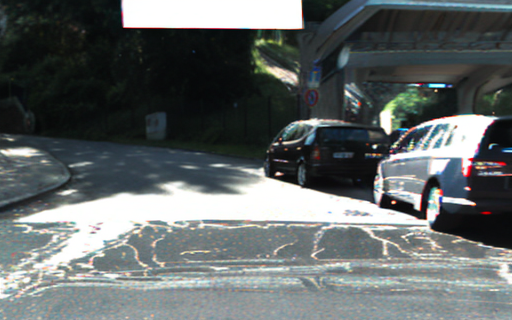} & 
    \includegraphics[width=\fgsize\textwidth]{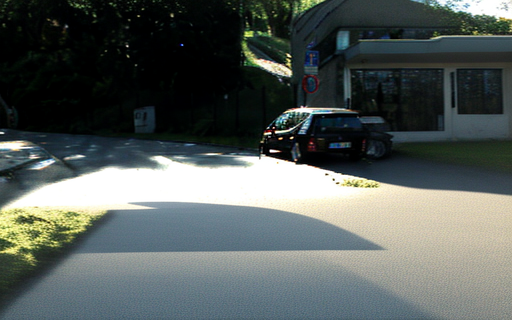} & 
    \includegraphics[width=\fgsize\textwidth]{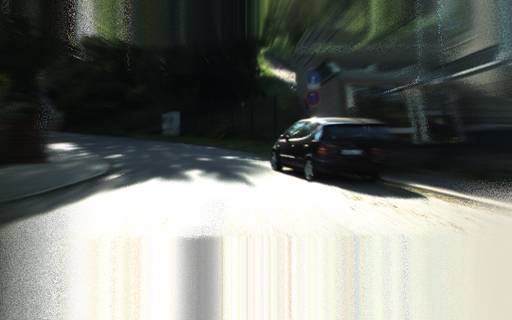} & 
    \includegraphics[width=\fgsize\textwidth]{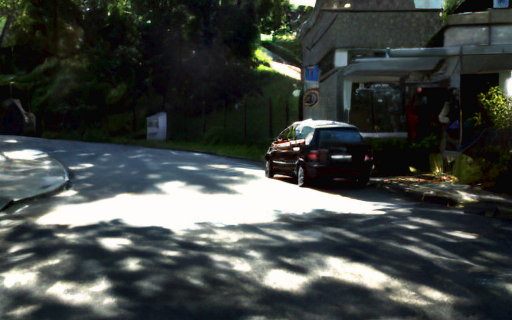}  & 
    \includegraphics[width=\fgsize\textwidth]{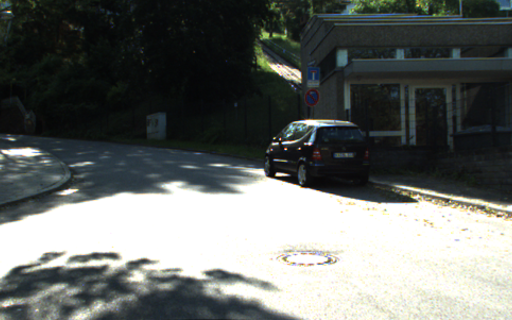}\\

    \includegraphics[width=\fgsize\textwidth]{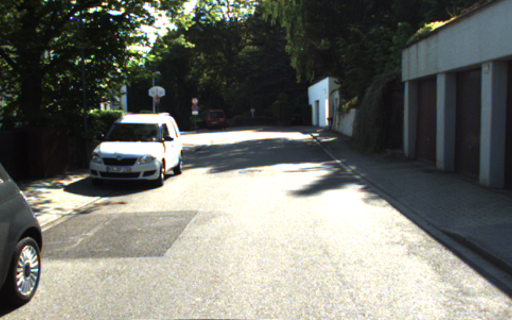} & 
    \includegraphics[width=\fgsize\textwidth]{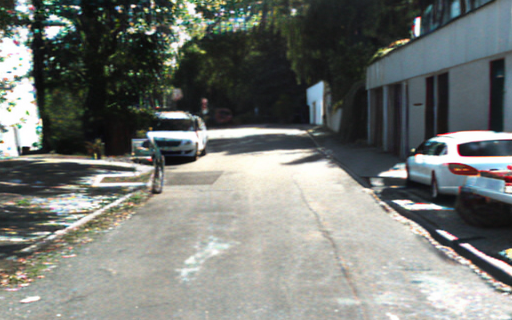} & 
    \includegraphics[width=\fgsize\textwidth]{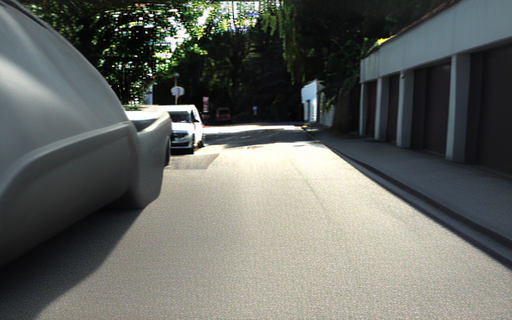} & 
    \includegraphics[width=\fgsize\textwidth]{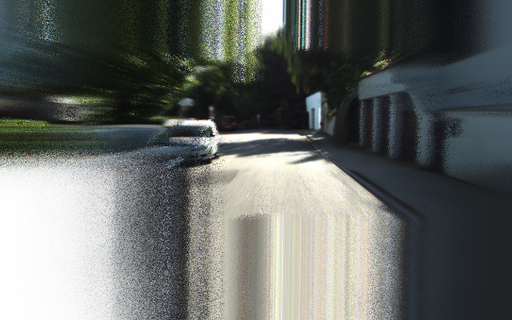} & 
    \includegraphics[width=\fgsize\textwidth]{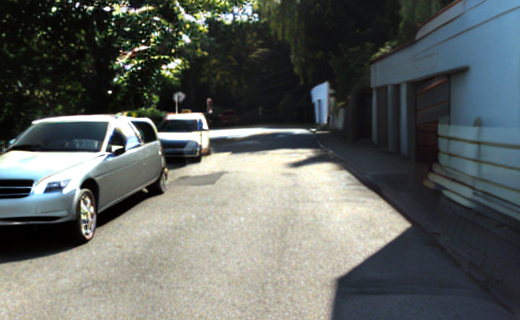} & 
    \includegraphics[width=\fgsize\textwidth]{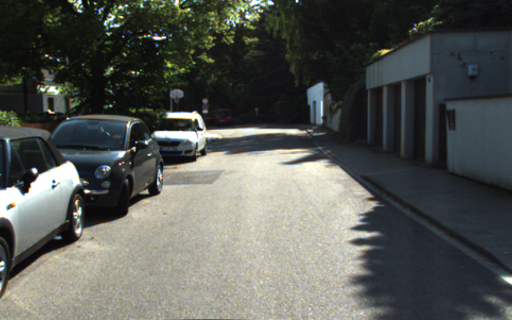}\\

\end{tabular}
\caption{Qualitative comparison for single-image NVS on KITTI-360~\cite{liao2022kitti}.}
\label{fig:single_nvs_kitti}
\vspace{-0.5cm}
\end{figure*}

\noindent\textbf{Setup and Baselines.}
In this experiment, we use a single source image as input to predict target images corresponding to 4-6 previous or following frames in the trajectory.
Our baselines include warping and inpainting method using SD Inpainting~\cite{rombach2022high}, GenWarp~\cite{seo2024genwarp}, and BTS~\cite{wimbauer2023behind} as one-shot implicit NVS. To ensure scene-specific visual fidelity, we fine-tune SD using LoRA~\cite{hu2022lora} on the same dataset. Since the baselines are not designed to handle LiDAR inputs directly, we condition with completed depth maps~\cite{zhang2023completionformer} instead. 
We use FID~\cite{heusel2017gans} and KID ($\times 100$)~\cite{binkowski2018demystifying}  to estimate the realism of the generated image distribution; depth metrics (absolute relative error, root mean square error, and threshold accuracy $\delta < 1.25$) between predicted depth~\cite{hu2024metric3d} of the generated images and LiDAR depth to measure the geometric consistency.
We also include indoor experiments on RealEstate10K~\cite{zhou2018stereo} and ScanNet++~\cite{yeshwanth2023scannet++} with more competitors in the supplementary material.

\begin{table}[htbp]
\centering
\resizebox{\columnwidth}{!}{%
\begin{tabular}{lccccc}
    \toprule
    & FID$\downarrow$ & KID$\downarrow$ & AbsRel$\downarrow$ & RMSE$\downarrow$ & $\delta_1\uparrow$\\
    \midrule
    Inpainting~\cite{rombach2022high} & \sbest{34.61} & \sbest{0.789} & \sbest{0.241} & \sbest{6.496} & \sbest{61.99}\\
    GenWarp~\cite{seo2024genwarp} & \tbest{45.41} & \tbest{2.080} & \tbest{0.270} & \tbest{6.563} & \tbest{56.05}\\
    BTS~\cite{wimbauer2023behind} & 74.66 & 6.120 & 0.351 & 7.131 & 44.33\\
    PointmapDiff & \best{28.31} & \best{0.392} & \best{0.185} & \best{5.030} & \best{73.36}\\
    \midrule
    Lower bound & 11.61 & 0.028 & 0.118 & 3.831 & 87.51\\
    \bottomrule
\end{tabular}}
\caption{Quantitative results for singe-image NVS on KITTI-360~\cite{liao2022kitti}. We include the inception distances between the source and the target view and the predicted depth of the ground truth target view as the lower bound.}
\label{tab:single_nvs_kitti}  
\end{table}

\cref{fig:single_nvs_kitti} and \cref{tab:single_nvs_kitti} show that PointmapDiff demonstrates the ability to follow the scene geometry to effectively generate plausible high-quality views, whereas GenWarp and BTS suffer heavily from stretching artifacts. We also outperform in most metrics.

\subsection{Ablation Study}
\begin{figure}
\centering    
    \begin{subfigure}{0.32\linewidth}
        \includegraphics[clip,trim=0 0 50 0,width=\linewidth]{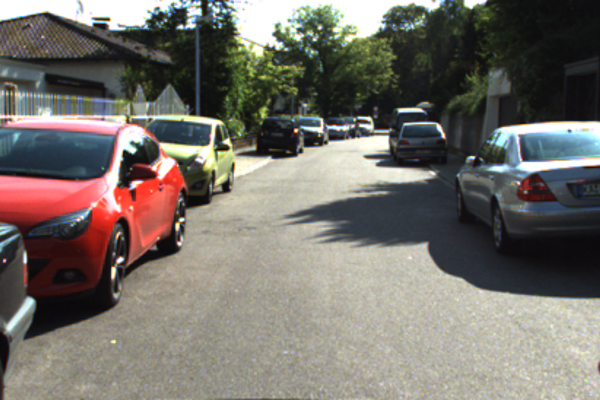}
        \caption{Source view}
    \end{subfigure}
    \begin{subfigure}{0.32\linewidth}
        \includegraphics[clip,trim=0 0 50 0,width=\linewidth]{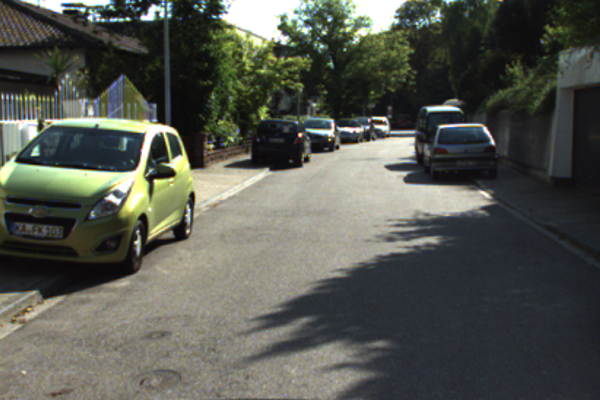}
        \caption{Target view}
    \end{subfigure}
    \begin{subfigure}{0.32\linewidth}
        \includegraphics[clip,trim=0 0 50 0,width=\linewidth]{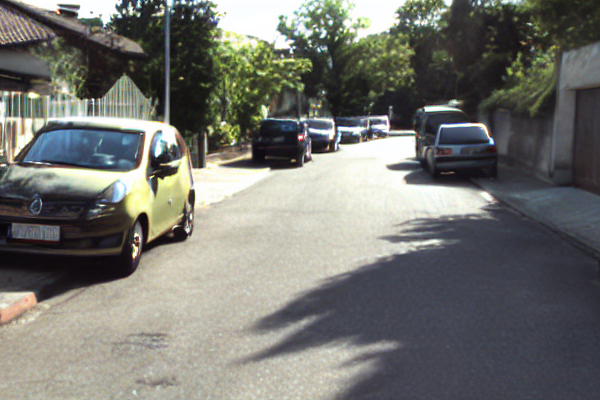}
        \caption{Full model}
        \label{fig:ablation:full_model}
    \end{subfigure}
    \begin{subfigure}{0.32\linewidth}
        \includegraphics[clip,trim=0 0 50 0,width=\linewidth]{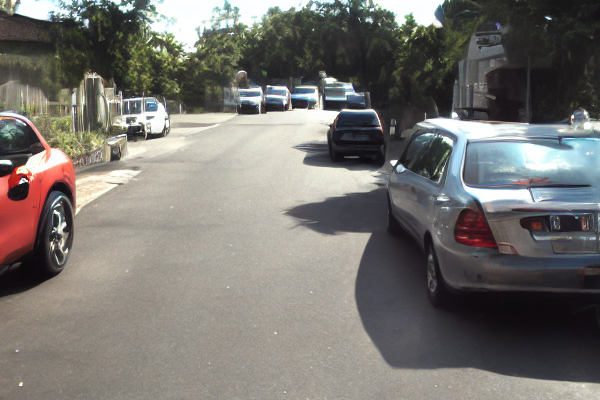}
        \caption{w/o Pointmap CN}
        \label{fig:ablation:no_pointmap}
    \end{subfigure}
    \begin{subfigure}{0.32\linewidth}
        \includegraphics[clip,trim=0 0 50 0,width=\linewidth]{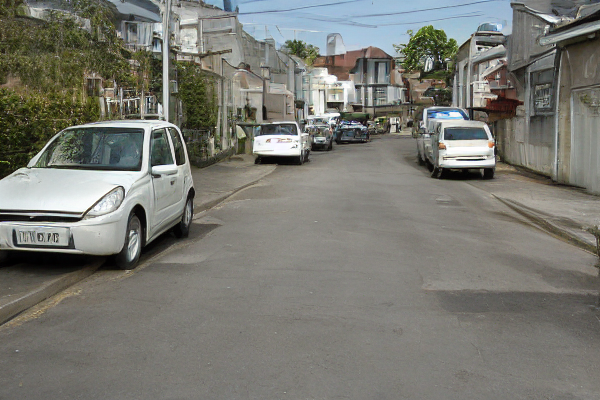}
        \caption{w/o Reference Attn.}
        \label{fig:ablation:no_crossattn}
    \end{subfigure}
    \begin{subfigure}{0.32\linewidth}
        \includegraphics[clip,trim=0 0 50 0,width=\linewidth]{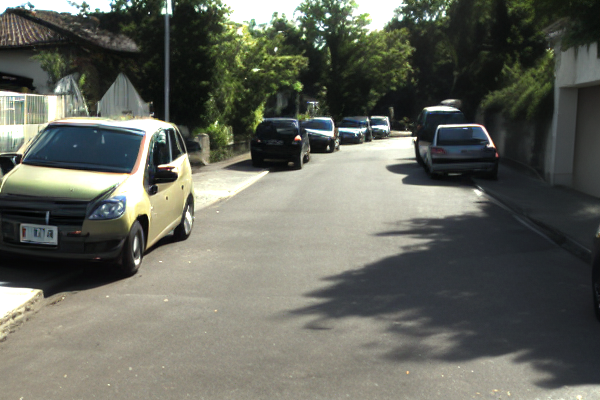}
        \caption{w/o Pointmap P.E.}
        \label{fig:ablation:no_pe}
    \end{subfigure}
    \caption{The full model effectively captures high-detail scene continuity, closely aligning with the target image, while removing components leads to a loss in both geometric structure or appearance fidelity.}
    \label{fig:ablation}
    \vspace{-0.5cm}
\end{figure}

To study the design of PointmapDiff, we conduct single-view NVS experiment on all the model variants. \cref{tab:ablation} and \cref{fig:ablation} demonstrate the results of our study.

\begin{table}[htbp]
    \centering
    \resizebox{\columnwidth}{!}{%
    \begin{tabular}{lccccc}
        \toprule
        Ablation & FID$\downarrow$ & KID$\downarrow$ & AbsRel$\downarrow$ & RMSE$\downarrow$ & $\delta_1\uparrow$\\
        \midrule
        w/o ControlNet & \tbest{39.50} & \tbest{0.936} & 0.422 & 10.937 & 39.70\\
        w/o Attention & 64.57 & 3.626 & \tbest{0.272} & \tbest{6.899} & \tbest{56.61}\\
        w/o Pointmap P.E. & \sbest{24.80} & \sbest{0.276} & \sbest{0.184} & \sbest{5.009} & \sbest{72.48}\\
        \midrule
        Full model & \best{21.26} & \best{0.268} & \best{0.181} & \best{4.880} & \best{74.46}\\
        \bottomrule
     \end{tabular}}
    \caption{Quantitative ablation of individual components.}
    \label{tab:ablation}
\end{table}

\noindent\textbf{Pointmap ControlNets.} When excluding the Pointmap ControlNets, the model loses access to the correct correspondences derived from the reference image. This omission impairs its ability to maintain spatial consistency, resulting in generated views that respect the reference appearance but deviate significantly in geometry (\cref{fig:ablation:no_pointmap}).

\noindent\textbf{Reference-Guided Cross-View Attention.} 
Without reference attention, the model operates similarly to a standard geometry-controlled SD model. Even when we employ LLaVA~\cite{liu2024improved} to input more detailed scene descriptions, the model struggles to respect the contents. However, this version provides valuable insights; specifically, it demonstrates that point maps are an effective conditioning source. They successfully encode the scene's geometry, helping the model recover the scene's structure reasonably, only without precise adherence to the reference image. In \cref{fig:ablation:no_crossattn}, while the model can place the cars in the correct position, it fails to transfer the appearance from the source image accurately.

\noindent\textbf{Pointmap Positional Encoding (P.E.).}
\cref{fig:ablation:no_pe} shows that directly passing point map coordinates into the network results in reduced image detail (\eg, texture on the road and the shadow regions) and lower metrics, whereas preprocessing the input with positional embedding (\cref{fig:ablation:full_model}) enables the model to represent higher frequency details.

\subsection{Discussion}
In this section, we explore additional capabilities granted by the architecture design of PointmapDiff, along with its current limitations.

\noindent\textbf{LiDAR-algined Generation.}
PointmapDiff demonstrates a remarkable ability to generate images that align closely with LiDAR-derived conditions, preserving spatial structure and depth cues inherent in point cloud data. This is critical for autonomous driving applications, where hallucinated regions must remain consistent with real-world geometry and context, rather than being entirely random. Such fidelity ensures that generated content supports safe and reliable perception in driving environments. We refer to the \cref{sec:lidar_align} for more illustrations.

\begin{figure}
\centering    
\begin{subfigure}{0.24\linewidth}
    \includegraphics[clip,trim=50 0 0 0,width=\linewidth]{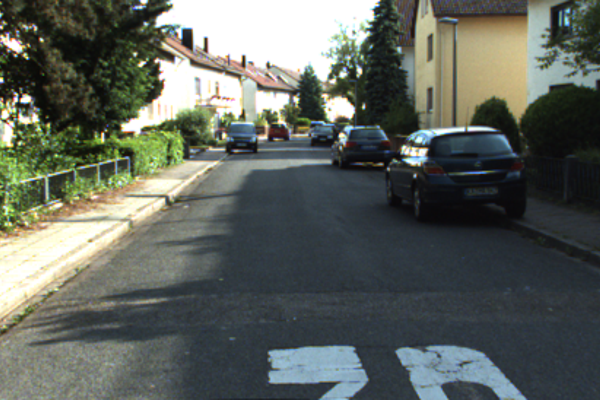}
    \caption*{Source view}
\end{subfigure}
\begin{subfigure}{0.24\linewidth}
    \includegraphics[clip,trim=50 0 0 0,width=\linewidth]{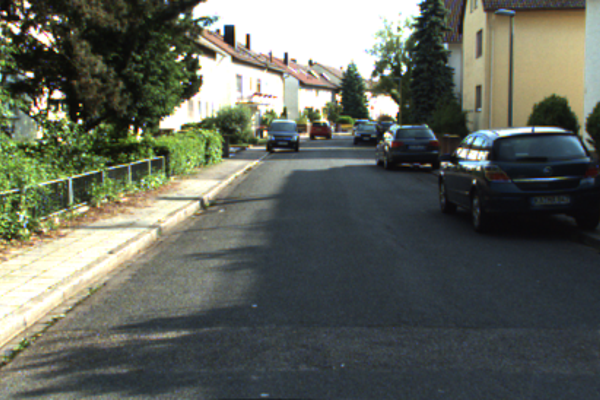}
    \caption*{Target view}
\end{subfigure}
\begin{subfigure}{0.24\linewidth}
    \includegraphics[clip,trim=50 0 0 0,width=\linewidth]{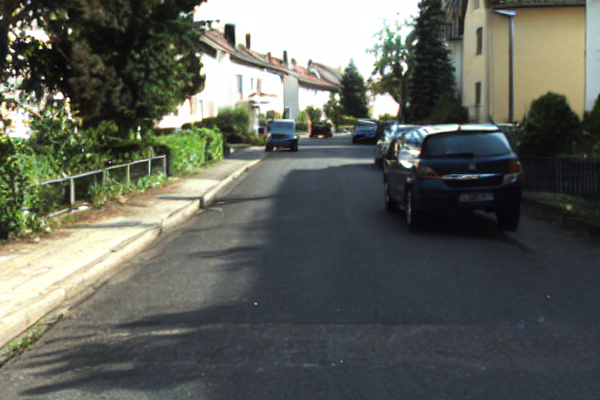}
    \caption*{Translation}
\end{subfigure}
\begin{subfigure}{0.24\linewidth}
    \includegraphics[clip,trim=50 0 0 0,width=\linewidth]{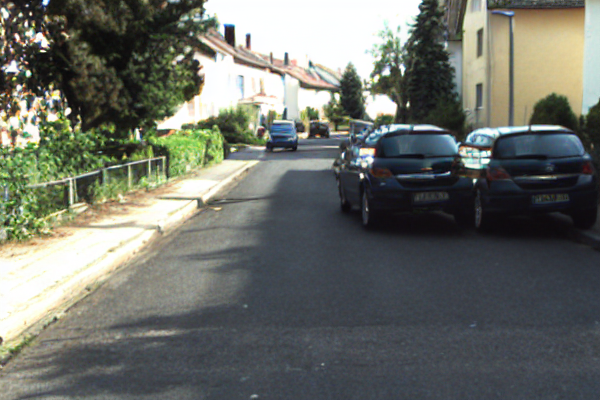}
    \caption*{Duplication}
\end{subfigure}
\caption{\textbf{Scene editing results.} We perform instance-level editing by manipulating point map values.}
\label{fig:scene_editting}
\vspace{-0.5cm}
\end{figure}

\noindent\textbf{Scene Editing.}
Our model allows image editing by manipulating the point map, which enables repositioning or duplicating objects within a scene without changing their visual appearances. 
We first isolate the points belonging to the objects of interest using 3D bounding boxes or instance labels. Then, spatial transformations are applied to these points while keeping their initial values in the point map. This helps the model to establish correspondences based on these transformations. 
Following this idea, we showcase in \cref{fig:scene_editting} two scenarios where we shift and duplicate a set of points that belong to a car. As a result, we can perform both novel view synthesis and spatial editing at the same time.
This provides a promising direction for future explorations in scene manipulation through point map-based editing.

\noindent\textbf{Limitations.} 
While PointmapDiff improves extrapolated views, the model can still introduce inconsistencies that cause blurriness, thus lower reconstruction and interpolation qualities. Additionally, the current design is optimized for static scenes and may struggle with dynamic environments. Future work will focus on extending PointmapDiff to better handle dynamic scenarios, incorporating temporal consistency and object tracking to improve generation quality in more complex settings.

\section{Conclusion}
We present a novel approach for view synthesis by integrating point map-conditioned diffusion, significantly improving scene reconstruction in extrapolating use cases. Our method enhances spatial consistency and enables high-quality novel view generation. Detailed evaluations confirm its robustness and effectiveness in urban driving data and open new editable view synthesis applications.

{
    \small
    \bibliographystyle{ieeenat_fullname}
    \bibliography{main}
}

\clearpage
\maketitlesupplementary
\appendix

\section{Implementation Details}
\label{sec:implementation}
\subsection{Design Motivation}
We further explain the motivation for using point maps as conditioning signals.
Given reference $r$ and target $t$ viewpoints, establishing correspondences $\mathcal{M}^{r,t}$ between pixels of two images can be trivially achieved by nearest neighbor (NN) search in the 3D point map space:
\begin{multline}
    \mathcal{M}^{r,t}=\left\{(a,b)~|~a=\text{NN}^{t,r}(b)~\text{and}~b=\text{NN}^{r,t}(a)\right\},
    \\
    \text{with}~\text{NN}^{m,n}(a)=\underset{b\in\{0,...,WH\}}{\arg\min}\left\|X^{n,n}_{b}-X^{m,n}_{a}\right\|.
\end{multline}

Here, $\text{NN}^{m,n}$ computes the nearest neighbor $b$ of pixel $a$ between views $m$ and $n$. While this explicit correspondence is computationally expensive and only operates on pixel space, it motivates our approach of leveraging implicit attention mechanisms.

\begin{figure}
\centering    
\begin{subfigure}{0.32\linewidth}
    \begin{tikzpicture}
    \node[anchor=south west,inner sep=0] (image) at (0,0) {\includegraphics[width=\linewidth]{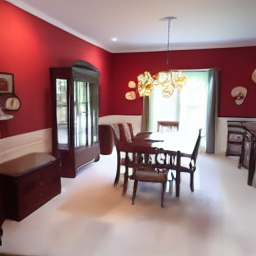}};
    \begin{scope}[x={(image.south east)},y={(image.north west)}]
        \draw[black,fill=green] (0.33,0.41) circle (0.04);
    \end{scope}
    \end{tikzpicture}
    \caption*{Predicted view}
\end{subfigure}
\begin{subfigure}{0.653\linewidth}
    \includegraphics[width=0.49\linewidth]{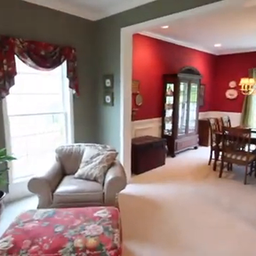}
    \includegraphics[width=0.49\linewidth]{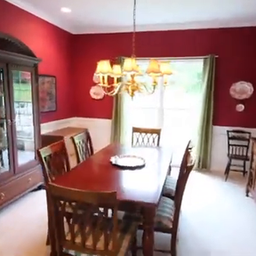}
    \caption*{Source views}
\end{subfigure}
\begin{subfigure}{0.32\linewidth}
    \includegraphics[width=\linewidth]{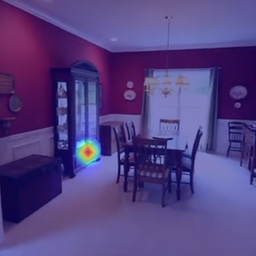}
    \caption*{Self-attention map}
\end{subfigure}
\begin{subfigure}{0.653\linewidth}
    \includegraphics[width=0.49\linewidth]{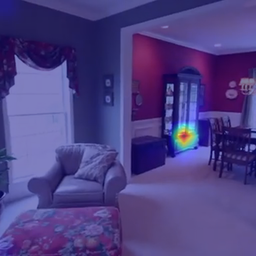}
    \includegraphics[width=0.49\linewidth]{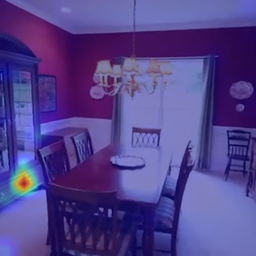}
    \caption*{Cross-attention maps}
\end{subfigure}
\caption{Given \textcolor{teal}{a query point} in the upper-left generated view and reference views, we extract PointmapDiff's intermediate layer activations through the keys and queries of self-attention and reference attention layers at a certain time step $\tau=0.2T$ during the denoising process and use them to visualize the attention maps~\cite{tang2023emergent, alaluf2024cross}. As a result, the method can learn and produce correct correspondences.}
\label{fig:attn_vis}
\end{figure}

We consider a simple positional encoding example of point maps, $\gamma(X)$, which maps the normalized input points to higher-dimensional Fourier features using a set of sine and cosine functions:
\begin{multline}
    \gamma(\mathbf{x})=[a_{1}\cos(2\pi F_1\mathbf{x}), a_{1}\sin(2\pi F_1\mathbf{x}),\dots,\\
    a_{N}\cos(2\pi F_N\mathbf{x}), a_{N}\sin(2\pi F_N\mathbf{x})]^{T},
\end{multline}
where $F_j$ are the Fourier basis frequencies and $a_j$ are their corresponding coefficients. Using this encoding, the spatial correlation between two point maps can be measured via a kernel function as:
\begin{equation}
    \gamma(\mathbf{x}_{1})\gamma(\mathbf{x}_2)^{T}= \sum_{j=1}^{N}a_{j}^{2}\cos\left(2\pi F_{j}(\mathbf{x}_{1}-\mathbf{x}_{2})\right).
\end{equation}
To adapt this to the nearest neighbor computation, we redefine $\text{NN}^{m,n}$ using the encoded point maps as follows:
\begin{multline}
    \text{NN}^{m,n}(a)=\underset{b\in\{0,...,WH\}}{\arg\max}\left(\gamma\left(X^{n,n}_{b}\right)\gamma\left(X^{m,n}_{a}\right)^{T}\right),
\end{multline}
by replacing $t \gets n$ and $r \gets m$, and applying this for all $a \in \{0, \dots, WH\}$, interestingly, this operation resembles the reference attention mechanism introduced in the main paper. Specifically, the attention matrix: $A = \text{softmax}\left(\frac{Q^{t}{K^{r}}^{T}}{\sqrt{d}}\right)$ serves a similar purpose by learning implicit correspondences between the query ($Q^t$) and key ($K^r$) representations extracted from Pointmap ControlNet's layers of the target and reference views. Thus, the point map conditioning acts as an intermediate signal to naturally establish correspondences within the attention layers, bridging the gap between explicit point matching and feature-based reasoning with the ability to dynamically attend to relevant regions.
Hence, we verify the roles of the keys and queries in \cref{fig:attn_vis}; they determine the regions in the source views that can be used for generation.

\subsection{Training}
Our method employs a pre-trained SD v1.5 as the backbone, thanks to its robust generative capabilities. Since SD v1.5 is also a text-to-image model, we incorporate simple text prompts, such as \textit{"a photo of a driving scene"} or \textit{"a photo of a room"}, to provide high-level semantic guidance.

\begin{figure*}
    \centering
    \includegraphics[width=1\linewidth]{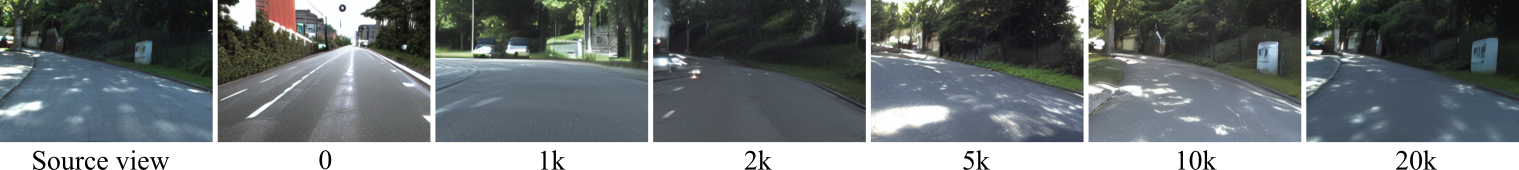}
    \caption{Validation sample observed in several training iterations.}
    \label{fig:convergence}
\end{figure*}

Unlike other methods~\cite{gao2024cat3d, seo2024genwarp}, we do not modify the latent input, allowing us to retain the U-Net backbone and instead adapt to the task by training the additional ControlNet.
For the positional encoding, we use a frequency range from $2^0$ to $2^3$, resulting in an input channel dimension of 24 for the ControlNet model. 
As the training progresses, we observe sudden convergence of ControlNet after approximately 10K iterations, portrayed in \cref{fig:convergence}. The model continues to refine fine-grained details beyond this point, yielding steady improvements in PSNR and SSIM.

\def\fgsize{0.16}

\begin{figure*}
\centering
\setlength{\tabcolsep}{0.002\linewidth}
\renewcommand{\arraystretch}{0.8}
\begin{tabular}{lcccccc}
    {} & Reference & NeRF~\cite{tancik2023nerfstudio} & 3DGS~\cite{kerbl20233d} & SGD~\cite{yu2025sgd} & VEGS~\cite{hwang2024vegs} & 3DGS+PointmapDiff\\
    
    \multirow{2}{*}[13mm]{\rotatebox[origin=c]{90}{Rotate $45^\circ$}} & 
    \includegraphics[clip=true,trim={0 0 0 0},width=\fgsize\textwidth]{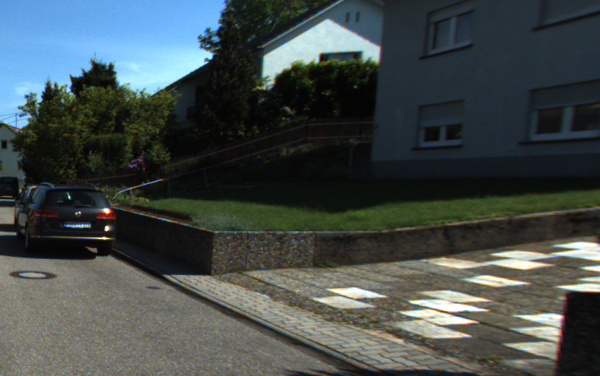} &
    \includegraphics[clip=true,trim={0 0 0 0},width=\fgsize\textwidth]{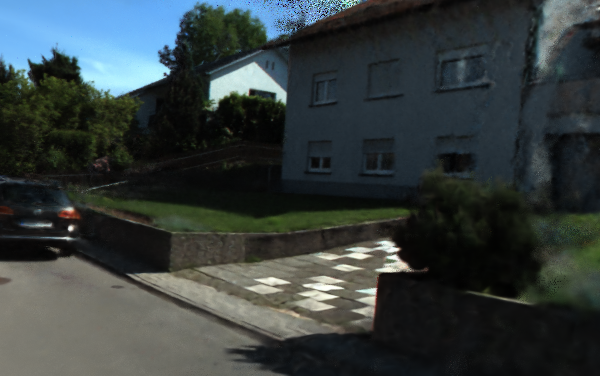} &
    \includegraphics[clip=true,trim={0 0 0 0},width=\fgsize\textwidth]{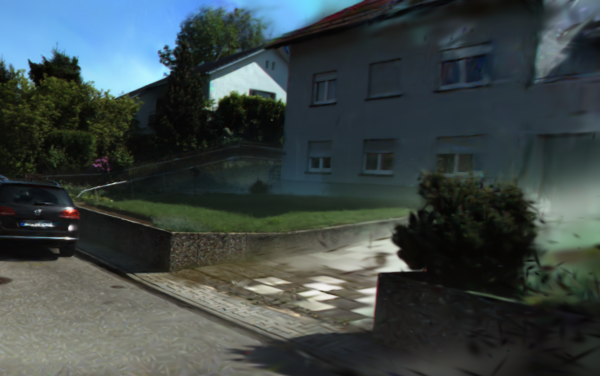} & 
    \includegraphics[clip=true,trim={0 0 0 0},width=\fgsize\textwidth]{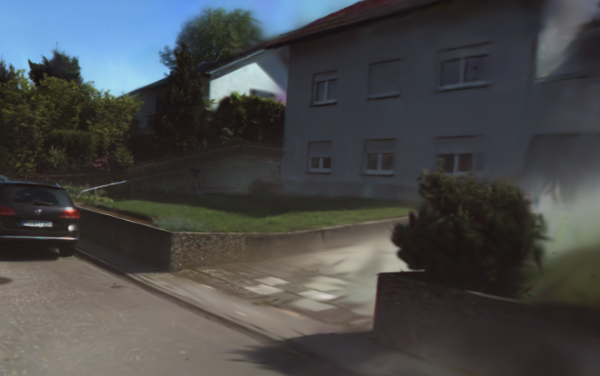} & 
    \includegraphics[clip=true,trim={0 0 0 0},width=\fgsize\textwidth]{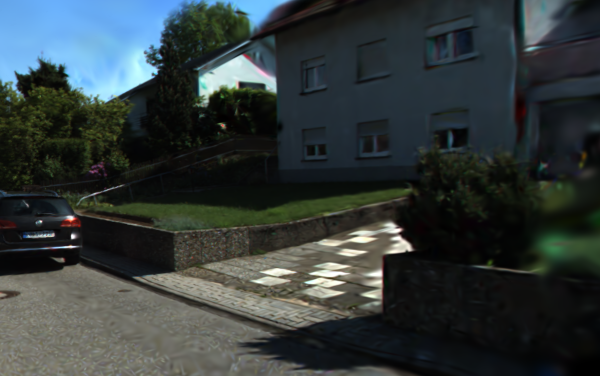} & 
    \includegraphics[clip=true,trim={0 0 0 0},width=\fgsize\textwidth]{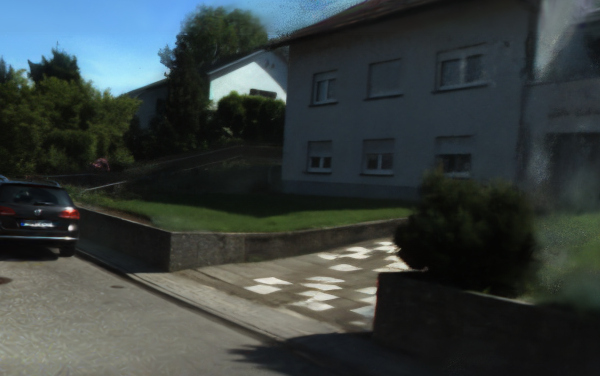}\\
    
    \multirow{2}{*}[12mm]{\rotatebox[origin=c]{90}{Shift 2m}} &
    \includegraphics[clip=true,trim={0 0 0 0},width=\fgsize\textwidth]{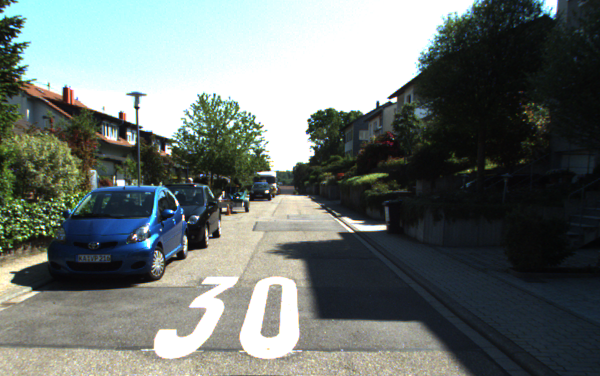} &
    \includegraphics[clip=true,trim={0 0 0 0},width=\fgsize\textwidth]{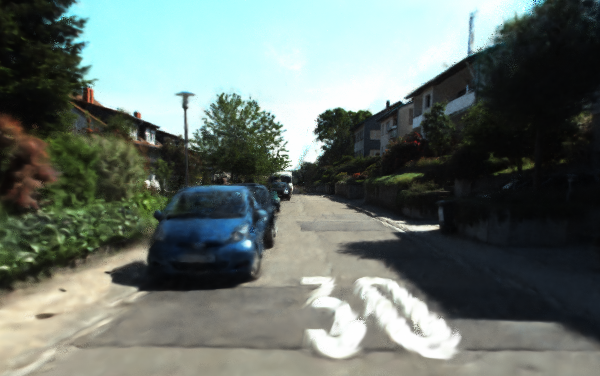} &
    \includegraphics[clip=true,trim={0 0 0 0},width=\fgsize\textwidth]{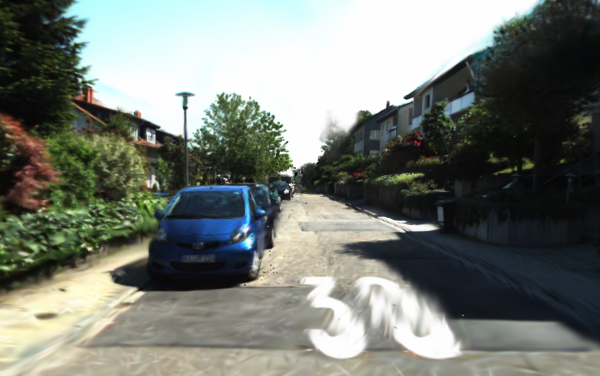} & 
    \includegraphics[clip=true,trim={0 0 0 0},width=\fgsize\textwidth]{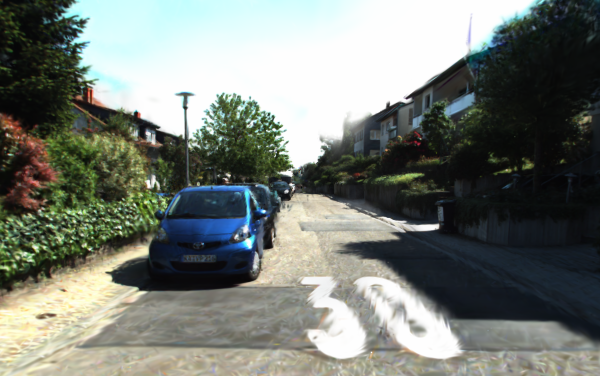} & 
    \includegraphics[clip=true,trim={0 0 0 0},width=\fgsize\textwidth]{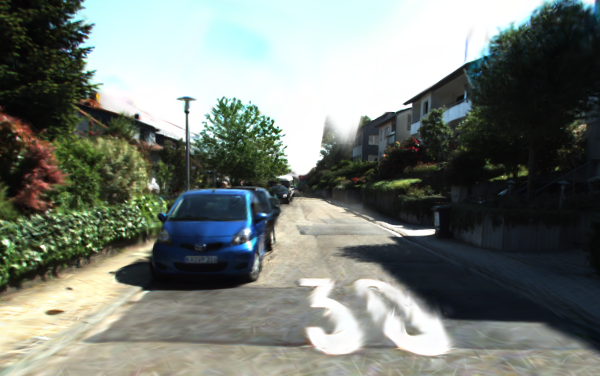} & 
    \includegraphics[clip=true,trim={0 0 0 0},width=\fgsize\textwidth]{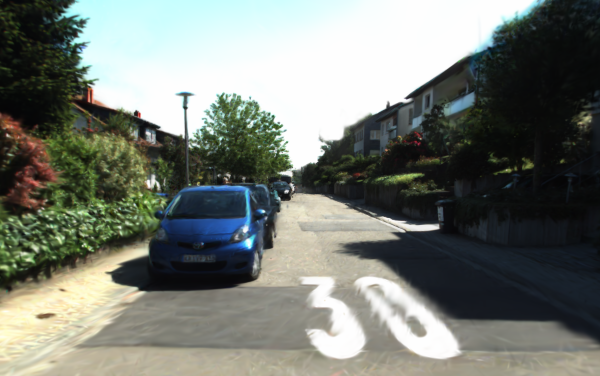}\\
    
    \multirow{2}{*}[15mm]{\rotatebox[origin=c]{90}{Upward 1m}} &
    \includegraphics[clip=true,trim={0 0 0 0},width=\fgsize\textwidth]{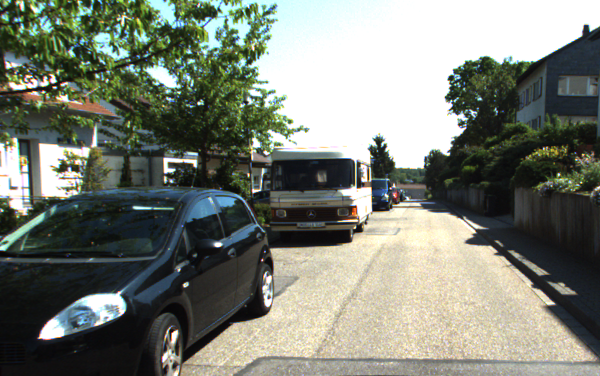} & 
    \includegraphics[clip=true,trim={0 0 0 0},width=\fgsize\textwidth]{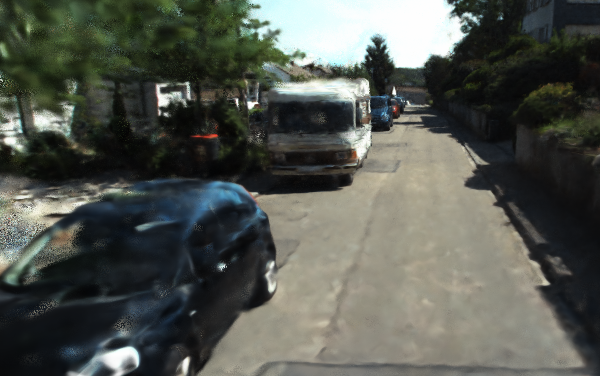} &
    \includegraphics[clip=true,trim={0 0 0 0},width=\fgsize\textwidth]{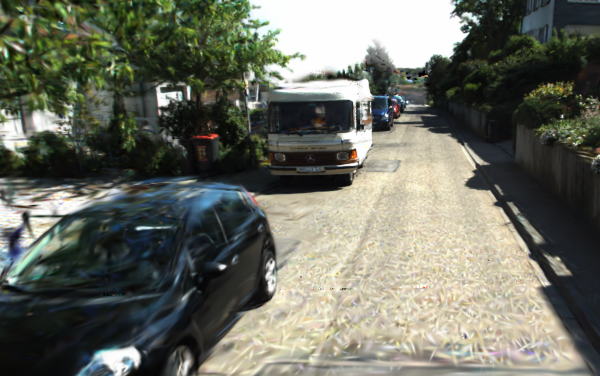} & 
    \includegraphics[clip=true,trim={0 0 0 0},width=\fgsize\textwidth]{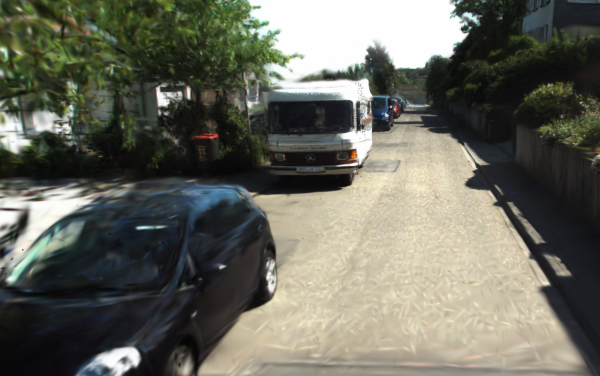} & 
    \includegraphics[clip=true,trim={0 0 0 0},width=\fgsize\textwidth]{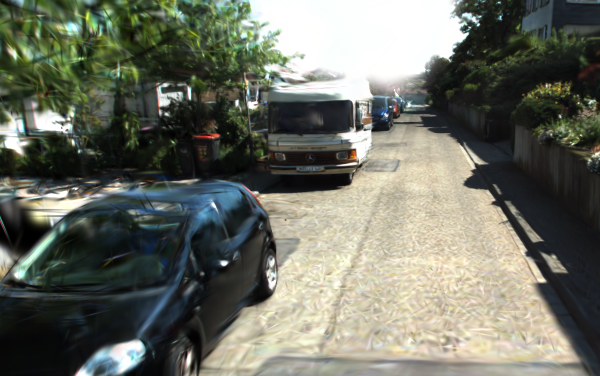} & 
    \includegraphics[clip=true,trim={0 0 0 0},width=\fgsize\textwidth]{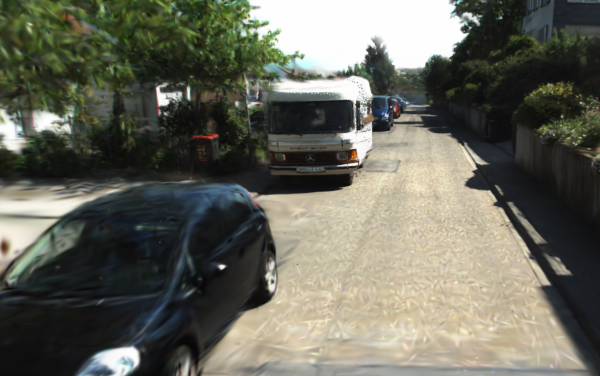}\\

\end{tabular}
\caption{Additional qualitative comparison on KITTI-360~\cite{liao2022kitti}.}
\label{fig:extrapolate_kitti_supmat}
\end{figure*} 

\def\fgsize{0.16}

\begin{figure*}
\centering
\setlength{\tabcolsep}{0.002\linewidth}
\renewcommand{\arraystretch}{0.8}
\begin{tabular}{cccccc}
    Reference & NeRF~\cite{tancik2023nerfstudio} & 3DGS~\cite{kerbl20233d} & FreeVS~\cite{wang2024freevs} & 3DGS+PointmapDiff\\
    \includegraphics[clip=true,trim={0 0 0 0},width=\fgsize\textwidth]{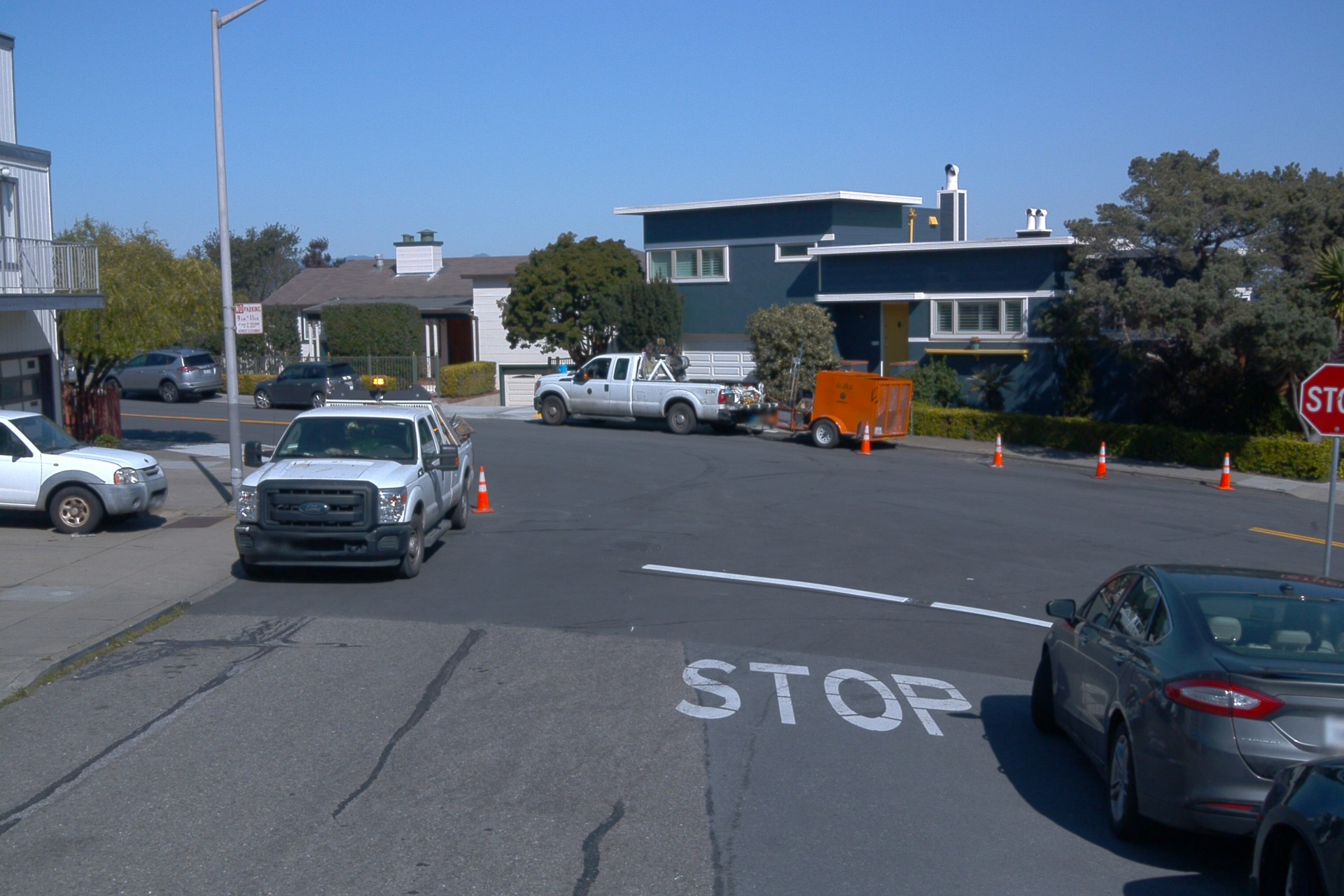} &
    \includegraphics[clip=true,trim={0 0 0 0},width=\fgsize\textwidth]{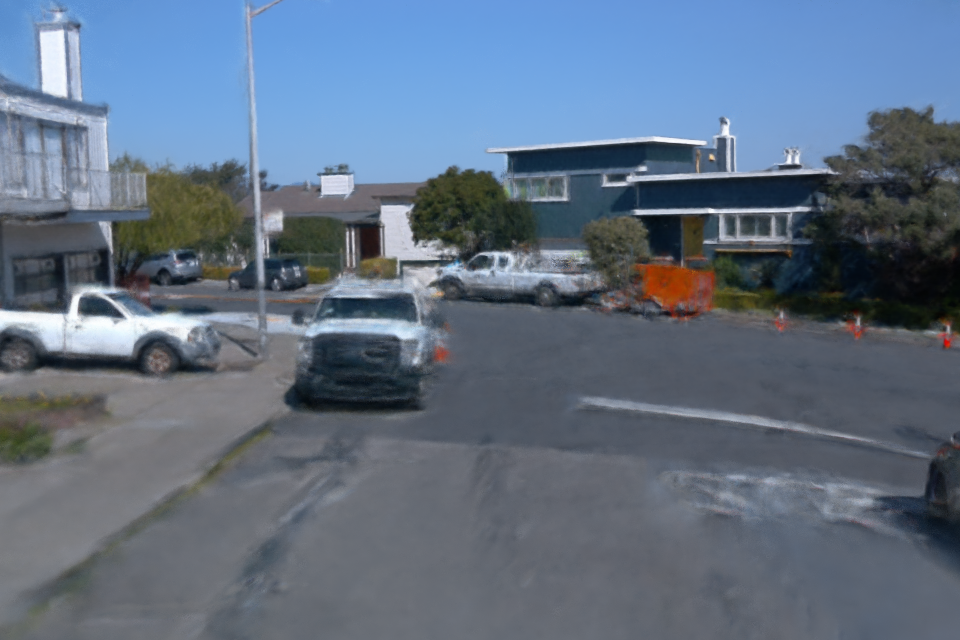} &
    \includegraphics[clip=true,trim={0 0 0 0},width=\fgsize\textwidth]{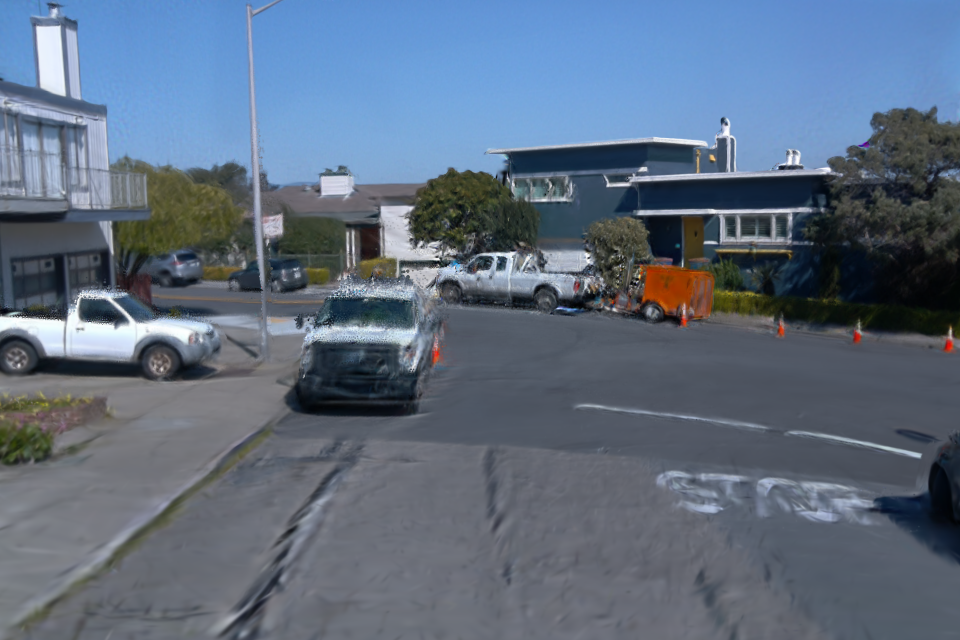} & 
    \includegraphics[clip=true,trim={0 0 0 0},width=\fgsize\textwidth]{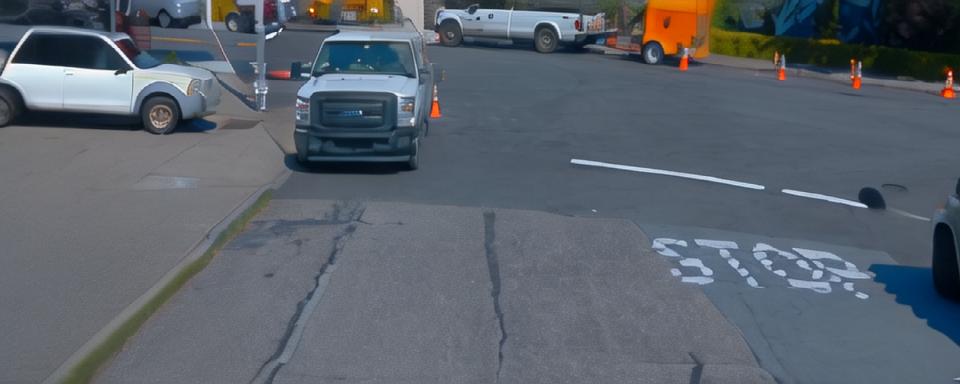} & 
    \includegraphics[clip=true,trim={0 0 0 0},width=\fgsize\textwidth]{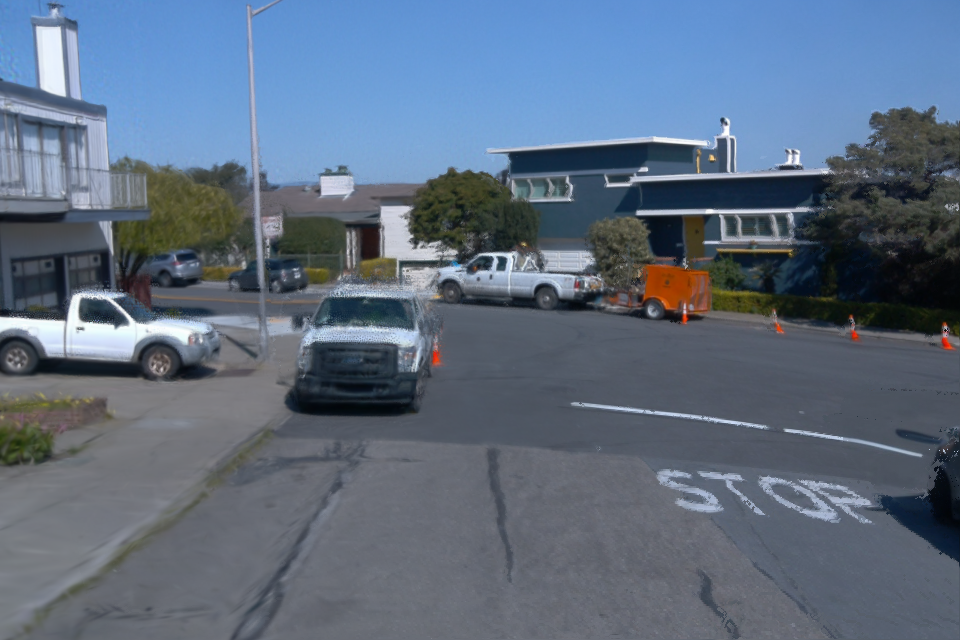}\\

    \includegraphics[clip=true,trim={0 0 0 0},width=\fgsize\textwidth]{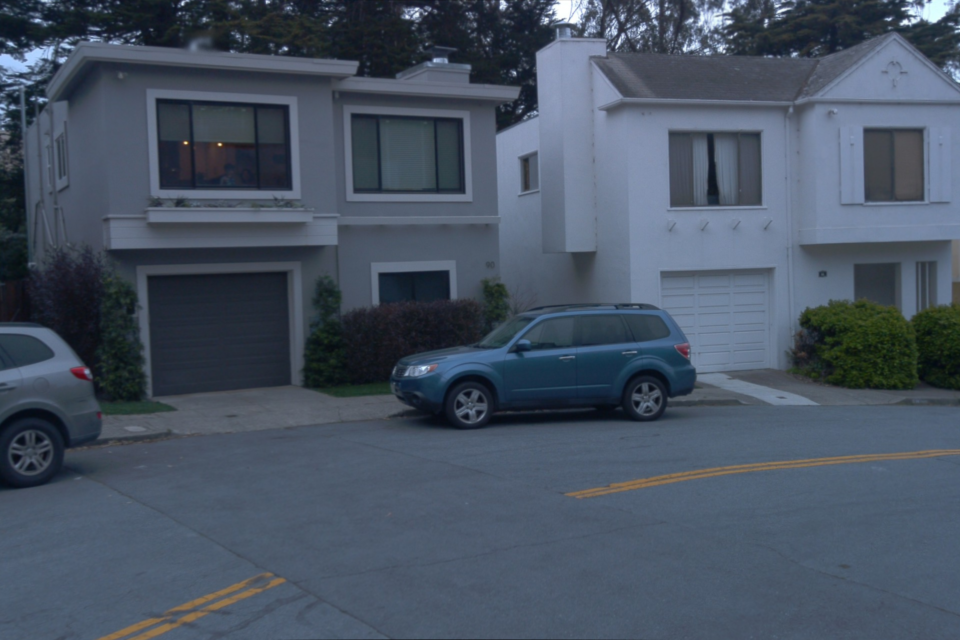} &
    \includegraphics[clip=true,trim={0 0 0 0},width=\fgsize\textwidth]{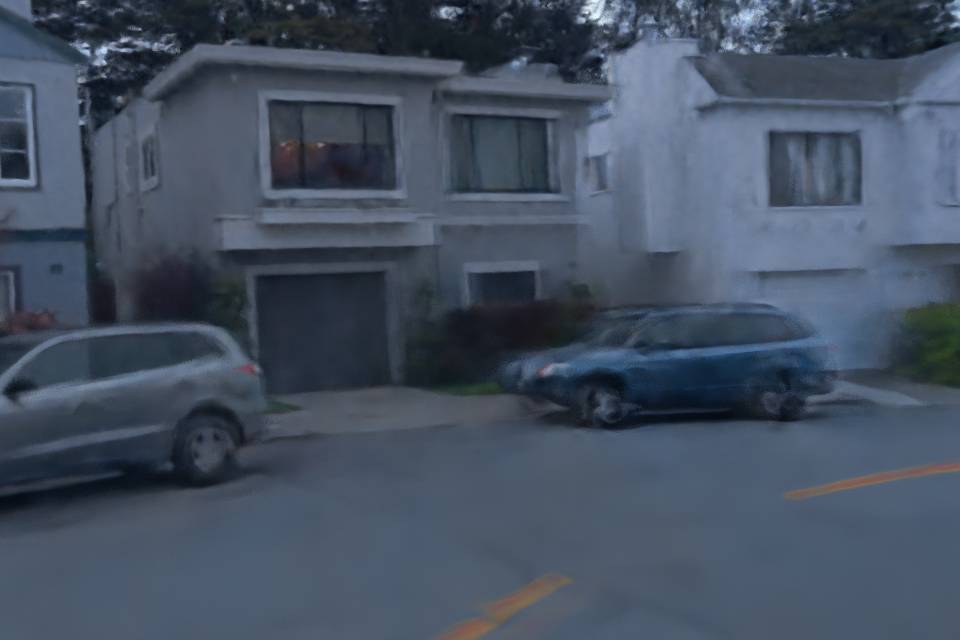} &
    \includegraphics[clip=true,trim={0 0 0 0},width=\fgsize\textwidth]{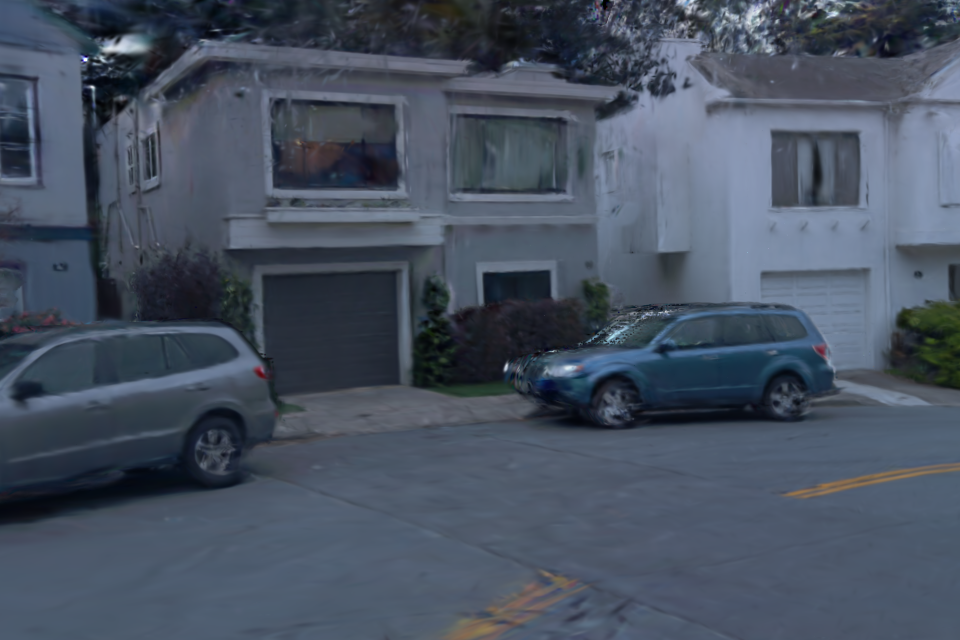} & 
    \includegraphics[clip=true,trim={0 0 0 0},width=\fgsize\textwidth]{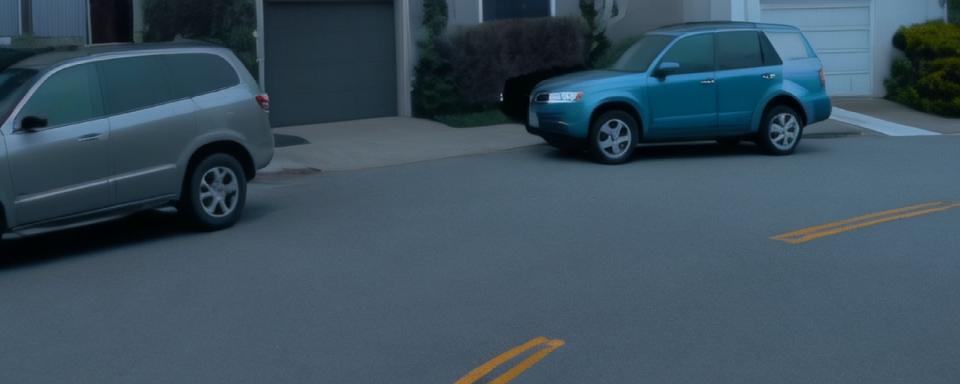} & 
    \includegraphics[clip=true,trim={0 0 0 0},width=\fgsize\textwidth]{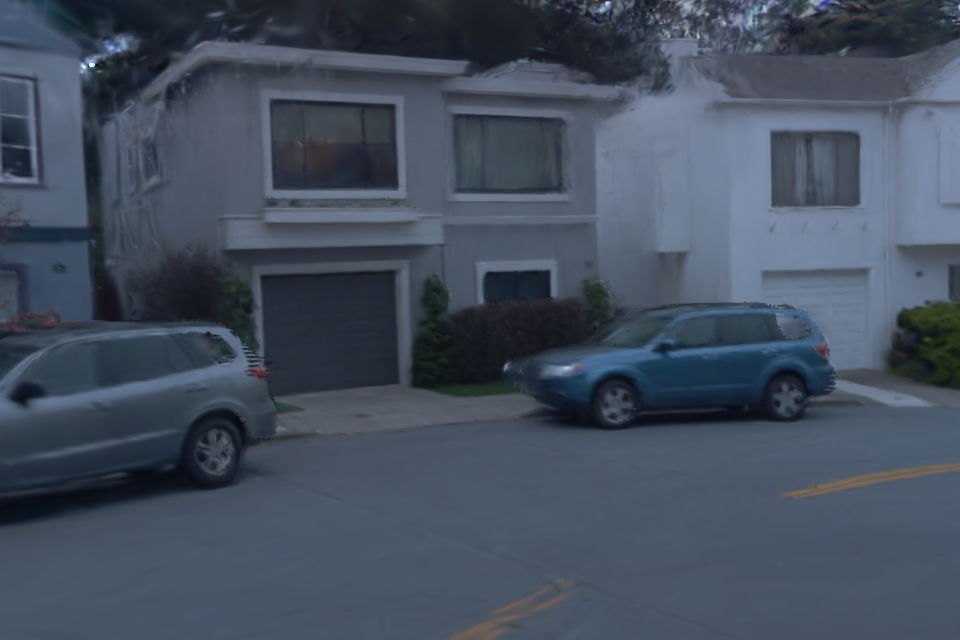}\\
    
    \includegraphics[clip=true,trim={0 0 0 0},width=\fgsize\textwidth]{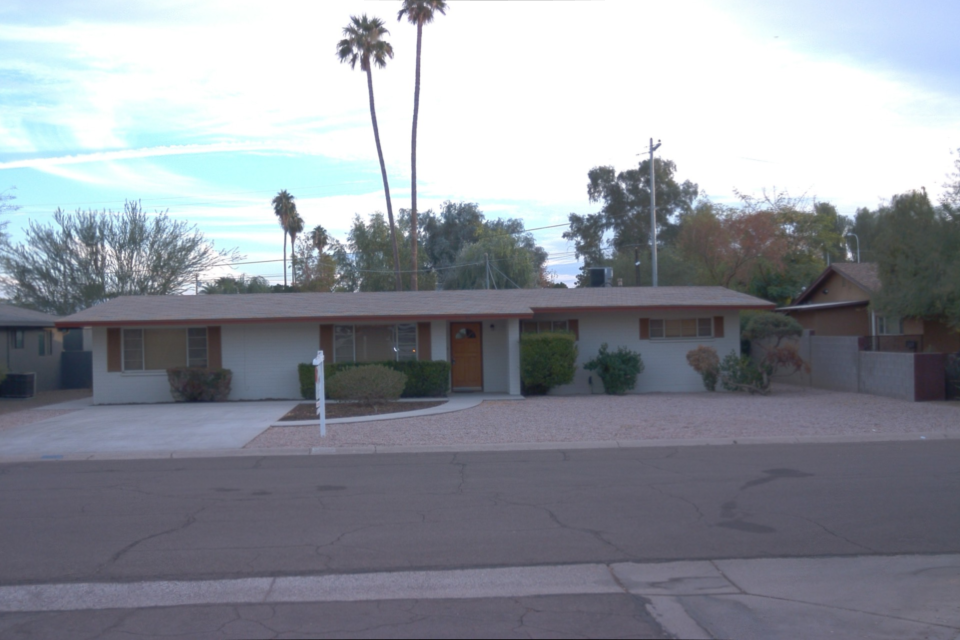} &
    \includegraphics[clip=true,trim={0 0 0 0},width=\fgsize\textwidth]{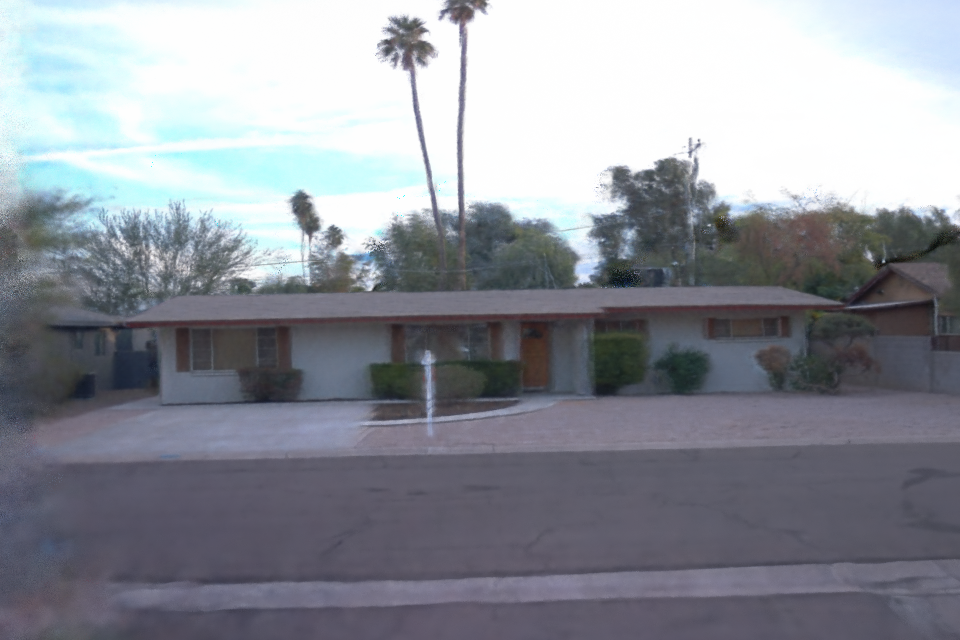} &
    \includegraphics[clip=true,trim={0 0 0 0},width=\fgsize\textwidth]{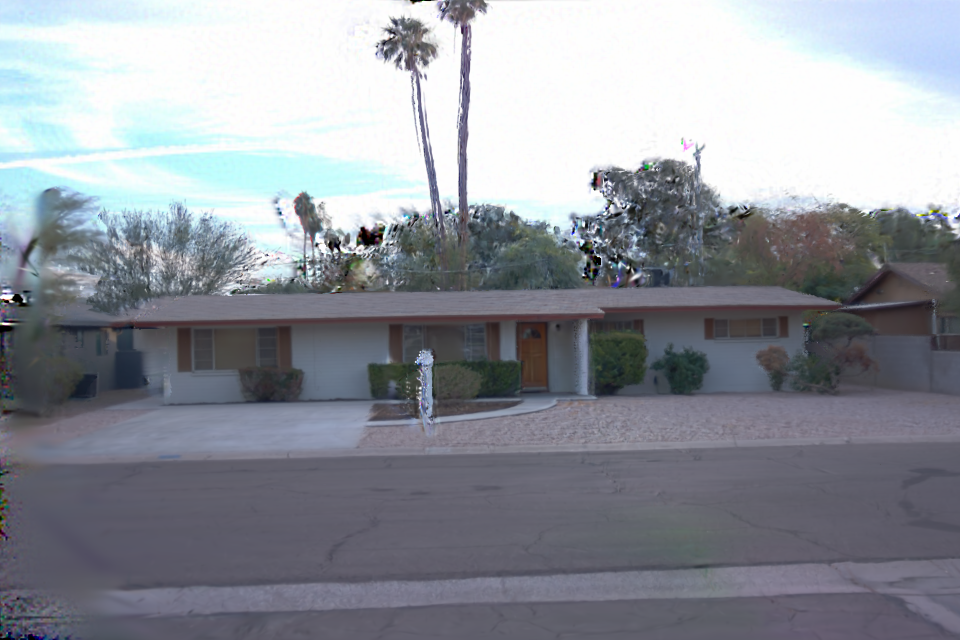} & 
    \includegraphics[clip=true,trim={0 0 0 0},width=\fgsize\textwidth]{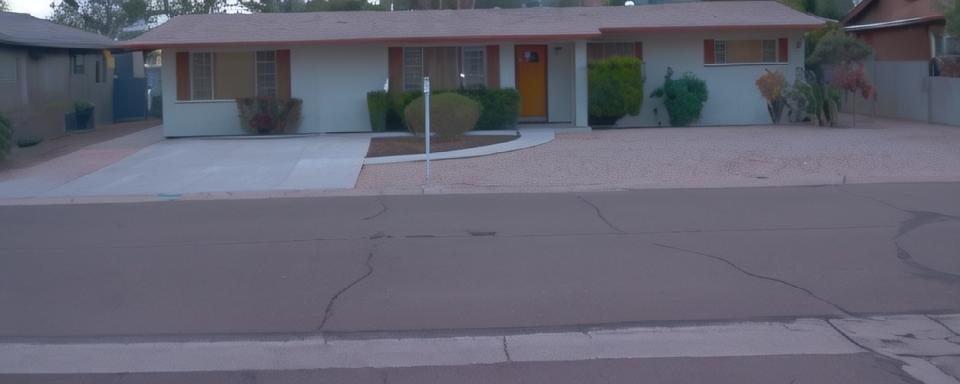} & 
    \includegraphics[clip=true,trim={0 0 0 0},width=\fgsize\textwidth]{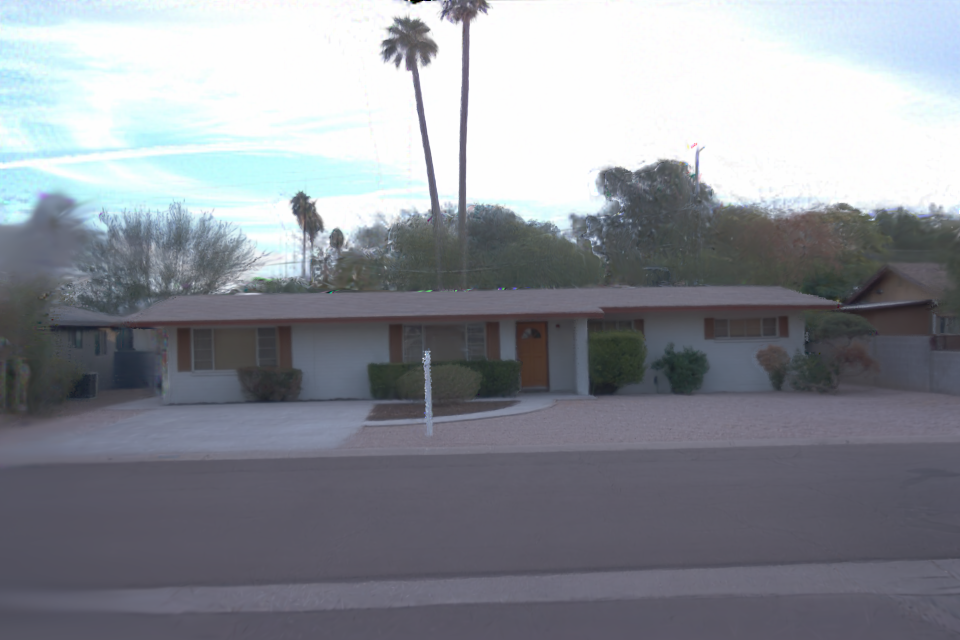}\\

    \includegraphics[clip=true,trim={0 0 0 0},width=\fgsize\textwidth]{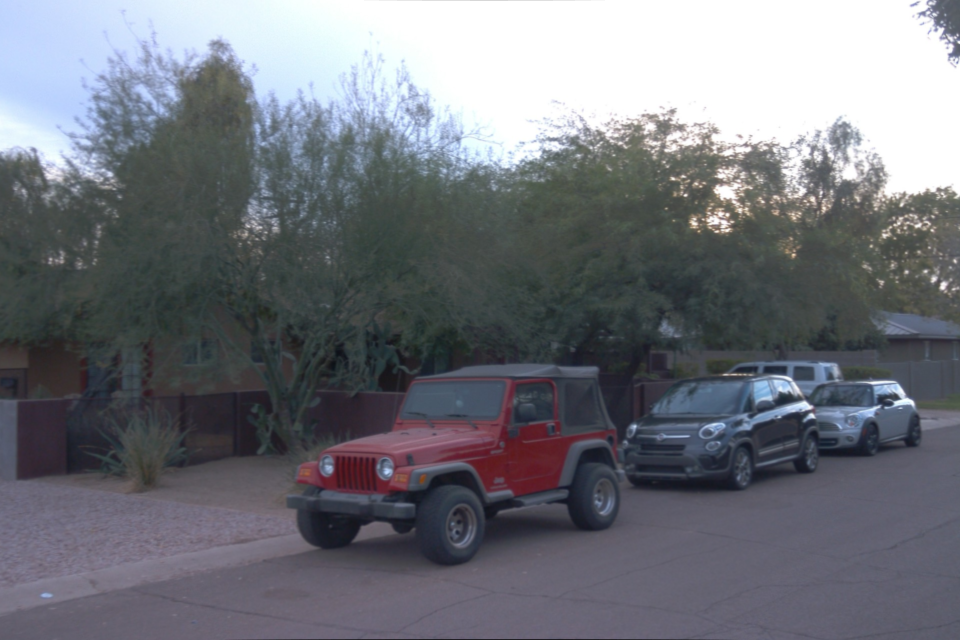} &
    \includegraphics[clip=true,trim={0 0 0 0},width=\fgsize\textwidth]{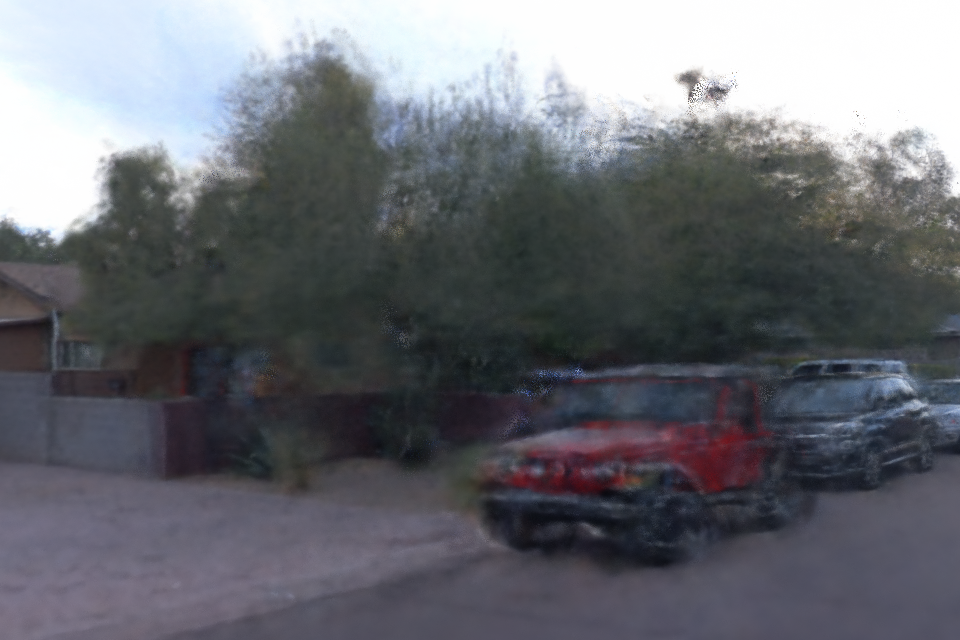} &
    \includegraphics[clip=true,trim={0 0 0 0},width=\fgsize\textwidth]{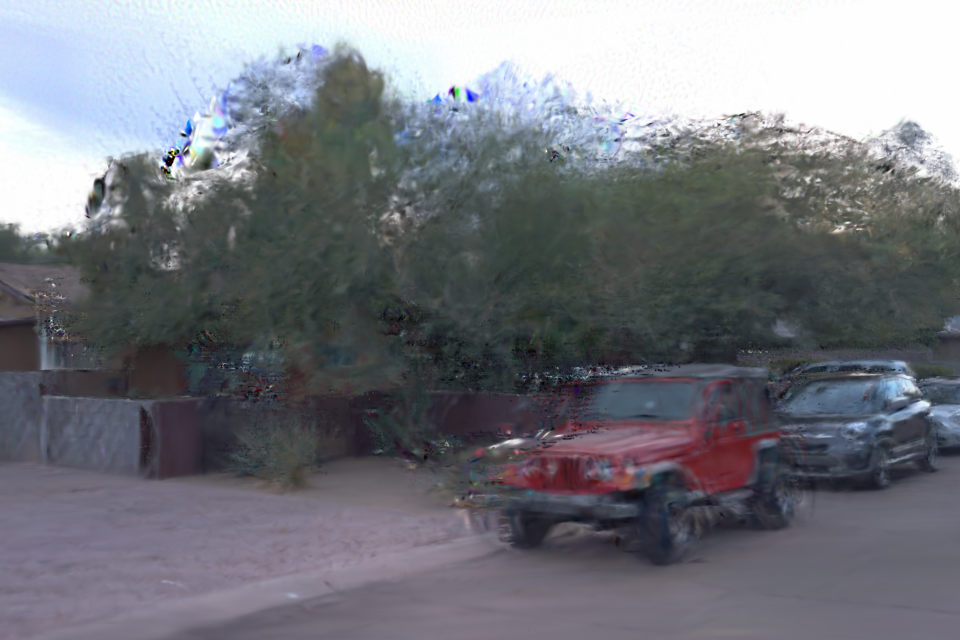} & 
    \includegraphics[clip=true,trim={0 0 0 0},width=\fgsize\textwidth]{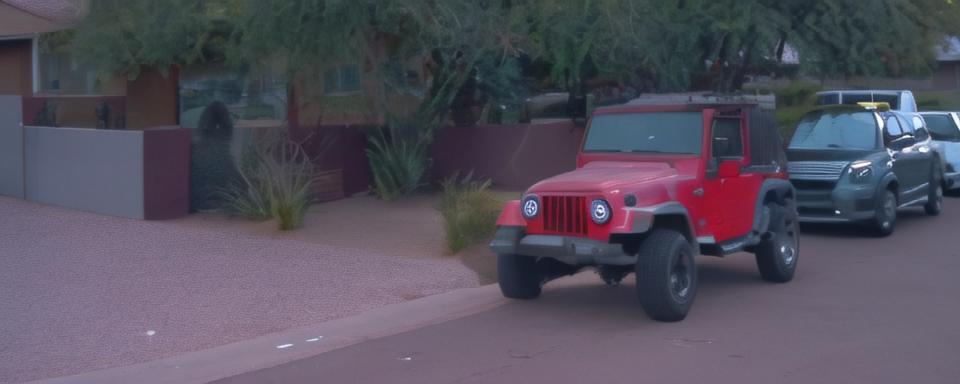} & 
    \includegraphics[clip=true,trim={0 0 0 0},width=\fgsize\textwidth]{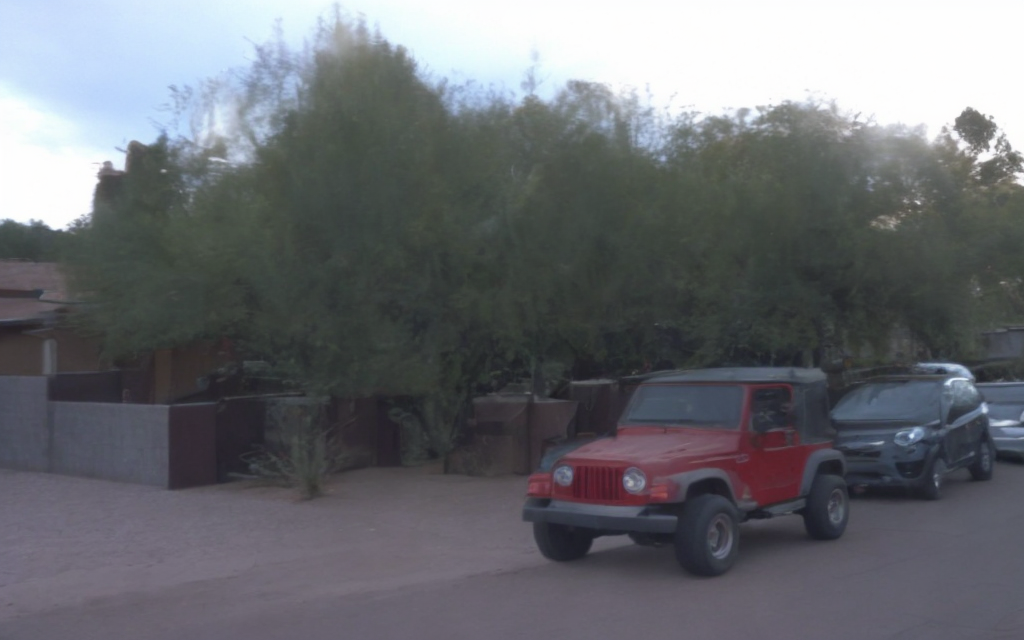}\\

\end{tabular}
\caption{Addtional qualitative comparison on Waymo~\cite{sun2020scalability}.}
\label{fig:extrapolate_waymo_supmat}
\end{figure*}

\section{Additional Results} 
\subsection{Extrapolation in Street View Reconstruction}
\label{sec:supmat_extrapolation}

\noindent \textbf{Optimizing 3DGS.}
We randomly select 10 static sub-sequences per dataset, each with 100-150 consecutive frames for evaluation. For KITTI-360, we train on two perspective images with full resolution $1408\times 376$, and for Waymo, we only use the front camera downsampled to $960 \times 640$.
We initialize the 3D Gaussian models with the accumulated LiDAR point cloud without using Structure-from-Motion (SfM) point clouds.

The loss weights $\lambda_{rgb}$, $\lambda_{ssim}$, $\lambda_{aug}$, $\lambda_{lpips}$, and $\lambda_{d}$ are set to 0.8, 0.2, 0.5, 0.1, and 0.01, respectively. Additionally, we progressively reduce the noise scale $s$ from 0.6 to 0.2 throughout training to ensure harmonization between generation and reconstruction.

\noindent \textbf{Baseline Implementations.}
We adapt the code in the official repository of VEGS, ViewCrafter, and FreeVS. We re-implement SGD since there is no public code base available. For ViewCrafter, we use ground truth images and rendered depth from 3DGS to achieve warped conditions, since predictions from MDE are extremely noisy and not aligned well with the shifted distance. Secondly, as ViewCrafter is a video diffusion model, and requires the first frame to be "clean" (\ie, from ground truth trajectory), we design shifting samples, gradually from the original to the novel trajectory to extract the most details from this initial frame. Since the sequences contain more frames than a video diffusion model can handle at once, the process is divided into smaller chunks and repeated across the entire sequence. The distillation process for ViewCrafter remains mostly the same as with PointmapDiff.
We show additional qualitative comparisons in \cref{fig:extrapolate_kitti_supmat} and \cref{fig:extrapolate_waymo_supmat}.

\subsection{Single-image NVS on Street View}
\begin{figure*}
\centering
\setlength{\tabcolsep}{0.002\linewidth}
\renewcommand{\arraystretch}{0.8}
\begin{tabular}{ccccc}
    Source view & Inpainting~\cite{rombach2022high} & GenWarp~\cite{seo2024genwarp} & PointmapDiff & Target view (GT)\\

    \includegraphics[clip=true,trim={0 0 0 0},width=\fgsize\textwidth]{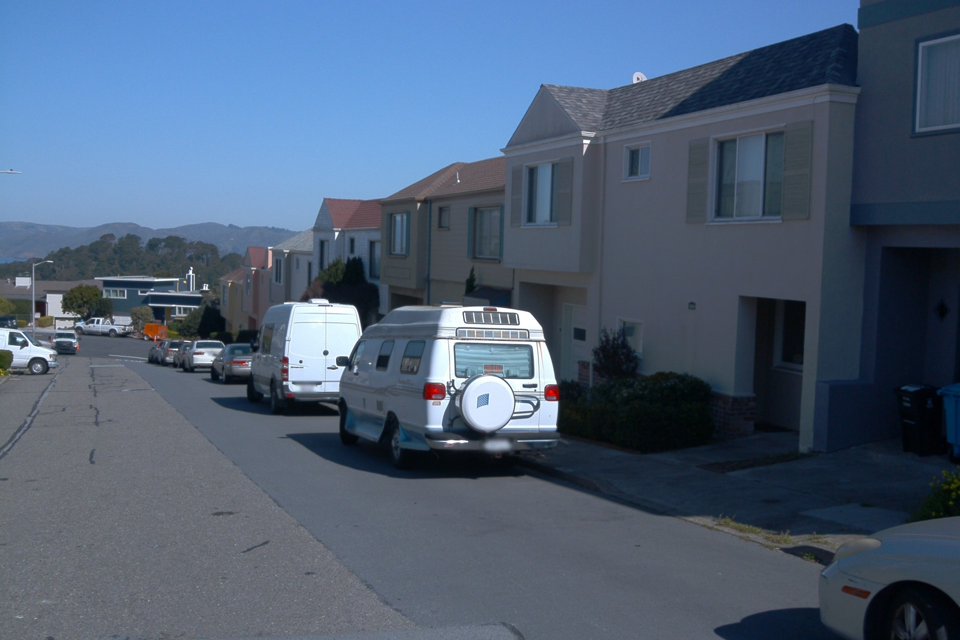} & 
    \includegraphics[clip=true,trim={0 0 0 0},width=\fgsize\textwidth]{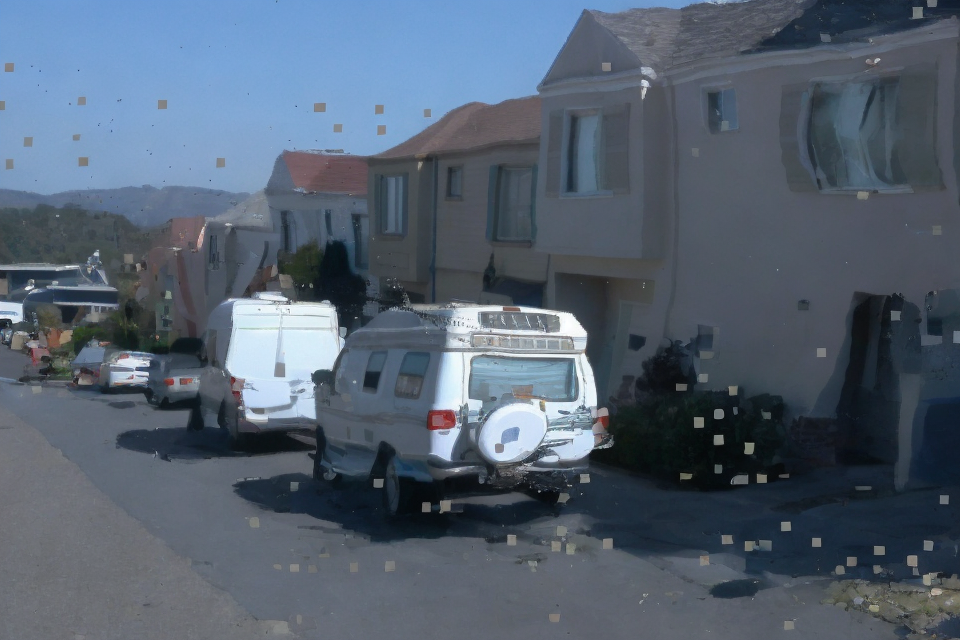} & 
    \includegraphics[clip=true,trim={0 0 0 0},width=\fgsize\textwidth]{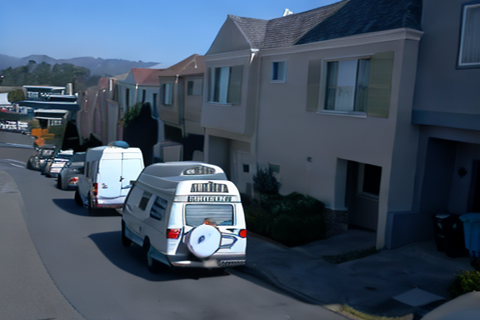} & 
    \includegraphics[clip=true,trim={50 18 0 18},width=\fgsize\textwidth]{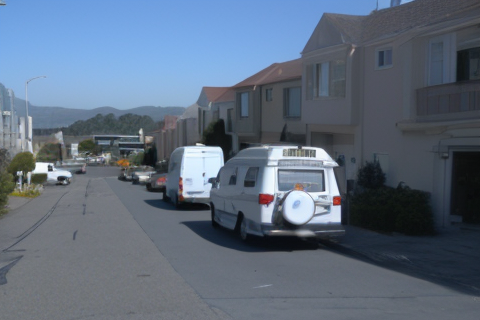}  & 
    \includegraphics[clip=true,trim={50 18 0 18},width=\fgsize\textwidth]{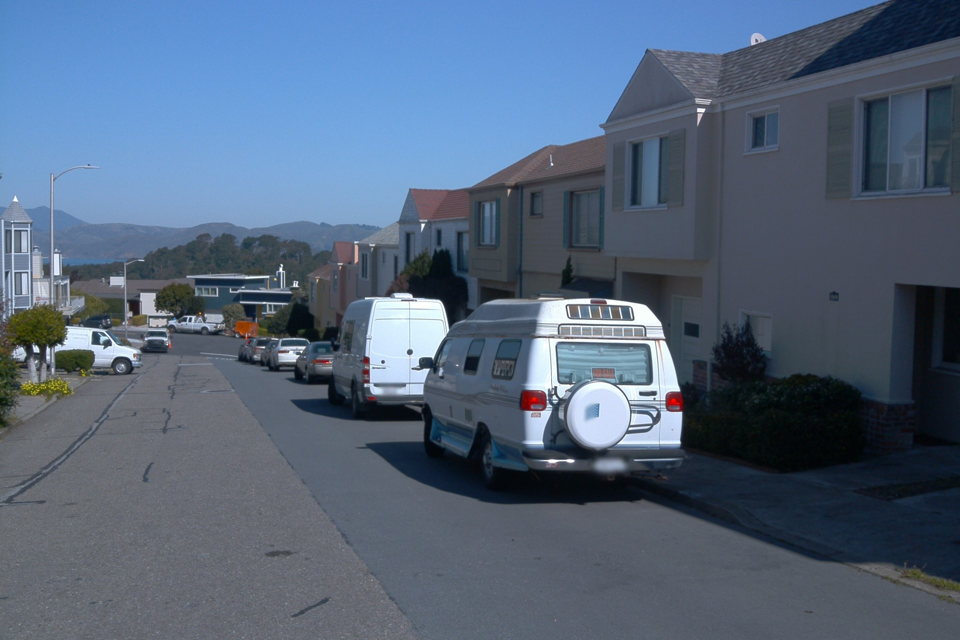}\\

    \includegraphics[clip=true,trim={0 0 0 0},width=\fgsize\textwidth]{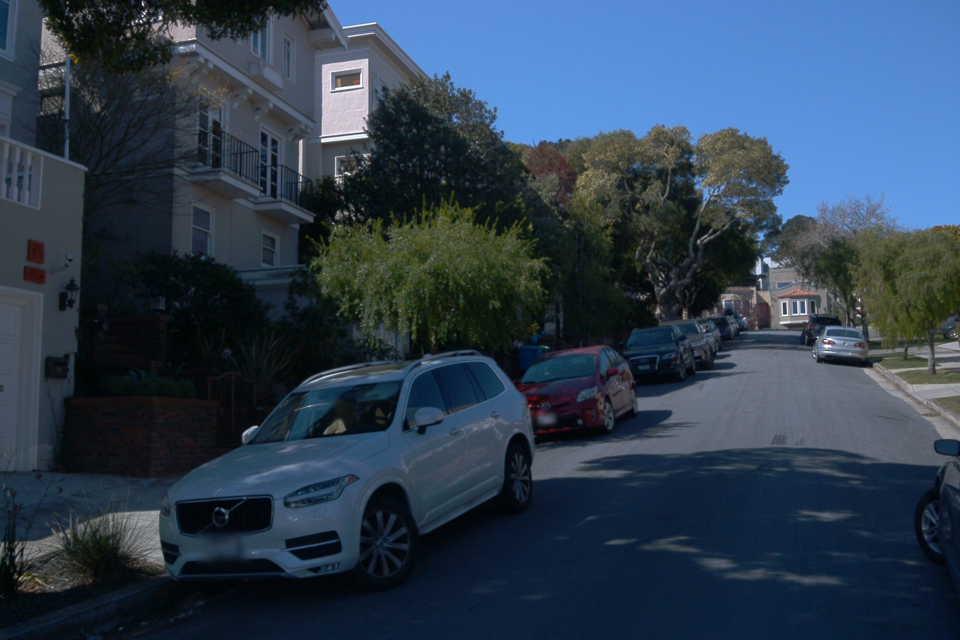} & 
    \includegraphics[clip=true,trim={0 0 0 0},width=\fgsize\textwidth]{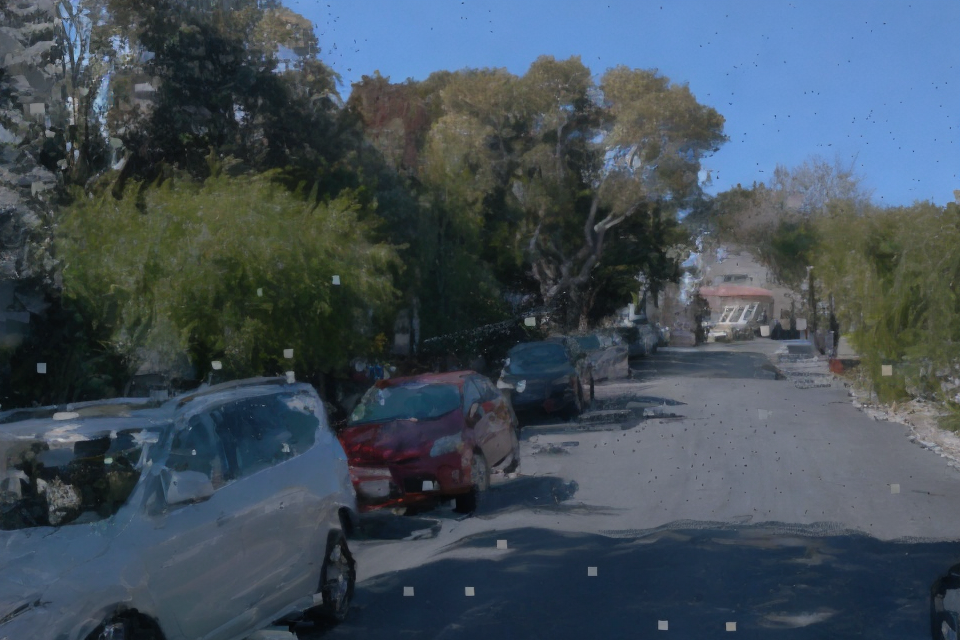} & 
    \includegraphics[clip=true,trim={0 0 0 0},width=\fgsize\textwidth]{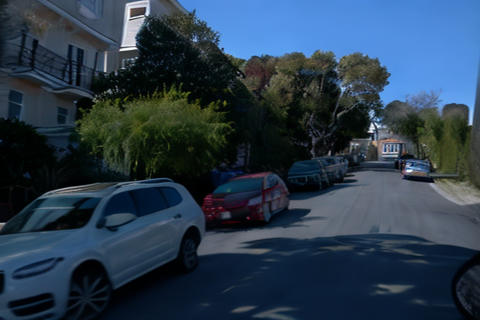} & 
    \includegraphics[clip=true,trim={0 0 0 0},width=\fgsize\textwidth]{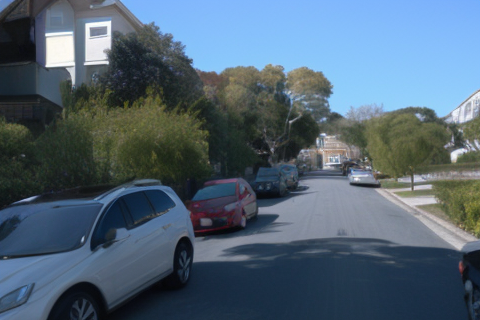} & 
    \includegraphics[clip=true,trim={0 0 0 0},width=\fgsize\textwidth]{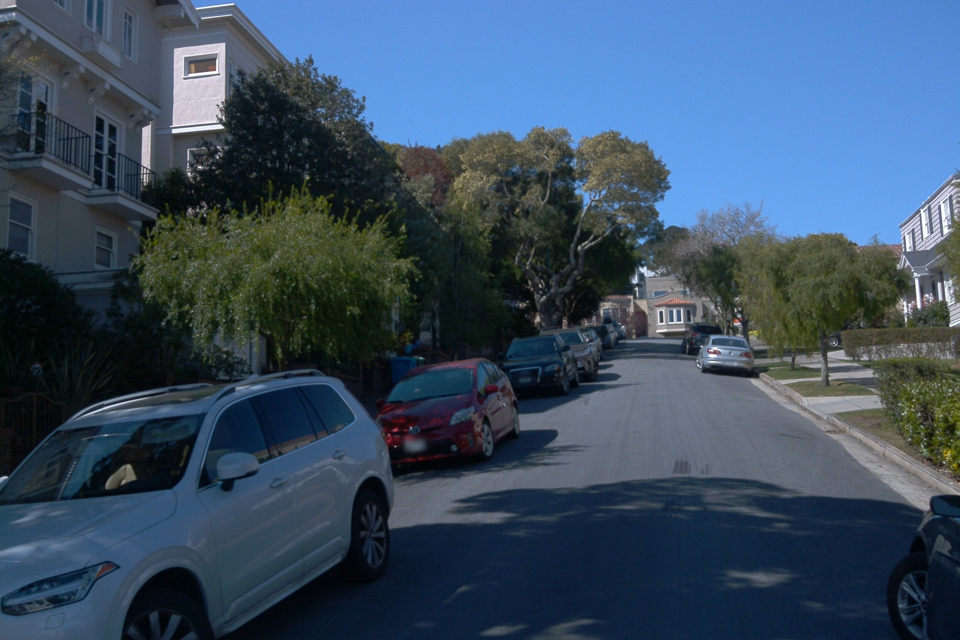}\\
\end{tabular}
\caption{Qualitative comparison for single-image NVS on Waymo~\cite{sun2020scalability}.}
\label{fig:single_nvs_waymo}
\end{figure*}

We provide results for single-image NVS task on Waymo~\cite{sun2020scalability} in  \cref{fig:single_nvs_waymo}. We utilize Metric3D~\cite{hu2024metric3d} to estimate depth, as there is no reliable depth completion model available for Waymo.

\subsection{Single-image NVS on Indoor Data}

\def\fgsize{0.136}

\begin{figure*}
\centering
\scriptsize
\setlength{\tabcolsep}{0.002\linewidth}
\renewcommand{\arraystretch}{0.8}
\begin{tabular}{lcccc|ccc}
    \multirow{1}{*}[17mm]{\rotatebox[origin=c]{90}{Source view}}
    & \multicolumn{2}{c}{\includegraphics[width=\fgsize\textwidth]{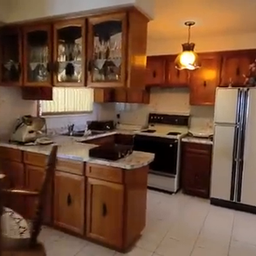}} & 
    \multicolumn{2}{c|}{\includegraphics[width=\fgsize\textwidth]{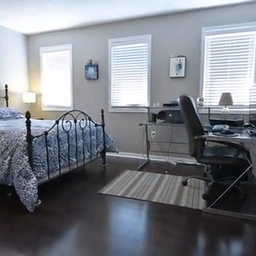}} &  
    {\includegraphics[width=\fgsize\textwidth]{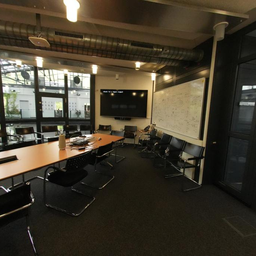}} &
    {\includegraphics[width=\fgsize\textwidth]{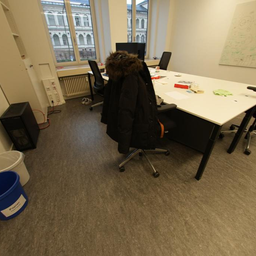}} & 
    {\includegraphics[width=\fgsize\textwidth]{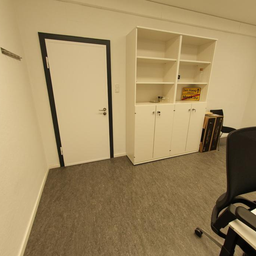}}\\

    \multirow{1}{*}[19mm]{\rotatebox[origin=c]{90}{Target view (GT)}}
    & \includegraphics[width=\fgsize\textwidth]{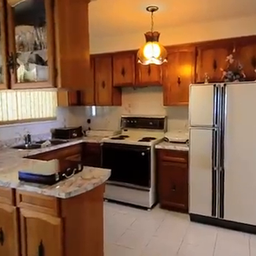} & 
    \includegraphics[width=\fgsize\textwidth]{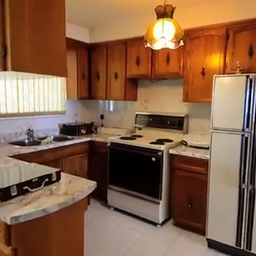} & 
    \includegraphics[width=\fgsize\textwidth]{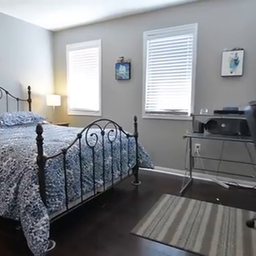} & 
    \includegraphics[width=\fgsize\textwidth]{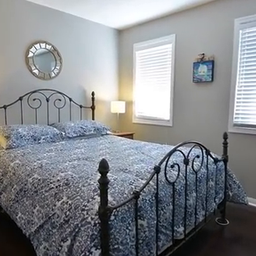}  & 
    \includegraphics[width=\fgsize\textwidth]{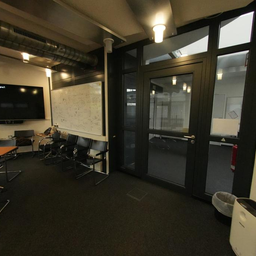} & 
    \includegraphics[width=\fgsize\textwidth]{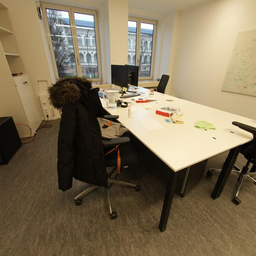} &
    {\includegraphics[width=\fgsize\textwidth]{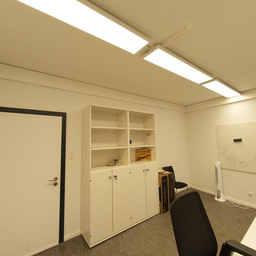}}\\
    
    \multirow{1}{*}[17mm]{\rotatebox[origin=c]{90}{GeoGPT~\cite{rombach2021geometry}}}
    & \includegraphics[width=\fgsize\textwidth]{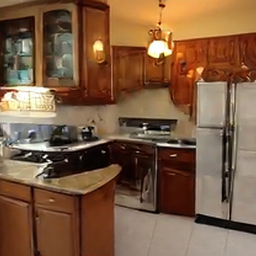} & 
    \includegraphics[width=\fgsize\textwidth]{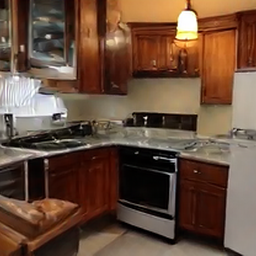} & 
    \includegraphics[width=\fgsize\textwidth]{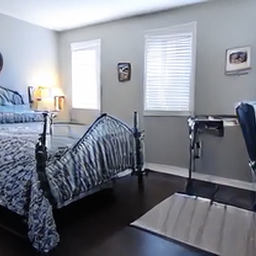} & 
    \includegraphics[width=\fgsize\textwidth]{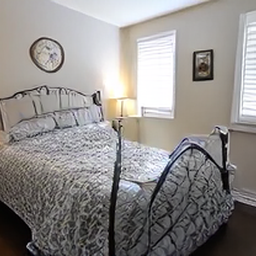}  & 
    \includegraphics[width=\fgsize\textwidth]{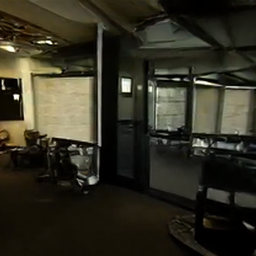} & 
    \includegraphics[width=\fgsize\textwidth]{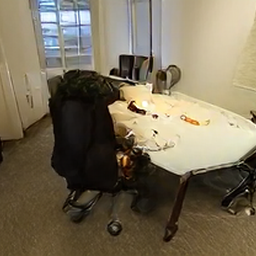} &
    {\includegraphics[width=\fgsize\textwidth]{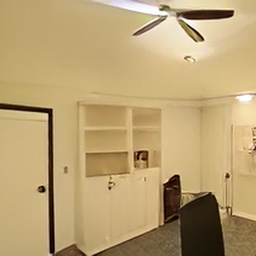}}\\

    \multirow{1}{*}[19mm]{\rotatebox[origin=c]{90}{Photo-NVS~\cite{saharia2022photorealistic}}}
    & \includegraphics[width=\fgsize\textwidth]{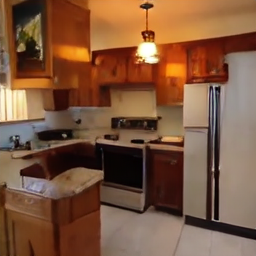} & 
    \includegraphics[width=\fgsize\textwidth]{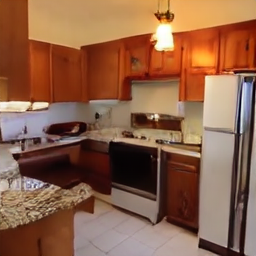} & 
    \includegraphics[width=\fgsize\textwidth]{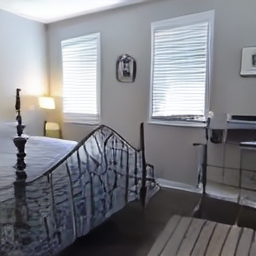} & 
    \includegraphics[width=\fgsize\textwidth]{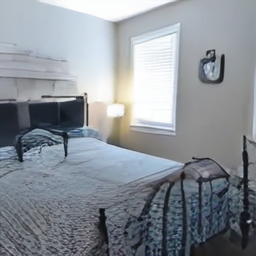}  & 
    \includegraphics[width=\fgsize\textwidth]{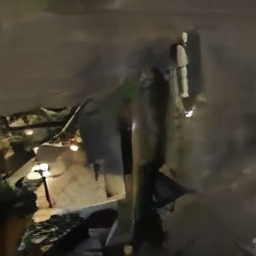} & 
    \includegraphics[width=\fgsize\textwidth]{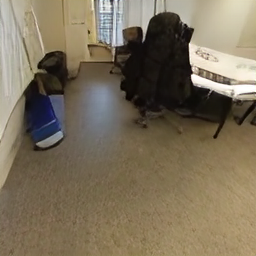} &
    {\includegraphics[width=\fgsize\textwidth]{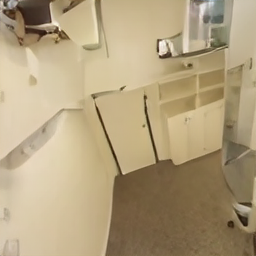}}\\

    \multirow{1}{*}[18mm]{\rotatebox[origin=c]{90}{Inpainting~\cite{rombach2022high}}}
    & \includegraphics[width=\fgsize\textwidth]{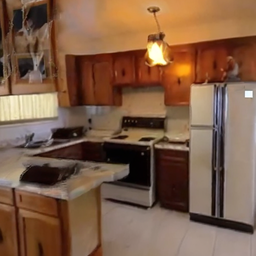} & 
    \includegraphics[width=\fgsize\textwidth]{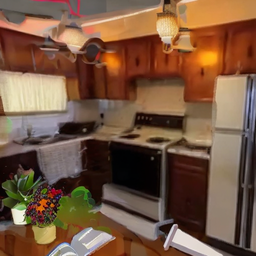} & 
    \includegraphics[width=\fgsize\textwidth]{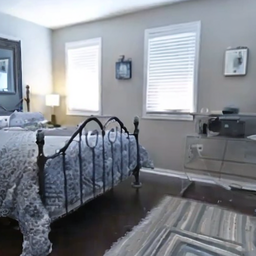} & 
    \includegraphics[width=\fgsize\textwidth]{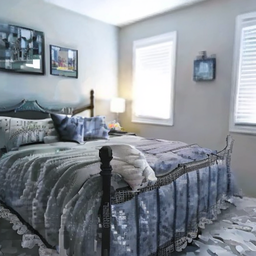}  & 
    \includegraphics[width=\fgsize\textwidth]{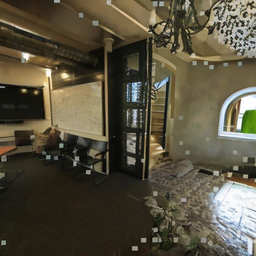} & 
    \includegraphics[width=\fgsize\textwidth]{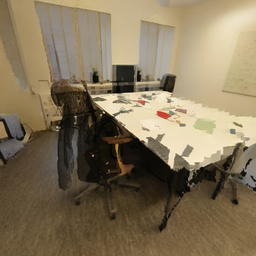} &
    {\includegraphics[width=\fgsize\textwidth]{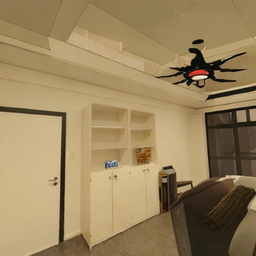}}\\

    \multirow{1}{*}[17mm]{\rotatebox[origin=c]{90}{GenWarp~\cite{seo2024genwarp}}}
    & \includegraphics[width=\fgsize\textwidth]{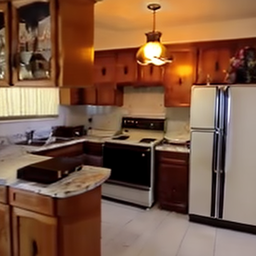} & 
    \includegraphics[width=\fgsize\textwidth]{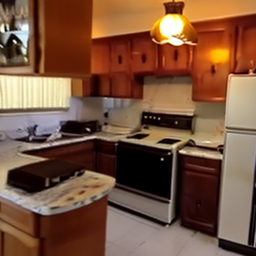} & 
    \includegraphics[width=\fgsize\textwidth]{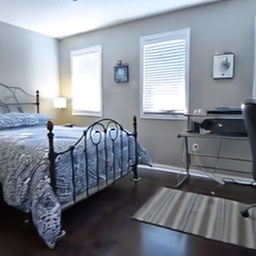} & 
    \includegraphics[width=\fgsize\textwidth]{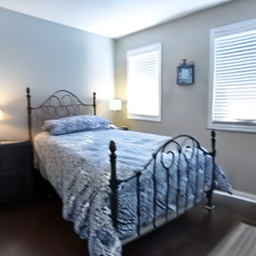} &
    \includegraphics[width=\fgsize\textwidth]{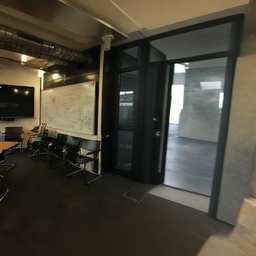} & 
    \includegraphics[width=\fgsize\textwidth]{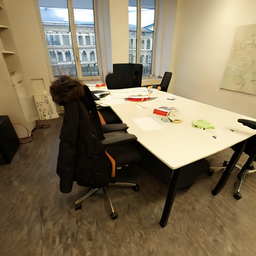} &
    {\includegraphics[width=\fgsize\textwidth]{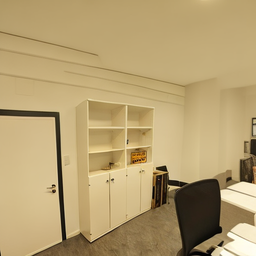}}\\

    \multirow{1}{*}[17mm]{\rotatebox[origin=c]{90}{PointmapDiff}}
    & \includegraphics[width=\fgsize\textwidth]{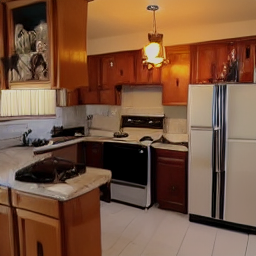} & 
    \includegraphics[width=\fgsize\textwidth]{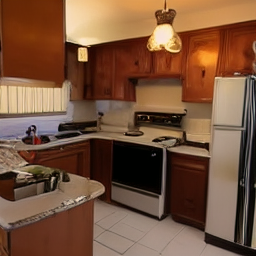} & 
    \includegraphics[width=\fgsize\textwidth]{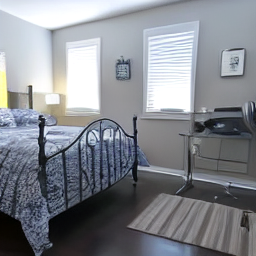} & 
    \includegraphics[width=\fgsize\textwidth]{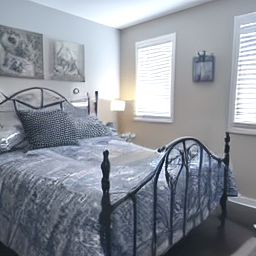}  & 
    \includegraphics[width=\fgsize\textwidth]{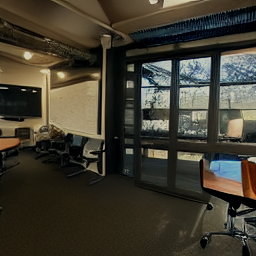} & 
    \includegraphics[width=\fgsize\textwidth]{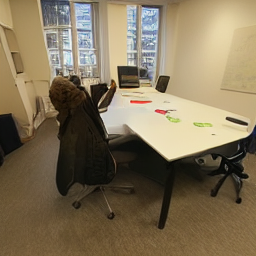} &
    {\includegraphics[width=\fgsize\textwidth]{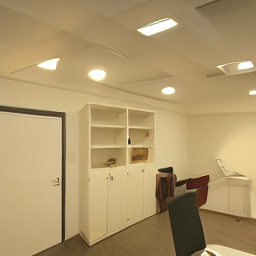}}\\
    & \multicolumn{4}{c}{RealEstate10K~\cite{zhou2018stereo}} & \multicolumn{3}{c}{ScanNet++~\cite{yeshwanth2023scannet++}}\\
\end{tabular}
\caption{Qualitative comparison for single-image NVS on RealEstate10K~\cite{zhou2018stereo} and ScanNet++~\cite{yeshwanth2023scannet++}.}
\label{fig:single_nvs_indoor}
\end{figure*}

\noindent\textbf{Baselines.}
These include GeoGPT~\cite{rombach2021geometry}, Photoconsistent-NVS ~\cite{yu2023long}, the warping and inpainting method using the SD Inpainting~\cite{rombach2022high}, and GenWarp~\cite{seo2024genwarp}. To ensure fair comparisons, we train our model only on RealEstate10K~\cite{zhou2018stereo}, aligning with the training data used by our baselines, and further evaluate on ScanNet++~\cite{yeshwanth2023scannet++} to assess performance on out-of-distribution scenarios. 
We use the officially provided checkpoint of all methods. For GeoGPT, we choose the \texttt{re\_impl\_depth} checkpoint as it requires reference depth maps and produces better results compared to the version that does not use depth information. Moreover, for SD-Inpainting, we apply interpolation on the warped images and dilate the inpaint mask using a $9\times 9$ kernel to reduce artifacts since the model performs inpainting on latent space (\textit{f8d4}). In contrast, only PhotoNVS does not require depth as an input.

\noindent\textbf{Setup.}
We utilize DUSt3R~\cite{wang2024dust3r} to generate point maps for training and as a depth estimator (by taking the z-values of the point map in local coordinate) for inference of all baselines.
Similar to~\cite{ren2022look, seo2024genwarp}, we consider dividing into short-term and long-term view synthesis. Specifically, we randomly select 1k sequences from the test set with more than 200 frames and evaluate the $50^{\text{th}}$ generated frame as short-term and the $100^{\text{th}}$ generated frame as long-term view synthesis on RealEstate10K. Due to the faster camera movement in ScanNet++, we focus solely on short-term synthesis. 

\noindent\textbf{Metrics.}
For short-term, we use pairwise reconstruction metrics PSNR and LPIPS~\cite{zhang2018unreasonable} to measure the difference between the generated and ground-truth images. Note that these metrics become less relevant in regions that are unseen.
For long-term, we value generated image quality, using the FID~\cite{heusel2017gans} and KID ($\times 100$)~\cite{binkowski2018demystifying} to estimate the realism of the generated image distribution.
Finally, all outputs are resized and cropped to $256 \times 256$ for evaluation. 

\cref{tab:quantitative_re10k} demonstrates that while GeoGPT achieves good FID and KID, indicating realistic generation quality, it struggles with misalignment issues from the input view, leading to worse PSNR and LPIPS scores. In contrast, the inpainting method excels in PSNR, benefiting from explicit warping. However, it often suffers from artifacts due to the imperfect depth, resulting in lower FID and KID.

For the out-of-distribution experiment, as shown in \cref{tab:quantitative_scannet}, GeoGPT and Photoconsistent-NVS struggle to generalize to out-of-domain scenarios, resulting in poor performance metrics and a noticeable drop in generation quality. On the other hand, our method achieves stable and consistent results across both in-domain and out-of-domain datasets, indicating improved adaptability and maintaining high-quality view synthesis under diverse conditions while mitigating overfitting.

\begin{table}[htbp]
    \centering
    \begin{tabular}{lcccc}
    \toprule
    {} & \multicolumn{2}{c}{Short-term} & \multicolumn{2}{c}{Long-term} \\
    \cmidrule(rl){2-3} 
    \cmidrule(rl){4-5}
     
    & PSNR$\uparrow$ & LPIPS$\downarrow$ & FID$\downarrow$ & KID$\downarrow$\\
    \midrule
    
    GeoGPT~\cite{rombach2021geometry} & 14.97 & 0.356 & \best{28.42} & \best{0.158} \\
    Photo-NVS~\cite{saharia2022photorealistic} & 15.74 & 0.309 & \tbest{30.96} & \sbest{0.305}\\
    Inpainting~\cite{rombach2022high} & \best{16.29} & \tbest{0.300} & 47.63 & 1.546\\
    GenWarp~\cite{seo2024genwarp} & \sbest{16.04} & \sbest{0.272} & 32.34 & \tbest{0.446}\\
    PointmapDiff & \tbest{15.87} & \best{0.237} & \sbest{29.65} & \tbest{0.446}\\
    \bottomrule
    \end{tabular}
    \caption{Quantitative results on RealEstate10K~\cite{zhou2018stereo}.}
    \label{tab:quantitative_re10k}  
\end{table}

\begin{table}[htbp]
\centering
\begin{tabular}{lcccc}
    \toprule
    & \multicolumn{4}{c}{Short-term}\\
    \cmidrule(rl){2-5} 
     
    & PSNR$\uparrow$ & LPIPS$\downarrow$ & FID$\downarrow$ & KID$\downarrow$\\
    \midrule
    
    GeoGPT~\cite{rombach2021geometry} & 14.50 & 0.328 & 62.70 & 2.256\\
    Photo-NVS~\cite{saharia2022photorealistic} & 11.72 & 0.525 & 90.05 & 4.143\\
    Inpainting~\cite{rombach2022high} & \tbest{15.09} & \tbest{0.312} & \tbest{56.08} & \tbest{1.647}\\
    GenWarp*~\cite{seo2024genwarp} & \best{15.95} & \best{0.248} & \best{29.63} & \best{0.336}\\
    PointmapDiff & \sbest{15.19} & \sbest{0.303} & \sbest{38.72} & \sbest{0.560}\\
    \bottomrule
\end{tabular}
\caption{Quantitative results on Scannet++~\cite{yeshwanth2023scannet++}. GenWarp achieves slightly better results because it is trained on datasets beyond RealEstate10K.}
\label{tab:quantitative_scannet}  
\end{table}

\cref{fig:single_nvs_indoor} shows qualitative comparisons on RealEstate10K and ScanNet++. 
The inpainting method performs well in regions where there is a clear overlap between the input and the novel views. However, in areas with sparse warped pixels, it produces inconsistent novel views, failing to take into account the information from the surrounding input pixels, which impacts the overall coherence.
Our method consistently synthesizes realistic and stable novel views across both small and large viewpoint changes, compatible with the quality of GenWarp despite training on less data.

\section{Additional Analysis}
\subsection{Multi-View Conditioning}
Our method can be easily extended to condition on a set of multiple reference images, $\left\{I^{r_1}, \dots, I^{r_k}\right\}$. This is achieved by concatenating the keys and values from all the reference images, as all point maps share the same coordinate system (i.e., the target coordinate). This allows the model to naturally integrate information from multiple reference views and inherently decide which views it should rely more on during generation, enhancing the quality and consistency of the output. Formally, the key and value with multiple images guidance are obtained with the following expressions:
\begin{equation}
    K^{r}=W^{K}[f^{r_1},\dots, f^{r_k}]; V^{r}=W^{V}[f^{r_1},\dots, f^{r_k}].
\end{equation}

\begin{figure}
\captionsetup{justification=centering}
\centering    
\begin{subfigure}{0.32\linewidth}
    \includegraphics[width=\linewidth]{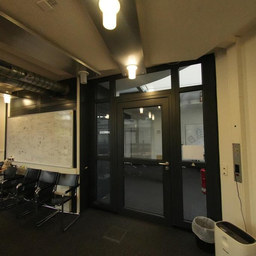}
    \caption*{Source view 1}
\end{subfigure}
\begin{subfigure}{0.32\linewidth}
    \includegraphics[width=\linewidth]{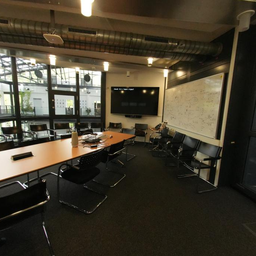}
    \caption*{Source view 2}
\end{subfigure}
\begin{subfigure}{0.32\linewidth}
    \begin{tikzpicture}
    \node[anchor=south west,inner sep=0] (image) at (0,0) {\includegraphics[width=\linewidth]{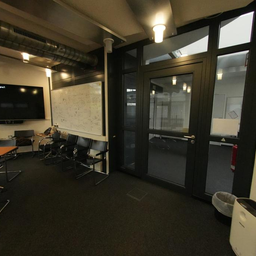}};
    \begin{scope}[x={(image.south east)},y={(image.north west)}]
        \draw[red,thick] (0.01,0.3) rectangle (0.2,0.7);
        \draw[red,thick] (0.5,0.3) rectangle (0.99,0.8);
    \end{scope}
    \end{tikzpicture}
    \caption*{Target view}
\end{subfigure}
\begin{subfigure}{0.32\linewidth}
    \begin{tikzpicture}
    \node[anchor=south west,inner sep=0] (image) at (0,0) {\includegraphics[width=\linewidth]{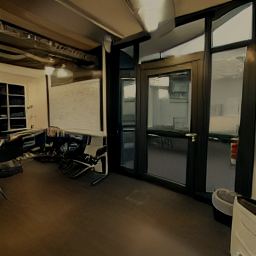}};
    \begin{scope}[x={(image.south east)},y={(image.north west)}]
        \draw[red,thick] (0.01,0.3) rectangle (0.2,0.7);
        \draw[green,thick] (0.5,0.3) rectangle (0.99,0.8);
    \end{scope}
    \end{tikzpicture}
    \caption*{Prediction from \\source view 1}
\end{subfigure}
\begin{subfigure}{0.32\linewidth}
    \begin{tikzpicture}
    \node[anchor=south west,inner sep=0] (image) at (0,0) {\includegraphics[width=\linewidth]{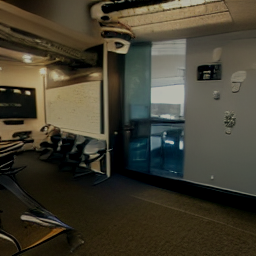}};
    \begin{scope}[x={(image.south east)},y={(image.north west)}]
        \draw[green,thick] (0.01,0.3) rectangle (0.2,0.7);
        \draw[red,thick] (0.5,0.3) rectangle (0.99,0.8);
    \end{scope}
    \end{tikzpicture}
    \caption*{Prediction from \\source view 2}
\end{subfigure}
\begin{subfigure}{0.32\linewidth}
    \begin{tikzpicture}
    \node[anchor=south west,inner sep=0] (image) at (0,0) {\includegraphics[width=\linewidth]{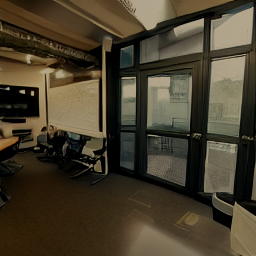}};
    \begin{scope}[x={(image.south east)},y={(image.north west)}]
        \draw[green,thick] (0.01,0.3) rectangle (0.2,0.7);
        \draw[green,thick] (0.5,0.3) rectangle (0.99,0.8);
    \end{scope}
    \end{tikzpicture}
    \caption*{Prediction from \\both source views}
\end{subfigure}
\captionsetup{justification=justified}
\caption{We demonstrate the results when generating viewpoints between two source views, effectively covering occluded regions by combining complementary information from both views. We use \textcolor{red}{red} to denote hallucinated regions and \textcolor{teal}{green} to indicate aligned regions compared to the target view.}
\label{fig:mv_conditioning}
\end{figure}

While our model has been trained using only one reference view, it is worth emphasizing that it can benefit from multiple reference view conditioning without further fine-tuning or modification. This approach helps reconstruct by leveraging details visible in alternate views, resulting in a more coherent and complete scene, as shown in \cref{fig:mv_conditioning}.

\subsection{Robust to Noisy Depth}
\begin{figure}
\centering  
\begin{subfigure}{0.32\linewidth}
    \includegraphics[width=\linewidth]{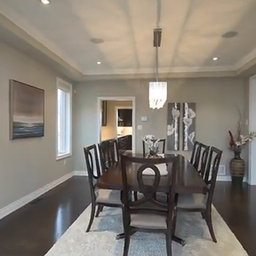}
    \caption*{Source view}
\end{subfigure}
\begin{subfigure}{0.32\linewidth}
    \includegraphics[width=\linewidth]{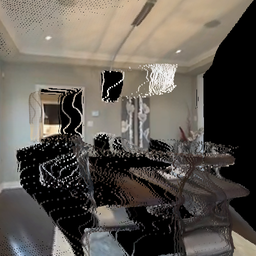}
    \caption*{Warped view}
\end{subfigure}
\begin{subfigure}{0.32\linewidth}
    \includegraphics[width=\linewidth]{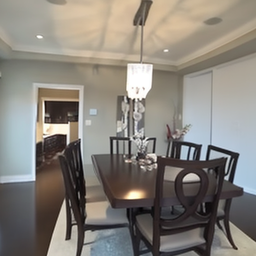}
    \caption*{Predicted view}
\end{subfigure}
\begin{subfigure}{0.49\linewidth}
    \includegraphics[width=\linewidth]{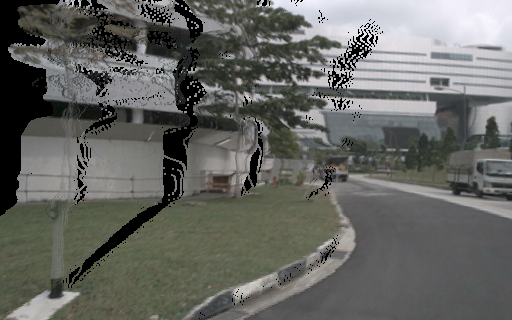}
    \caption*{Warped view}
\end{subfigure}
\begin{subfigure}{0.49\linewidth}
    \includegraphics[width=\linewidth]{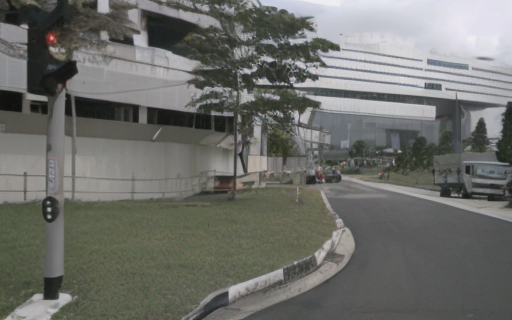}
    \caption*{Predicted view}
\end{subfigure}
\caption{Robustness to noisy depth on RealEstate10K~\cite{zhou2018stereo} and nuScenes~\cite{caesar2020nuscenes}.}
\label{fig:noisy_depth}
\end{figure}

\begin{figure}
\centering
\begin{subfigure}{0.32\linewidth}
    \includegraphics[width=\linewidth]{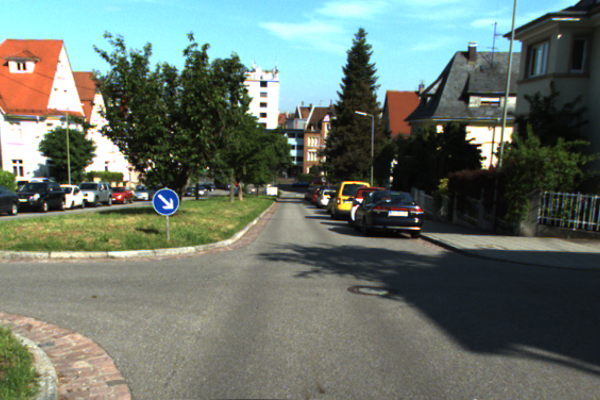}
    \caption*{Source view}
\end{subfigure}
\begin{subfigure}{0.32\linewidth}
    \includegraphics[width=\linewidth]{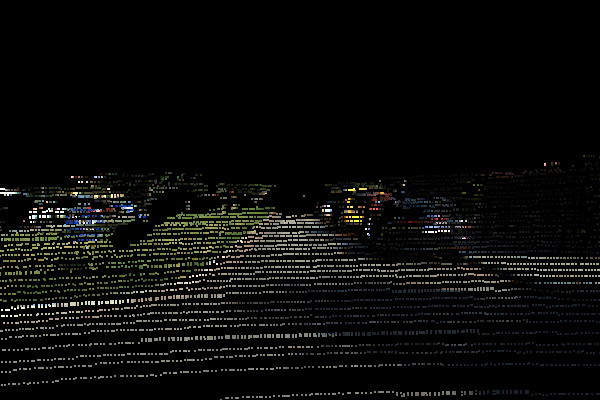}
    \caption*{Warped view}
\end{subfigure}
\begin{subfigure}{0.32\linewidth}
    \includegraphics[width=\linewidth]{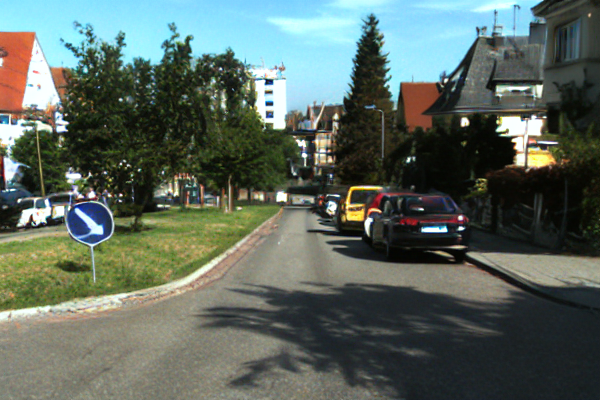}
    \caption*{Predicted view}
\end{subfigure}
\caption{Robustness to sparse depth on KITTI-360~\cite{liao2022kitti}.}
\label{fig:sparse_depth}
\end{figure}

Additionally, when leveraging off-the-shelf MDE models~\cite{ranftl2020towards, bhat2023zoedepth}, the generated depth maps $D^{r}$ used for wrapping and establishing point correspondences can be noisy. However, our reference attention mechanism additionally injects both semantic and geometric multi-resolution information from the reference image as a guiding signal. This enables the model to be naturally more robust to noisy or incomplete depth within the generative prior compared to the explicit warping~\cite{rockwell2021pixelsynth, cai2023diffdreamer, chung2023luciddreamer, shriram2024realmdreamer} approaches.
We show in \cref{fig:noisy_depth} a scenario where using monocular depth can lead to ill-warping artifacts and \cref{fig:sparse_depth} where sparsity of LiDAR points makes inpainting infeasible. As said, PointmapDiff demonstrates a strong ability to fill in missing regions and correct inaccurate geometry, highlighting its capacity to understand scene structure without overfitting to misaligned inputs.

\subsection{LiDAR-aligned Generation}
\label{sec:lidar_align}
\begin{figure}
\centering  
\begin{subfigure}{0.49\linewidth}
    \includegraphics[width=\linewidth]{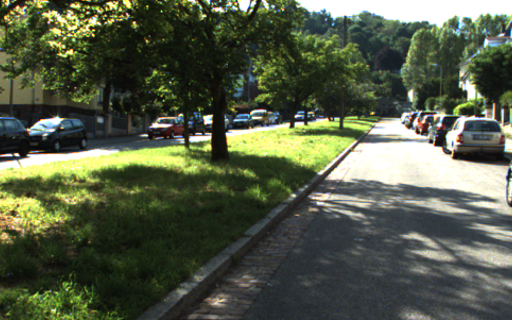}
\end{subfigure}
\begin{subfigure}{0.49\linewidth}
    \includegraphics[width=\linewidth]{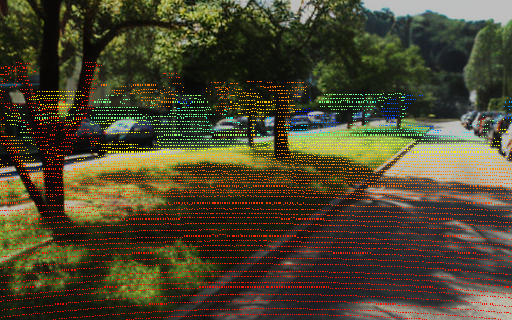}
\end{subfigure}
\begin{subfigure}{0.49\linewidth}
    \includegraphics[width=\linewidth]{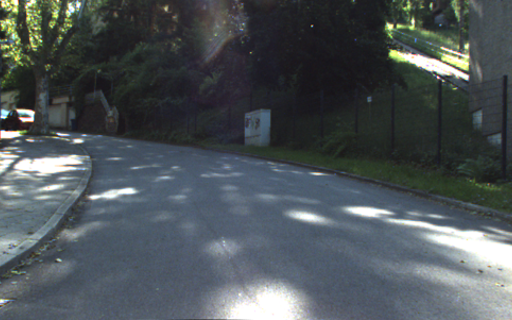}
\end{subfigure}
\begin{subfigure}{0.49\linewidth}
    \includegraphics[width=\linewidth]{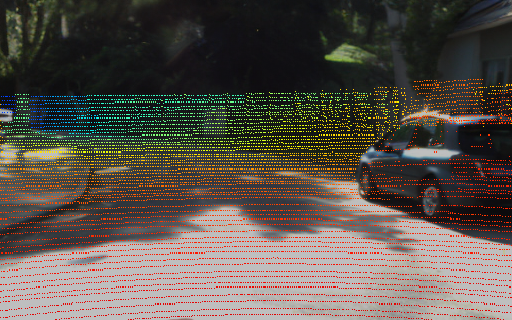}
\end{subfigure}
\begin{subfigure}{0.49\linewidth}
    \includegraphics[width=\linewidth]{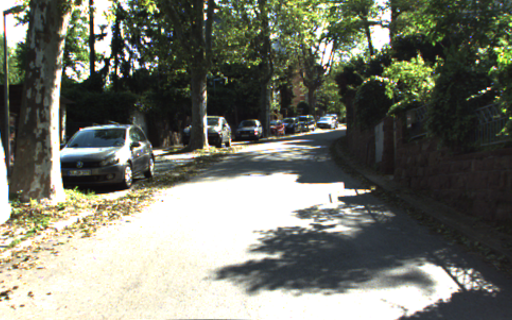}
\end{subfigure}
\begin{subfigure}{0.49\linewidth}
    \includegraphics[width=\linewidth]{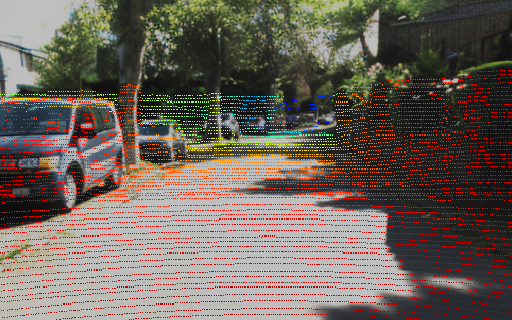}
\end{subfigure}
\begin{subfigure}{0.49\linewidth}
    \includegraphics[width=\linewidth]{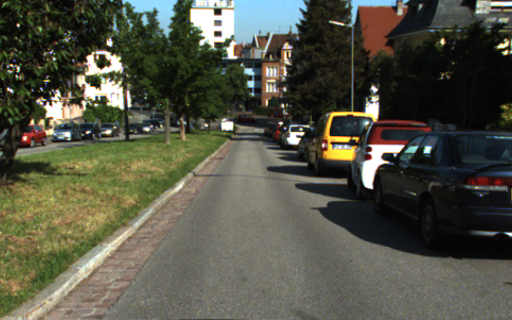}
    \caption*{Source views}
\end{subfigure}
\begin{subfigure}{0.49\linewidth}
    \includegraphics[width=\linewidth]{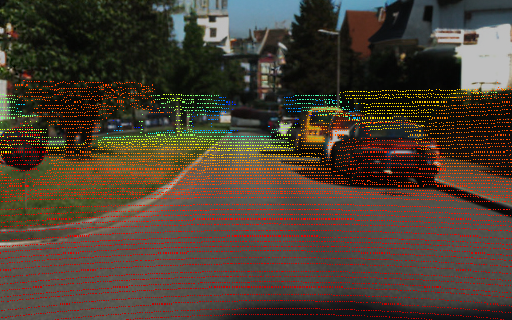}
    \caption*{Predicted views}
\end{subfigure}
\caption{We overlap projected LiDAR points onto the images on KITTI-360~\cite{liao2022kitti}, showing that our generated views are aligned with the geometry given by the LiDAR.}
\label{fig:lidar_align}
\end{figure}

\cref{fig:lidar_align} shows that by integrating LiDAR data from regions not visible in the reference views, we can generate images that accurately adhere to the underlying LiDAR measurements, ensuring high-fidelity scene reconstruction with enhanced geometric consistency.

\subsection{Limitations and Future Work}
In this section, we discuss the primary limitations of our work and propose some preliminary mitigation strategies for future research.
The diffusion model is trained to remove noise, but stochasticity persists in the final prediction. Moreover, lossy compression of VAE can remove contents in the prediction, particularly in small details. When using these images to train 3DGS, this can lead to blurry results, even in regions that are well-observed in the training ground truths, leading to lower PSNR and SSIM during interpolation. An interesting focus for future work would be to study the uncertainty in both the 3DGS and diffusion models. This involves updating only the regions where 3DGS is uncertain, while the diffusion model is confident, and vice versa.
Additionally, to adapt to dynamic scenes, it is necessary to introduce a temporal dimension to the diffusion model. This approach, commonly used in video diffusion, could complement object movement where static point maps cannot provide correct correspondences.

\end{document}